**Master's Thesis**
MSC-026

# Obstacle avoidance-driven controller for safety-critical aerial robots

by
Johann Lange

Supervisors: Prof. Dr.-Ing. R. Seifried
Asst.-Prof. Dr. K. Sreenath

Hamburg University of Technology
Institute of Mechanics and Ocean Engineering
Prof. Dr.-Ing. R. Seifried

Hamburg, September 2019

# Contents

# Chapter 1

# Introduction

Obstacle avoidance for aerial robots is a non-trivial task. Most aerial robots can stop and even hover in-flight and have the maneuverability to avoid obstacles. However, due to the underactuation of the aerial robot system and based on the current position and orientation with respect to the obstacle(s), the robot cannot necessarily avoid the obstacle easily. Therefore, both the underactuation as well as the relative orientation of the robot have to be accounted for. There are two types of obstacles: known and unknown obstacles. Known obstacles are obstacles whose position is known prior to the start of the mission, that is when the aerial robot starts its flight. Those include the limits of the flight space, such as walls, trees, or other fixed objects. These kind of obstacles can be safely avoided using state-of-the-art path-planning algorithms [LaiEtAl16, JiEtAl17]. Unknown obstacles are not known prior to mission start, these obstacles occur randomly. Common examples are humans, other robots, or other moving objects. Nevertheless, even fixed objects can be unknown prior to mission start, if for example the flight space is not completely mapped. Therefore, an algorithm is needed which can adapt to the environment in real-time and guarantee obstacle avoidance while keeping the aerial robot stable.

The following work is divided into multiple chapters. Chapter 2 gives a brief overview of the currently developed technologies regarding aerial robots and obstacle avoidance. In chapter 3, the development and design of the aerial robot is discussed. The chapter also explains the used soft- and hardware necessary for the flight of the aerial robot. This also includes external systems like the motion capture system, see subsection 3.2.2. The controls for the aerial robot are discussed in chapter 4. First, the general control scheme for aerial robots is shown, followed by a simple PID-like controller and an optimal control approach for obstacle avoidance. Finally, the Model Predictive Control (MPC) approach is highlighted. The aforementioned controllers are validated in chapter 5 to prove the ability for obstacle avoidance. This chapter also contains the flight tests conducted at the flight space of the University of California, Berkeley. Lastly, chapter 6 summarizes the work of this thesis as well as the contributions and gives an outlook on the next steps for this project.

## 1.1 Problem Statement

Consider the above described scenario of an aerial robot flying through an unknown and unmapped area. While the aerial robot would know its current state, due to available sensor data, and possibly the final state, it may have incomplete knowledge of the possible distribution of static and dynamic obstacles at every time step. Therefore, it might not be able to directly fly from one point in space to any other on the shortest path. Depending on the environment the robot could encounter everything from trees, wall, humans or other aerial robots. In order to keep the aerial robot and its surroundings safe at all time, a collision is to be avoided at all costs.

Assume that the aerial robot is already equipped with a technology which enables obstacle detection, for example an RGB(D)-camera or a different sensor, providing depth information. Additionally, an obstacle avoidance algorithm is needed to safely reach the robot's destination without colliding with the detected obstacles. Detecting obstacles only when they are already close will lead to a fast changing environment for the controller to navigate in. To react to the changes in the environment and safely navigate through it, the controller has to be able to take the obstacles into account. In general, such a controller has been developed and exclusively been used in simulations [WuEtAl16b]. This controller only takes the current position, velocity, and acceleration of the aerial robot and the obstacle(s) into account. It cannot predict the next time-step and optimize the robots trajectory over a longer time period. Such a prediction could potentially increase the controller's performance as the robot, given a known path for the obstacles, would not need to react to obstacles coming close but not crossing paths. Such scenarios are not uncommon in highly dense environments or in cases of multi-agent flights.

The scenario considered in this thesis is a flight through an unknown area, which might be inaccessible to humans, for example in dangerous rescue missions. The aerial robot can no longer rely on other systems, it has to operate fully autonomous. Especially a secondary ground station which can compute the control outputs is not available. Thus, a platform has to be developed which can handle the necessary computations for obstacle avoidance on-board and in real-time.

## 1.2 Mathematical Preliminaries

Throughout this work, some mathematical preliminaries are required. The first is the hat map, used to transform a cross product operation into a matrix multiplication. For that, let $\hat{\cdot}\colon \mathbb{R}^3 \to \mathfrak{so}(3)$ be defined as the hat map and its inverse $\check{\cdot}\colon \mathfrak{so}(3) \to \mathbb{R}^3$ as the vee map [MurrayEtAl94, Hall15] such that:

$$\hat{\boldsymbol{x}}\boldsymbol{y} = \boldsymbol{x} \times \boldsymbol{y}, \quad \forall \quad \boldsymbol{x}, \boldsymbol{y} \in \mathbb{R}^3 \tag{1.1}$$

$$\check{\hat{\boldsymbol{x}}} = \boldsymbol{x}, \quad \forall \quad \hat{\boldsymbol{x}} \in \mathsf{SO}(3). \tag{1.2}$$

Furthermore, throughout this thesis, whenever Euler angles are used, it shall be noted, that those are in fact the commonly used $z$-$y$-$x$-Euler-angles, also known as the roll-pitch-yaw-angles.

# Chapter 2

# State of the Art

Over the last years, aerial robots have become an attractive platform for robotics and control research [LeeEtAl10a, LeeEtAl10b, MellingerEtAl11, HehnEtAl11, CaiEtAl14]. Especially the coordinated use of aerial robots has gained popularity. Using swarms of aerial robots [KushleyevEtAl13] enables for example surveillance of large areas [HrabarEtAl10, AcevedoEtAl13] and transportation of heavy payloads with small scale robots [SreenathEtAl13a, SreenathEtAl13b] including grabbing payloads in-flight [ThomasEtAl14]. Aerial robots have also found application in the civil sector, supporting or leading search and rescue missions [TomicEtAl12, Murphy12, GhamryEtAl17].

However, for aerial robots to work fully autonomously, they need to be equipped with obstacle detection and avoidance algorithms [KoyasuEtAl02, KoyasuEtAl03, BeardEtAl03]. The obstacle detection algorithms are often offloaded to a ground station [CarloniEtAl13], limiting the range of the aerial robot such that the robot is always able to communicate with the ground station. But efforts have been made for intercommunication of swarms to overcome the problem of the limited-range [BeardEtAl03, PalatEtAl05]. Nevertheless, to maneuver fully autonomously, independent from other aerial robots or a ground station, aerial robots have to detect obstacles and react accordingly using only on-board hard- and software. Thus, obstacle avoidance is a very important research area, especially for swarms in which the robots can also interfere with each other, both on ground [BorrmannEtAl15] and air-bound [AlejoEtAl09, MoriEtAl13].

To solve the problem of obstacle avoidance, path-planning is often applied [Hrabar08, PepyEtAl06, ShenEtAl15, JiEtAl17, ZhangEtAl18]. The path-planing algorithm calculates an optimal path to reach a certain state while avoiding all obstacles. The resulting path, often referred to as reference trajectory is then send to a controller. In the case of an aerial robot, this is most likely a position controller. The position controller in turn will calculate a desired thrust vector to minimize the error between the current and reference position. Nevertheless, a simple position controller is not constrained by an obstacle avoidance algorithm. Thus, there is no guarantee, it will follow the reference trajectory close enough to avoid the obstacle. Augmenting the position controller with an obstacle avoidance constraint will guarantee avoidance. The next step in the control chain is often an attitude controller, minimizing the attitude error. Its input is the desired thrust vector outputted by the position controller. Most aerial robots have two degrees of underactuation, see [WuEtAl16a] and section 3.2. Therefore, their potential to avoid obstacles depends on their relative orientation to the obstacle [WuEtAl16a, WuEtAl16b]. Thus, obstacle avoidance algorithms are applied to the attitude controller. The drawback is, that the attitude controller cannot influence the position of the aerial robot, as it has no knowledge about where the aerial robot is in space compared to the reference trajectory. The topic of obstacle avoidance on the position and attitude level and its impact on the controller design are further discussed in subsection 4.4.2.

Due to the weight limitation of aerial robots, flight time is always a problem. To maximize the flight time [TagliabueEtAl19] optimal control [WuEtAl16a, WuEtAl16b] is often used instead of ordinary control [LeeEtAl10b]. Optimal control can optimize flight by reduction unnecessary movements. It can also impose constraints, guaranteeing exponential stability and safety. Yet, optimal control poses higher strain on the hardware than ordinary control laws. To guarantee safety, a new set of constraint function was recently developed [AmesEtAl14b, AmesEtAl19]. Based on exponentially stable Control Lyapunov Functions (CLF) [GhandhariEtAl01, AmesEtAl14a] this new controller guarantees safety at all times and uses a barrier function [TeeEtAl09b, TeeEtAl09a] as a safety constraint. The resulting controller is referred to as a Control Barrier Function (CBF). It will optimize the control output at each time step, trying to achieve exponential stability while never violating the safety constraint defined by the barrier function.

In general, obstacle avoidance controllers do not predict future positions of the aerial robot and the obstacle. Thus, one-step optimizers may not follow the optimal path around obstacles. A possible solution is Receding Horizon Control (RHC), often referred to as Model Predictive Control (MPC) [KwonEtAl05, GrimmEtAl05, ShenEtAl15]. These controllers optimize over a receding time horizon to calculate the optimal set of control outputs up until a horizon. It has been proven to work with aerial robots [MuellerEtAl13, BanguraEtAl14] and also with safety-guaranteeing-constraints [WuEtAl18]. Currently, there is no controller combining the safety guarantee described in [AmesEtAl14b, WuEtAl16a, WuEtAl16b, AmesEtAl19] with the idea of MPC.

Parallel to the rapid development of control algorithms, in the last years, numerous software-platforms for aerial robot research have been developed. Most notably are Crazyflie [GiernackiEtAl17], Ardupilot [ArduPilot09], Px4 [MeierEtAl15], and Rosflight [ROSflight19], all of which are open-source and designed for research, thus allowing heavy customization. Crazyflie has been developed for small-scale aerial robots, especially useful for swarms. The software is designed to be ran on the corresponding hardware, a tiny 9 g aerial robot. In comparison, Ardupilot, Px4, as well as Rosflight all have been developed independent from a specific hardware. While there exist a custom flight controller for Px4, called Pixhawk [pixhawk18], the software can easily be ported onto different hardware. With the rising popularity of Ros [QuigleyEtAl09], an open-source robotics framework, Rosflight has been developed for easy integration of flight control into an existing Ros installation. Contrary to that, Ardupilot and Px4 both use an intermediate protocol, called Mavlink, to communicate with Ros and allow for the integration of flight controllers into Ros.

To deploy the developed control algorithm, different scales of aerial robots are available with a maximum take-off mass of <10 g up to around 25 kg [CaiEtAl14, GiernackiEtAl17]. A commonly used platform for aerial robots in research is the "DJI Flamewheel" series [DJI12], designed for a maximum take-off mass of around 1 kg to 5 kg. The Flamewheel platform provides a light-weight structure for the aerial robot with multiple mounting point for different accessory, like a flight controller, battery, and a gimbal equipped with a camera.

# Chapter 3

# Test Platform Development

This chapter discusses the ideas behind the development and design of the aerial robot used in this work. While the number of commercially available aerial robots increased, so did the supply [CaiEtAl14], most aerial robots are trimmed for flight time and use only light weight control schemes with no on-board obstacle avoidance. In addition, most commercially available aerial robots are closed-source, regarding the hard- and software. Still, to incorporate obstacle avoidance, an aerial robot with access to its software and potentially hardware is needed. The following chapter discusses the design and development of an aerial robot, which is open-source and equipped with the necessary hard- and software for obstacle detection and avoidance.

## 3.1 Requirements

As mentioned above, the aerial robot has to be customizable in terms of hard- and software. To detect obstacles, the aerial robot needs to be equipped with a device, which is able to detect obstacles, like an RGBD-camera. In addition to the RGB color channels, a RGBD-camera also provides a channel with depth information for each pixel. But the aerial robot also has to have a device which can handle the information streams. This additional channel can be used for the obstacle detection as well as the pose information of the robot. Also, the device has to be able to calculate the control laws. Lastly, the device has to determine the control laws for obstacle avoidance and execute the control commands using the available actuators.

Aerial robots have to be designed around a low mass with high thrust. The higher the thrust to mass ratio is, the better the aerial robots flight performance will be. With a ratio of approximately 1, the aerial robot can barely hover at a constant height. But with a ratio of 2 and even higher, it can accelerate and decelerate quickly, allowing for quick turns to avoid obstacles. For that reason, the aforementioned device has to be as light as possible. Increasing the size to add a stronger actuation system, is, in theory, a valid option. But as the available flight space is limited to $3\,\mathrm{m} \times 5\,\mathrm{m} \times 6\,\mathrm{m}$ ($x \times y \times z$), increasing the size of the aerial robot will significantly decrease its relative size compared to the flight space. Therefore, the size of the aerial robot shall not exceed one-tenth of the size in each direction. Otherwise, the maneuverability of the robot will be negatively influenced.

## 3.2 Hardware

A reliable and robust platform is needed for testing the proposed control strategy in chapter 4. A commonly used platform for aerial robots is the quadrotor, also known as quadrocopter, named after its four rotors, generating the thrust and moments. Of the six degrees of freedom in $\mathsf{SE}(3)$, the quadrotor can only directly manipulate the three rotations in $\mathsf{SO}(3)$ and only one translation in $\mathbb{R}^3$, that is the translation along the thrust axis, commonly defined as the $z$-axis. To manipulate the translations along the $x$- and $y$-axis, the robot has to rotate. Therefore, the system has two degrees of underactuation. Nevertheless, the problem of underactuation holds true for other multirotors, such a hexa- and octorotors, as the thrust can still be only applied in one direction. The resulting implications will be discussed in section 4.2.

The quadrotor is built around the commonly used "DJI Flamewheel F330" [DJI12] frame. The frame offers multiple mounting points for additional hardware, such as flight-controllers, batteries, and cameras. With its beam length of 330 mm it is big enough to carry enough additional hardware without negatively impacting the required flight space to operate. Bigger frames limit the quadrotor's maneuverability. In contrast, a smaller frame reduces the available mount space for additional hardware.

The quadrotor is actuated by four Emax MT 2208 II [Emax18] motors with $8 \times 4.5$ propellers with matching ESCs (Electronic Speed Controller). This configuration creates a maximum thrust of $f_{\max} = 28.51\,\mathrm{N}$ and maximum moments of $\boldsymbol{\tau}_{x,\max} = 1.66\,\mathrm{N\,m}$, $\boldsymbol{\tau}_{y,\max} = 1.66\,\mathrm{N\,m}$, and $\boldsymbol{\tau}_{z,\max} = 0.215\,\mathrm{N\,m}$. For the definition of $f$ and $\boldsymbol{\tau}$ see section 4.1 and Figure 4.1. Using a 3S LiPo battery, this setup can achieve a flight time of approximately 10 min.

Additional hardware is required to calculate the control commands and to send them to the ESCs. Most flight controllers, like the Pixhawk [pixhawk18], are only able to do this for simple control laws in real-time. Calculating more advanced control laws requires computationally more powerful hardware. For that reason, a NVIDIA Jetson TX2 [NVIDIA19] is used. With its six CPU cores, 256 CUDA GPU cores, and 8 GB RAM, it can handle optimal control. It can also run ROS [QuigleyEtAl09] to e.g. combine multiple sensor readings. Nonetheless, the Jetson TX2 cannot communicate with the ESCs. This is done through the Aerocore 2 [Gumstix18] running PX4 [GiernackiEtAl17], an open-source software for flight control of aerial robots, see subsection 3.3.2. In short, the Aerocore 2 receives the desired control outputs calculated by the Jetson TX2 and translates it to ESC readable signals, which control the motors. More on the exact data- and workflow can be found in section 3.3.

### 3.2.1 Camera

The quadrotos is equipped with a stereovision camera, which can be used for obstacle detection and pose estimation of the quadrotor. The latter feature can be especially useful if the quadrotor was to be flown in open areas where no exact positioning service is available.

For the development of this platform, three different cameras were available, each with distinct features and their consequently advantages and disadvantages. The three cameras are the “ZED Mini” [Stereolabs19] by Stereolabs, the “D435i” [Intel18] as well as the “T265” [Intel19], which are both by Intel. The features of which are listed in Table 3.1.

Table 3.1: Features of different RGB(D)-cameras.

| Feature | ZED Mini | Intel D435i | Intel T265 |
|---|---|---|---|
| Resolution | from $672 \times 376$ to $2\,208 \times 1\,242$ | $1\,920 \times 1\,080$ | $848 \times 800$ |
| Framerate | up to 100 fps at $672 \times 376$ | 30 fps | 30 fps |
| Field of View (diag.) | 110° | $(95 \pm 3)°$ | $(163 \pm 5)°$ hemispherical |
| Depth technique | passive | active | none |
| Depth resolution | same as RGB | up to $1\,280 \times 720$ at 90 fps | – |
| Depth range | 12 m | 10 m | – |
| Pose information | ✓ | ✗ | ✓ |
| Accelerometer | up to 800 Hz | 250 Hz | 62 Hz |
| Gyroscope | up to 800 Hz | 400 Hz | 200 Hz |
| Computation | offloaded | on-device | on-device |

As listed in the table, the ZED Mini offers acceptable pose estimation and an IMU, but it is offloading all computations to a CUDA device resulting in the necessity of a high performance NVIDIA GPU. While the Jetson TX2 is equipped with 256 CUDA cores, even on the lowest possible resolution, the depth map and the pose estimation are only available in less than 10 fps, which is insufficient for proper pose estimation and especially velocity estimation. Therefore, this camera is not a valid option. The Intel cameras on the other hand do all their calculations on-board, running independently from the Jetson TX2. This design allows for higher framerate streams for both image and pose data. While the T265 delivers high accuracy pose estimation with a supposed drift of $< 1\,\%$, only the D435i natively delivers a depth image. As all flight tests where conducted in a secured flight space, equipped with a motion capturing setup, see subsection 3.2.2, pose information could be obtained using said setup. Therefore, the D435i was chosen as the primary camera for the quadrotor platform.

### 3.2.2 Motion-Capture

Motion-capture is a technique used in the industry [MoeslundEtAl01, PullenEtAl02] and research [YamaneEtAl09, ThomasEtAl13, ThomasEtAl14] to capture highly accurate pose information of objects in $\mathsf{SE}(3)$. To do so, the setup uses multiple cameras to track reflective markers mounted on an object. As the positions of the markers on the object are known, the position and orientation of the object can be calculated based on the marker's position in $\mathbb{R}^3$. Such a setup is exemplary shown in Figure 3.1. Using its cameras, the system can calculate the current position and orientation of any marker array. The markers are attached to the quadrotor in between the arms. Adding a filter provides additional velocity information. To incorporate the data into the quadrotor system, a VRPN (Virtual Reality Peripheral Network) publishes the pose data via ROS [BovbelEtAl19]. From here, every other system running ROS can subscribe to the data and for example log it or process the received pose signals.

## 3.3 Software

The quadrotor platform consists of multiple devices with different interfaces, which have to be accounted for. While most software runs in ROS on the Jetson TX2 the interface to the Aerocore 2 is managed via MAVLINK. This interface is fully handled by MAVROS and PX4, see subsection 3.3.2. Nevertheless, some changes had to be made to pass the calculated control outputs from the Jetson TX2 through to the ESCs. Figure 3.2 provides a basic overview over the systems involved. As the only external system, the motion capture system publishes the data to Ros from where all other Ros-nodes can subscribe and read the messages. Same holds true for the camera system. Ros provides the interface for the camera to publish the data. Likewise, the controller and Mavros read and write from and to the Ros message system. On the Aerocore 2, Px4 recieves the control outputs and sends to the mixing module. The mixing module calculates the rotation speed of the motors based on the control outputs and communicates with the ESCs, which again control the motors.

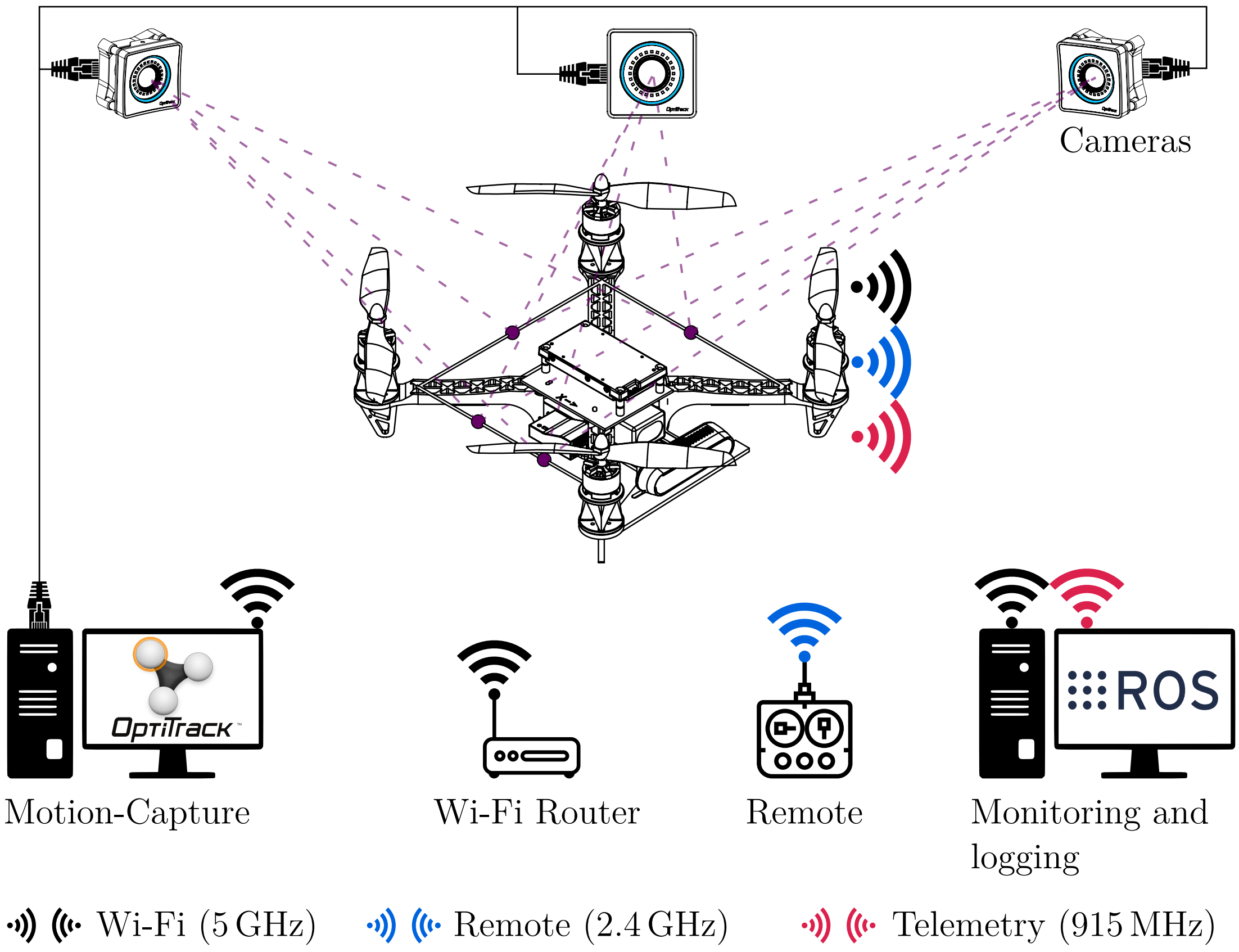

Figure 3.1: Overview of all systems necessary for flight tests.

### 3.3.1 Ros

The Robot Operating System (Ros) has been developed as an open source framework for distributed robotic systems [QuigleyEtAl09]. It is mainly written in C++ and Python but is also supported by Matlab. Ros provides thread scheduling, package management and message distribution. In Ros, executables are called nodes, which, based on the interconnection between the nodes, form a graph representing the architecture of the robotic system. The messaging system in Ros provides each node with a publish/subscribe-system and a service-call-system. The former allows nodes to broadcast predefined messages into a certain topic, a named bus. This topic can then be subscribed to by other nodes. It also allows for multiple nodes to broadcast and to subscribe, an omnidirectional message exchange. Ros-service on the other hand, is a unidirectional peer-to-peer connection between two nodes. Here, a node can call the service of another node to request a message instead of waiting for the other node to publish it. Another difference is, that using a service, the information in the message can be adapted to the request of the requesting node, a kind of customized message compared to the public messages using a Ros-topic.

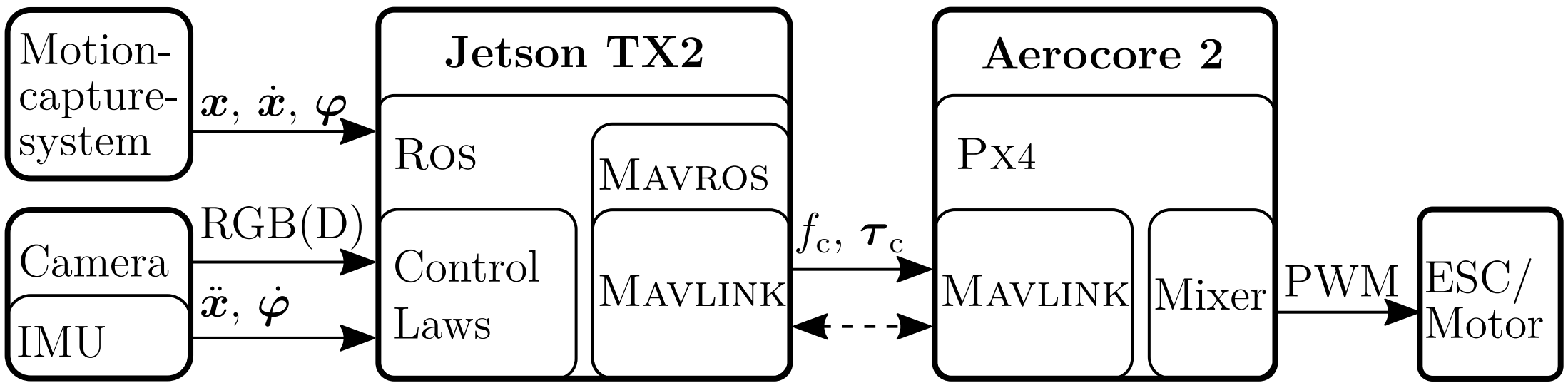

Figure 3.2: General schematic of the high-level software design.

Ros is able to run on distributed systems with different hardware. A Ros-master handles the communication between the nodes and acts as a central point of contact for nodes to initially connect to. All nodes can be run in multiple instances and on multiple devices in parallel. This allows for a modular software design, for example with a node reading sensor values on multiple robots in a swarm. Moreover, nodes can quickly be exchanged with other nodes, or a new version of the same node, possibly even in operation. Using Ros has the benefit that new controllers can easily be included, as they only have to call the corresponding services and subscribe to the necessary topics to gather the input data and publish the control output.

### 3.3.2 Px4

Of the in chapter 2 mentioned available software packages, only Px4 is available pre-build for the Aerocore 2, it is used as the operation system. With its support for multiple aerial robots, such as quad-, hexa-, octorotors, helicopters, and even wheeled robots, as well as different hardware platforms, it is well supported by a huge community.

Px4 does not support $f$-$\boldsymbol{\tau}$-control via the companion devices (Jetson TX2). By default, it only allows higher level control, such as path-planning, on the compan-ion device, because the delay cannot be verified to be low enough for the system to be stable. But as the Jetson TX2 and the Aerocore 2 share a high speed connection, it is possible to send $f_c$ and $\boldsymbol{\tau}_c$ not only at a sufficient frequency and a low enough delay for the control to be stable. The proof of concept can be found in subsection 5.2.1. Therefore, a modified version of [MeierEtAl18] is used.

To send the normalized $f_c$ and $\boldsymbol{\tau}_c$ from the Jetson TX2 to the Aerocore 2, a new Mavlink-plugin had to be written. The plugin receives $f_c$ and $\boldsymbol{\tau}_c$ from Ros via Mavros, converts it to data readable by the Mavlink-protocol, sends it via the Mavlink-bridge to the Aerocore 2, converts it back to a message readable by Px4 and distributes it inside of Px4.

The values are read by the attitude control module inside of Px4, overwriting the control inputs from the default controller [D'AndreaEtAl13]. As the default controller is still running in the background, it can be used as a fallback in case the connection between the Jetson TX2 and the Aerocore 2 drops out. The final control input is then passed to the mixer module. This module uses a matrix similar to the matrix described in [MellingerEtAl11] to calculate the individual PWM values for each motor.

### 3.3.3 Cvxgen

Often, an ordinary controller is not sufficient to guarantee stability and obstacle avoidance. One possible solution is using an optimal controller, as derived in section 4.3. Using optimal control relies on solving an optimization problem. The optimization problems for the CLF-CBF-QP, see section 4.3, and the MPCBF, see section 4.4, can be classified as convex [BoydEtAl10]. Those can be solved faster and more efficient due to having one global optimum rather then multiple local optima. These problems can, for example, be solved using linearly constrained quadratic program [Hildreth57, NocedalEtAl06], an efficient way of solving convex problems. Most solvers are written to be ambiguous, leading to slower overall solve times. However, to run an optimization in real time, the solver itself has to be faster than the loop time of the controller. For that reason, Cvxgen [MattingleyEtAl12] was chosen. It is a solver designed for these types quadratic programs, designed to run as fast as possible, even on embedded systems. In comparison to traditional solvers, Cvxgen runs faster, due to running custom C-code. According to [MattingleyEtAl12], the solver created by Cvxgen reduces the solving time by 10 000 compared to ordinary solvers, such as Cvx.

Suppose the following linearly constrained quadratic program:

$$\boldsymbol{x}^* = \underset{\boldsymbol{x}}{\text{minimize}} \; \frac{1}{2}\boldsymbol{x}^\mathsf{T}\boldsymbol{Q}\boldsymbol{x} + \boldsymbol{f}^\mathsf{T}\boldsymbol{x}, \tag{3.1}$$

$$\text{subj. to } \boldsymbol{A}\boldsymbol{x} \leq \boldsymbol{b}, \tag{3.2}$$

$$\boldsymbol{A}_{\text{eq}}\boldsymbol{x} = \boldsymbol{b}_{\text{eq}}, \tag{3.3}$$

where $\boldsymbol{x} \in \mathbb{R}^n$ is the optimization variable, $\boldsymbol{x}^* \in \mathbb{R}^n$ is the optimal solution, $\boldsymbol{Q} \in \mathbb{S}_+^n$ is a symmetric weight matrix, often referred to as the quadratic penalty, $\boldsymbol{f} \in \mathbb{R}^n$ is the linear cost or bias term, $\boldsymbol{A} \in \mathbb{R}^{m \times n}$ and $\boldsymbol{b} \in \mathbb{R}^m$ are the matrix and vector, respectively, used to represent the $m$ inequality constraint, and $\boldsymbol{A}_{\text{eq}} \in \mathbb{R}^{m_{\text{eq}} \times n}$ and $\boldsymbol{b}_{\text{eq}} \in \mathbb{R}^{m_{\text{eq}}}$ are the matrix and vector, respectively, used to represent the $m_{\text{eq}}$ equality constraint. To create a highly efficient solver for the linear quadratic program, the following code would be used:

```
dimensions
  n   = 10 # states of x
  m   = 5  # num. of ineq constraints
  meq = 3  # num. of eq constraints
end

parameters
  Q   (n, n) psd # quadratic penalty (positive semi definite)
  f   (n)        # linear cost term
  A   (m, n)
  b   (m)
  Aeq (meq, n)
  beq (meq)
end

variables
  x (n)
end

minimize
  quad(x, Q) + f'*x
subject to
  A*x   <= b
  Aeq*x == beq
end
```

Furthermore, CVXGEN can also create solvers for MPC, see section 4.4, natively [MattingleyEtAl11]. While linear Model Predictive Control problems can be converted to a quadratic program [BuijsEtAl02], the problem can also be written down directly in CVXGEN. For that, mostly the minimization-block is changed with an alternative minimization function:

```
minimize
  sum[t=0..T](quad(x[t], Q) + quad(u[t], R)) + quad(x[T+1], Q_final)
subject to
  x[t+1] == A*x[t] + B*u[t], t=0..T     # dynamics constraints.
end
```

Due to the similarity of both codes, the definition of the corresponding dimensions, parameters, and variables has been omitted.

Since the generated code is written in C, CVXGEN is highly portable, especially considering its use case for embedded systems. Furthermore, C can be easily incorporated into C++-code, which is natively supported by ROS. Therefore, CVXGEN can be used in combination with ROS, making it a very good candidate for systems running ROS.

### 3.3.4 Controller

The control laws with the necessary in- and outputs are written in C++ running as ROS-nodes. Thus, each task is handled by a specific node. The fundamental tasks are the input and output from the virtual controller space. An input-node gathers all different inputs and provides a ROS service for each, such that other nodes always fetch the most recent data. The corresponding inputs are the IMU and gyroscopic readings from either the camera or the Aerocore 2, the position, orientation, and velocity from the motion capture system, via the VRPN. Also, all sensor data is filtered using a second order low pass filter to reduce the noise in the system. Depending on the used camera, the pose can also be acquired from the camera, or, as long as an IMU and gyro is present, via a Kalman filter. Thus, the full pose information, $\boldsymbol{x}$, $\dot{\boldsymbol{x}}$, $\ddot{\boldsymbol{x}}$, $\boldsymbol{\varphi}$, $\dot{\boldsymbol{\varphi}}$, is gathered. Furthermore, the orientation is provided in terms of a rotation matrix and the quaternion representation. As the input node handles the decision which IMU, gyro, and position data is used, the logic does not have to be implemented inside the controller nodes. Yet, the camera's video stream is not subscribed to by the input node, it is, if necessary, directly subscribed to by each node. The output node works exactly the other way around. It gathers the control output from the currently active controller via a ROS service and publishes it to the ROS topic. MAVROS then gathers the data and further processes it. There also exists a trajectory planning node, which provides the controllers with the reference trajectory, for example a fixed setpoint or a circular trajectory similar to the one used in section 5.1.

The general framework of the controller is a service call to obtain the pose, a service call for the trajectory, and a service call to pass the control output to the output node. In the case of the CLF-CBF-QP, see section 4.3, or the MPCBF, see section 4.4, CVXGEN is used to solve the optimal control problem. Finally, as for each controller, see chapter 4, there exists a separate node, it can be easily switched among them. Since all controllers are running at all times, even an in-flight replacement is possible.

## 3.4 Final Design

The final quadrotor can be seen in Figure 3.3, both as a rendering and a picture. While the motors had to be raised to make space for the Aerocore 2 and Jetson TX2, in theory this should only improve the aerodynamics. The total mass of the quadrotor comes out to around 980 g. With the maximum total thrust of $f_{\mathrm{max}} = 28.51\,\mathrm{N}$, this results in a maximum positive acceleration of $a_{\mathrm{max}} \approx 29.1\,\mathrm{m\,s^{-2}} > 2g$. Accounting for the $1g$ of earths gravitational acceleration, this still leaves the quadrotor with over $1g$ possible acceleration to perform maneuvers. This is often seen as the lower limit for the maximum acceleration for aerial robots. A maximum acceleration lower than $2g$ would reduce its agility. In total, the platform is equipped with eight motion capturing markers, four on the beams between the arms, two on top of the Jetson TX2, and two by using the nuts on two opposite motors. This placement ensures that there are always at least three markers visible from the viewpoint of the cameras.

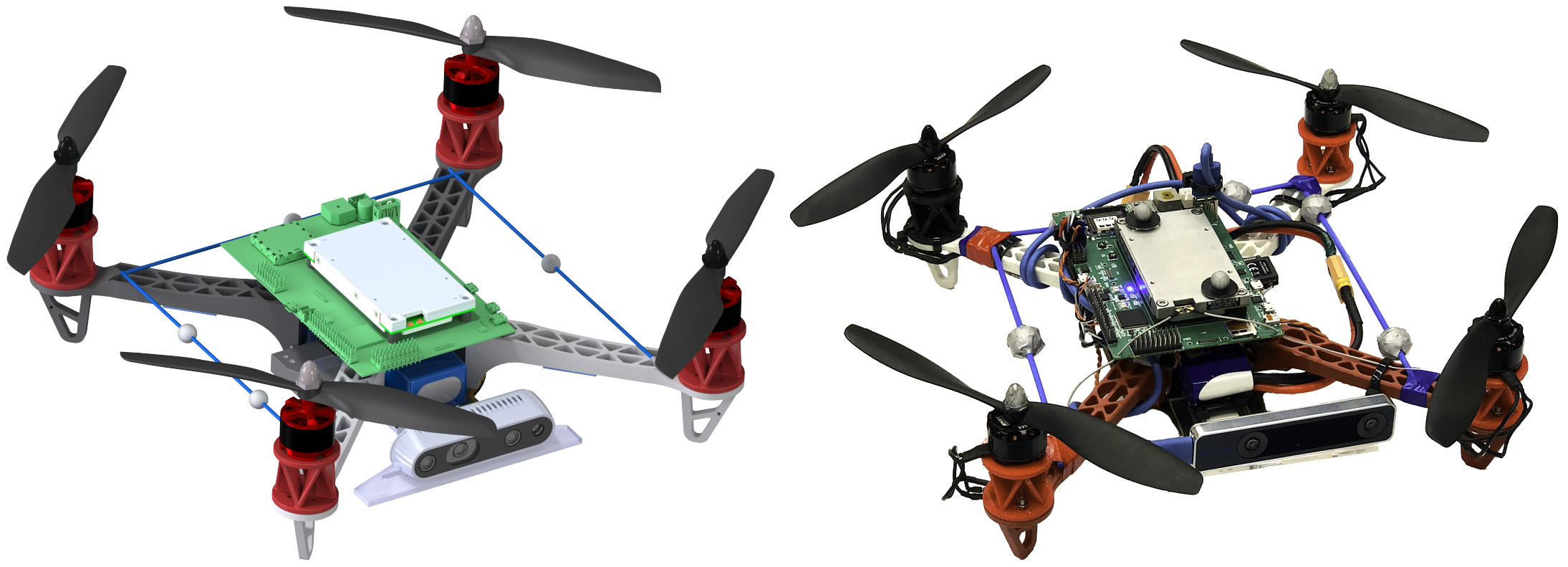

(a) Rendering (with D435i). (b) Real platform (with T265).

Figure 3.3: Final design of the platform.

A remote control has been added to the design, such that an operator can arm, start, fly, land, and disarm the platform if needed. Otherwise, the Jetson TX2's on-board software takes care of it. Still, the remote has another very important function. In case of an emergency or malfunctioning of the software, the motors can be stopped immediately using a “kill switch”. A telemetry system is added for monitoring the status of the Px4 system, for example the battery level and flight mode, either flying manual using the remote or fully autonomously. Another computer running Ros is used to monitor and log the general behavior of the platform, including the pose, velocity, and control in- and outputs.

Lastly, the software running on the Jetson TX2 can be found online [Lange19]. As well as the modified version of Px4 which is used on the Aerocore 2 [LangeEtAl19].

# Chapter 4

# Control

In this chapter, the control of an aerial robot is discussed. First, the robot's dynamics are described in section 4.1. A simplified, yet sufficient model is introduced. Next, in section 4.2, the general control scheme of an aerial robot is reviewed. Here, the Euler angle controller will be introduced, a simple, PD-like controller based on the dynamics of the aerial robot. Afterwards, section 4.3 examines Control Lyapunov Functions (CLF) as well as Control Barrier Functions (CBF) and its application to Optimal Control. Lastly, section 4.4 introduces the Model Predictive Control and how it is adapted to the problem of obstacle avoidance on an aerial robot. Finally, the newly developed controller, the Model Predictive Control Barrier Function (MPCBF), will be presented.

## 4.1 Dynamics of an Aerial Robot

Plausible control actions require a reasonable model of the system's underlying dynamics. There are different models for a quadrotor. While some models, such as those discussed in [PoundsEtAl06, MellingerEtAl11], include each motor with its own thrust, torque, and the resulting forces, said models can be too computationally expensive. To calculate the dynamics in real time, which is required for the Model Predictive Controller in section 4.4, a simple model is better suited as it is computationally cheaper. With this simplification, the dynamics of a quadrotor are the same as for any other (simplified) aerial robot. Such simplified dynamics of an aerial robot are given by [WuEtAl16b, LeeEtAl10a] with its necessary quantities indicated in Figure 4.1:

$$m\ddot{\boldsymbol{x}} = f\boldsymbol{R}\boldsymbol{e}_3 - m \cdot g\boldsymbol{e}_3^{(\mathcal{Q})}, \tag{4.1}$$

$$\dot{\boldsymbol{R}} = \boldsymbol{R}\hat{\dot{\boldsymbol{\varphi}}}, \tag{4.2}$$

$$\boldsymbol{J}\ddot{\boldsymbol{\varphi}} = \boldsymbol{\tau} - \dot{\boldsymbol{\varphi}} \times \boldsymbol{J}\dot{\boldsymbol{\varphi}}, \tag{4.3}$$

where $\boldsymbol{x} \in \mathbb{R}^3$ denotes the position of the robot's center of mass, $f$ $(= \sum_{i=1}^{4} f_i) \in \mathbb{R}$ the total scalar thrust in Newton, $\boldsymbol{R} \in \mathsf{SO}(3)$ the rotation matrix from the inertial frame $\mathcal{I}$ to the robot's frame $\mathcal{Q}$. Furthermore, $\boldsymbol{e}_1, \boldsymbol{e}_2, \boldsymbol{e}_3 \in \mathbb{R}^3$ are the first, second, and third unit vector respectively. Moreover, $m, g \in \mathbb{R}$ are the robot's mass and the local gravitational acceleration, respectively, $\boldsymbol{\varphi} \in \mathbb{R}^3$ is the vector of Euler angles in the body frame $\mathcal{Q}$. Since only very small angles are considered, the angular velocities turn out to be $\boldsymbol{\varphi}$ Likewise, $\boldsymbol{\tau} \in \mathbb{R}^3$ is the moment vector also in the body frame $\mathcal{Q}$, not to be confused with the torques of the single motors, and $\boldsymbol{J} \in \mathbb{R}^{3\times3}$ denotes the moment of inertia of the robot.

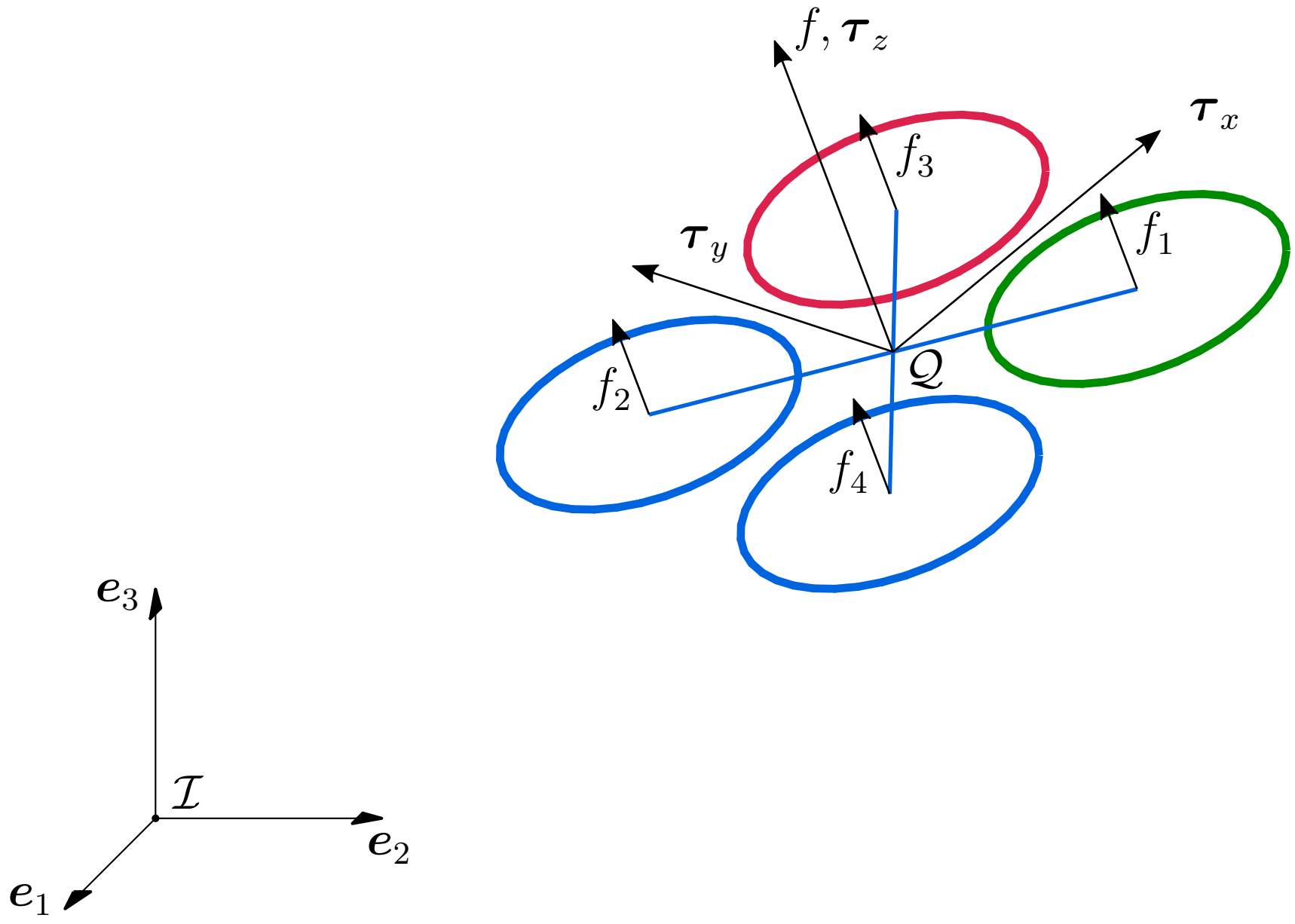

Figure 4.1: Simplified dynamic model of an aerial robot.

## 4.2 General Control Scheme of a Quadrotor

As mentioned in chapter 3, the quadrotor is underactuated. However, most control laws assume a fully actuated system with the virtual dynamics

$$\boldsymbol{v} = \dot{\boldsymbol{x}}, \tag{4.4}$$

$$m\dot{\boldsymbol{v}} = \boldsymbol{f}, \tag{4.5}$$

such that the virtual thrust $\boldsymbol{f} \in \mathbb{R}^3$ can be fully controlled. That way, it is easier to design a controller for the robot's dynamics. Based on that idea, most controllers first calculate the control on the virtual dynamics in $\mathbb{R}^3$, Equations 4.4 and 4.5, and after that on the fully actuated dynamics in $\mathsf{SO}(3)$, Equations 4.2 and 4.3. This splits the controller into two parts, a position controller and an attitude controller. A further advantage of this technique is that, as the attitude dynamics are faster than the position dynamics, both can be run at different updates rates. Therefore, the latter do not have to run at such a high frequency, reducing the strain on the computing device. In general, the position controller is run at as low as approximately 10 Hz, while the attitude controller is most often run at around 250 Hz to 500 Hz for standard applications. This concept is visualized in Figure 4.2. It follows, that the attitude controller could stabilize the aerial robot without input from the position controller. While the position offset would be non-zero, possibly even increasing, the robot's attitude would remain stable. The attitude controllers output, $f_\text{c}$, $\boldsymbol{\tau}_\text{c}$, are the command values for the actuation system of the quadrotor. In most cases those have to be translated to PWM values, based on the matrix described in [MellingerEtAl11].

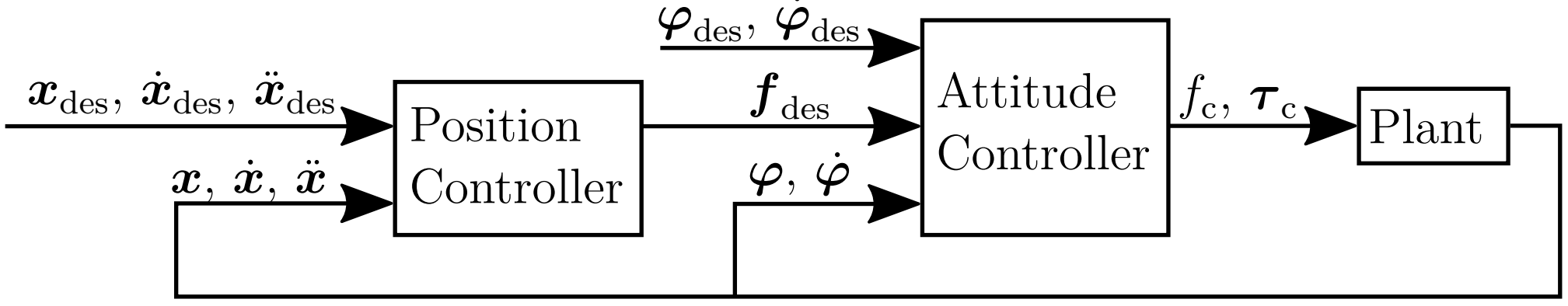

Figure 4.2: General control scheme of an aerial robot.

### 4.2.1 Position Control

As mentioned above, the position control is one of the two parts of the general control scheme of an aerial robot. Its task is to minimize the position error while ignoring the attitude dynamics. Based on that idea, the aerial robot is often seen as a point mass in $\mathbb{R}^3$. Therefore, the control inputs are the robot's position $\boldsymbol{x}$, velocity $\dot{\boldsymbol{x}}$, and acceleration $\ddot{\boldsymbol{x}}$, as well as the corresponding desired values $\boldsymbol{x}_\text{des}$, $\dot{\boldsymbol{x}}_{\text{des}}$, $\ddot{\boldsymbol{x}}_{\text{des}}$. The control output is, in accordance to Equations 4.4 and 4.5, the desired virtual thrust $\boldsymbol{f}_{\text{des}}$.

The simplest position control would be an Euler Angle controller, based on [LeeEtAl10a], which is similar to a PD-controller:

$$\boldsymbol{f}_{\text{des}} = \boldsymbol{k} \odot (\boldsymbol{x}_{\text{des}} - \boldsymbol{x}) + \boldsymbol{k}_{\text{v}} \odot (\dot{\boldsymbol{x}}_{\text{des}} - \dot{\boldsymbol{x}}) + m\ddot{\boldsymbol{x}}_{\text{des}} + m \cdot g\boldsymbol{e}_3, \tag{4.6}$$

where $\boldsymbol{k}, \boldsymbol{k}_{\text{v}} \in \mathbb{R}^3$ are the gains for the position and velocity error, respectively, and $\odot\colon \mathbb{R}^{n\times m} \to \mathbb{R}^{n\times m}$ denotes the Hadamard product defined as

$$(\boldsymbol{A} \odot \boldsymbol{B})_{ij} = (\boldsymbol{A})_{ij}(\boldsymbol{B})_{ij}, \quad \forall \quad \boldsymbol{A}, \boldsymbol{B} \in \mathbb{R}^{n\times m}. \tag{4.7}$$

But is augmented by the feed-forward terms $m\ddot{\boldsymbol{x}}_{\text{des}}$ and $m\cdot g\boldsymbol{e}_3$, adding the desired acceleration and compensating for the robot's mass. More advanced position controllers will be discussed in sections 4.3 and 4.4.

### 4.2.2 Attitude Control

Compared to the position control of an aerial robot, the attitude control is crucial for its stability. Without a stable attitude control, the aerial robot will not stay stable as a global system. The proof is as follows: assuming an aerial robot, whose position is tracking a reference without error but with an unstable or even non-existing attitude controller. This would reduce the free dynamics from $\mathsf{SE}(3)$ to $\mathsf{SO}(3)$. Now assuming the aerial robot could apply its thrust in any direction it desires, it could apply said thrust to always be positive along $\boldsymbol{e}_3$ and counteract any small disturbances in $\mathbb{R}^3$. Nonetheless, the thrust can only be applied along the aerial robot's positive $z$-axis. Even neglecting the $\mathbb{R}^3$dynamics, as soon as the aerial robot's positive $z$-axis has a positive part along $-\boldsymbol{e}_3$, it cannot counteract the gravitational pull, it would even accelerate towards the ground due to the thrust pointing downwards.

The inputs of the attitude control, as shown in Figure 4.2, are the desired thrust $\boldsymbol{f}_{\text{des}} \in \mathbb{R}^3$ from the position control, as well as $\boldsymbol{\varphi}$, $\dot{\boldsymbol{\varphi}}$, $\boldsymbol{\varphi}_{\text{des}}$, $\dot{\boldsymbol{\varphi}}_{\text{des}} \in \mathbb{R}^3$ denoting the robot's current Euler angles and rotation rates as well as their reference counterpart. From that, the attitude controller computes the command values, the scalar $f_{\text{c}} \in \mathbb{R}$ and the torque vector $\boldsymbol{\tau}_{\text{c}} \in \mathbb{R}^3$. For example, the proposed Euler Angle position controller has its attitude PID counterpart with:

$$\boldsymbol{\varphi}_{\text{des}} = \frac{1}{m \cdot g} \begin{bmatrix} \sin\boldsymbol{\varphi}_3 & -\cos\boldsymbol{\varphi}_3 & 0 \\ -\cos\boldsymbol{\varphi}_3 & \sin\boldsymbol{\varphi}_3 & 0 \\ 0 & 0 & 0 \end{bmatrix} \boldsymbol{f}, \tag{4.8}$$

$$f_{\text{c}} = \boldsymbol{f}_{\text{des}}^{\mathsf{T}}\boldsymbol{R}\boldsymbol{e}_3, \tag{4.9}$$

$$\boldsymbol{\tau}_{\text{c}} = \boldsymbol{k}_{\text{P}} \odot (\boldsymbol{\varphi}_{\text{des}} - \boldsymbol{\varphi}) + \boldsymbol{k}_{\text{I}} \odot \int_0^T (\boldsymbol{\varphi}_{\text{des}} - \boldsymbol{\varphi})\mathrm{d}t + \boldsymbol{k}_{\text{D}} \odot (\dot{\boldsymbol{\varphi}}_{\text{des}} - \dot{\boldsymbol{\varphi}}). \tag{4.10}$$

Here, $\boldsymbol{k}_\mathrm{P}, \boldsymbol{k}_\mathrm{I}, \boldsymbol{k}_\mathrm{D} \in \mathbb{R}^3$ are the corresponding PID gains.

This control scheme is sufficient for correcting small angular errors. Controlling huge angular errors, closer to 180°, requires the application of more sophisticated control strategies [WuEtAl16a, Mueller18]. Most of them differ from the way they compute the angular error. For example [Mueller18] discusses multiple different ways, which promise good recovery behavior for high angular errors. From the proposed algorithms, the following has been adapted, it is the same as used by [WuEtAl16a]:

$$\boldsymbol{e}_{\boldsymbol{R}} = \frac{1}{2}\left(\boldsymbol{R}_\mathrm{c}^\mathsf{T}\boldsymbol{R} - \boldsymbol{R}^\mathsf{T}\boldsymbol{R}_\mathrm{c}\right)^{\vee} \tag{4.11}$$

where $\boldsymbol{R}_\mathrm{c} \in \mathsf{SO}(3)$ is derived from the direction of $\boldsymbol{f}_\mathrm{des}$ and the desired rotation $\boldsymbol{R}_\mathrm{des} \in \mathsf{SO}(3)$ [WuEtAl16a].

## 4.3 Control Lyapunov Functions and Optimal Control

Suppose a given nonlinear system

$$\dot{\boldsymbol{x}} = f(\boldsymbol{x}) + g(\boldsymbol{x})\boldsymbol{u}, \quad \boldsymbol{x}(t_0) = \boldsymbol{x}_0, \tag{4.12}$$

with an equilibrium point at the origin, and $\boldsymbol{x} \in \mathbb{R}^n$, $\boldsymbol{u} \in \mathbb{R}^m$. Then, a positive definite, continuously differentiable function $V(\boldsymbol{x})$, $V\colon \mathbb{R}^n \to \mathbb{R}$ is called a Control Lyapunov Function (CLF) [AmesEtAl14b, WuEtAl16b] if and only if

$$c_1 \left\lVert \boldsymbol{x} \right\rVert_2^2 \leq V(\boldsymbol{x}) \leq c_2 \left\lVert \boldsymbol{x} \right\rVert_2^2 \tag{4.13}$$

and

$$\dot{V}(\boldsymbol{x}, \boldsymbol{u}) = \inf_{\boldsymbol{u} \in \mathbb{R}^m} \left(\mathcal{L}_f V + \mathcal{L}_g V \boldsymbol{u} + c_3 V\right) \leq 0, \tag{4.14}$$

with $c_1, c_2, c_3 > 0$ and $\mathcal{L}_f V, \mathcal{L}_g V$ denoting the Lie derivatives of $V$, defined as $\mathcal{L}_f V = \frac{\partial V}{\partial \boldsymbol{x}} f$ and $\mathcal{L}_g V = \frac{\partial V}{\partial \boldsymbol{x}} g$ [Hall15, RöbenackEtAl14]. Any Lyapunov candidate $V$ is exponentially stable, therefore guaranteeing stability for all $\boldsymbol{x} \in \mathbb{R}^n$.

Given the aforementioned properties, Control Lyapunov Functions are often used to guarantee stability and are especially useful for aerial robots. Since the attitude dynamics have to be stabilized for all $t \geq 0$ to guarantee general stability of the robot.

Determining a CLF, which satisfies the above conditions, is nontrivial. Moreover, different functions for the position and attitude control have to be defined. Based on [WuEtAl16a] two novice CLF candidates are proposed. In contrast to [WuEtAl16a] the candidates have improved tuning capabilities. For the position and attitude control the CLF candidates are

$$V_{\boldsymbol{x}} = \frac{1}{2} k_{\boldsymbol{x}} \boldsymbol{e}_{\boldsymbol{x}}^{\mathsf{T}} \boldsymbol{e}_{\boldsymbol{x}} + \frac{1}{2} m \boldsymbol{e}_{\boldsymbol{v}}^{\mathsf{T}} \boldsymbol{e}_{\boldsymbol{v}} + \boldsymbol{\varepsilon}_{\boldsymbol{x}} \boldsymbol{e}_{\boldsymbol{x}}^{\mathsf{T}} \boldsymbol{e}_{\boldsymbol{v}} \tag{4.15}$$

and

$$V_{\boldsymbol{R}} = \frac{1}{2 \min \lambda(\boldsymbol{J})} \boldsymbol{e}_{\dot{\varphi}}^{\mathsf{T}} \boldsymbol{K}_{\boldsymbol{J}} \boldsymbol{J} \boldsymbol{e}_{\dot{\varphi}} + \frac{1}{2} \boldsymbol{e}_{\boldsymbol{R}}^{\mathsf{T}} \boldsymbol{C}_{\boldsymbol{R}} \boldsymbol{e}_{\boldsymbol{R}} + (\boldsymbol{E}_{\boldsymbol{R}} \boldsymbol{e}_{\boldsymbol{R}})^{\mathsf{T}} \boldsymbol{e}_{\dot{\varphi}}, \tag{4.16}$$

respectively. Here, the individual errors are defined as $\boldsymbol{e}_{\boldsymbol{x}} = \boldsymbol{x} - \boldsymbol{x}_{\mathrm{des}}$, $\boldsymbol{e}_{\boldsymbol{v}} = \dot{\boldsymbol{x}} - \dot{\boldsymbol{x}}_{\mathrm{des}}$, $\boldsymbol{e}_{\boldsymbol{R}}$ as in Equation 4.11, and $\boldsymbol{e}_{\dot{\varphi}} = \dot{\boldsymbol{\varphi}} - \boldsymbol{R}^{\mathsf{T}} \boldsymbol{R}_{\mathrm{c}} \dot{\boldsymbol{\varphi}}_{\mathrm{des}}$. Furthermore, $\boldsymbol{\varepsilon}_{\boldsymbol{x}}, k_{\boldsymbol{J}} \in \mathbb{R}$ and $\boldsymbol{K}_{\boldsymbol{J}}, \boldsymbol{C}_{\boldsymbol{R}}, \boldsymbol{E}_{\boldsymbol{R}} \in \mathbb{R}^{3\times 3}$ are the tuning parameters.

## 4.3.1 Control Barrier Functions

Control Lyapunov Functions are not sufficient for obstacle avoidance, because they cannot incorporate obstacles into the candidate function. Assuming a free (non-)linear system of the form

$$\dot{\boldsymbol{x}} = f(\boldsymbol{x}), \tag{4.17}$$

with $\boldsymbol{x} \in \mathbb{R}^n$. Furthermore, assuming a set $\mathcal{C} \subset \mathbb{R}^n$ with the following properties:

$$\mathcal{C} = \{\boldsymbol{x} \in \mathbb{R}^n : h(\boldsymbol{x}) \geq 0\}, \tag{4.18}$$

$$\delta\mathcal{C} = \{\boldsymbol{x} \in \mathbb{R}^n : h(\boldsymbol{x}) = 0\}, \tag{4.19}$$

$$\mathrm{Int}(\mathcal{C}) = \{\boldsymbol{x} \in \mathbb{R}^n : h(\boldsymbol{x}) > 0\}, \tag{4.20}$$

where $\delta\mathcal{C}$ and $\mathrm{Int}(\mathcal{C})$ are the boundary and the interior of the set $\mathcal{C}$, respectively, and given a continuously differentiable function $h\colon \mathbb{R}^n \to \mathbb{R}$. Now the function $B\colon \mathcal{C} \subset \mathbb{R}^n \to \mathbb{R}$ is a barrier function [AmesEtAl14a, AmesEtAl19] if and only if there exist locally Lipschitz class $\mathcal{K}$ functions $\alpha_1, \alpha_2, \alpha_3$ [Khalil02], such that the following conditions hold true for all $\boldsymbol{x} \in \mathrm{Int}(\mathcal{C})$:

$$\inf_{\boldsymbol{x} \in \mathrm{Int}(\mathcal{C})} B(\boldsymbol{x}) \geq 0, \tag{4.21}$$

$$\lim_{\boldsymbol{x} \to \delta\mathcal{C}} B(\boldsymbol{x}) = \infty, \tag{4.22}$$

$$\frac{1}{\alpha_1(h(\boldsymbol{x}))} \leq B(\boldsymbol{x}) \leq \frac{1}{\alpha_2(h(\boldsymbol{x}))}, \tag{4.23}$$

$$\dot{B}(\boldsymbol{x}) \leq \alpha_3(h(\boldsymbol{x})). \tag{4.24}$$

A function $\alpha\colon [0, a) \to [0, \infty)$ is a Lipschitz class $\mathcal{K}$ function, if and only if $\alpha$ is strictly increasing and $\alpha(0) = 0$. Using $\alpha_1(z) = \alpha_2(z) = z$ and $\alpha_3(z) = \gamma z$ will lead to the limiting case of

$$B(\boldsymbol{x}) = \frac{1}{h(\boldsymbol{x})}, \tag{4.25}$$

$$\dot{B}(\boldsymbol{x}) = -\frac{\dot{h}(\boldsymbol{x})}{h^2(\boldsymbol{x})}. \tag{4.26}$$

Now, imposing

$$\dot{B}(\boldsymbol{x}) \leq \frac{\gamma}{B(\boldsymbol{x})}, \quad \forall \quad t > 0 \tag{4.27}$$

will guarantee a forward invariance of $\boldsymbol{x} \in \mathcal{C}$ if $\boldsymbol{x}_0 \in \mathcal{C}$.

Imposing such conditions for free systems is a nontrivial task. Thus, assuming the following control system:

$$\dot{\boldsymbol{x}} = f(\boldsymbol{x}) + g(\boldsymbol{x})\boldsymbol{u} \tag{4.28}$$

with $\boldsymbol{x} \in \mathbb{R}^n$, $\boldsymbol{x}_0 \in \mathcal{C}$ and $\boldsymbol{u} \in \mathcal{U} \subset \mathbb{R}^m$. Now, modifying Equations 4.23 and 4.24 will lead to the new definition of $B$. Which is, $B\colon \mathcal{C} \to \mathbb{R}$ is a Control Barrier Function (CBF), if and only if there exist locally Lipschitz class $\mathcal{K}$ functions $\alpha_1, \alpha_2$, with $\gamma > 0$, such that, for all $\boldsymbol{x} \in \mathrm{Int}(\mathcal{C})$,

$$\frac{1}{\alpha_1(\|\boldsymbol{x}\|_{\delta\mathcal{C}})} \leq B(\boldsymbol{x}) \leq \frac{1}{\alpha_2(\|\boldsymbol{x}\|_{\delta\mathcal{C}})}, \tag{4.29}$$

$$\inf_{\boldsymbol{u}\in\mathcal{U}} \left( \mathcal{L}_f B(\boldsymbol{x}) + \mathcal{L}_g B(\boldsymbol{x})\boldsymbol{u} - \frac{\gamma}{B(\boldsymbol{x})} \right) \leq 0. \tag{4.30}$$

Finally, it can be proven [WuEtAl16a], that any Lipschitz continuous controller

$$\left\{ \boldsymbol{u}(\boldsymbol{x}, t) \in \mathcal{U}\colon \mathcal{L}_f B(\boldsymbol{x}(t)) + \mathcal{L}_g B(\boldsymbol{x}(t))\boldsymbol{u}(\boldsymbol{x}, t) - \frac{\gamma}{B(\boldsymbol{x}(t))} \leq 0 \right\}, \quad \forall \quad t > 0 \tag{4.31}$$

will result in $\mathcal{C}$ being forward invariant as long as there is a feasible controller for all $t > 0$. In other words, enforcing the condition in Equation 4.30, will guarantee forward invariance, and therefore safety, as $\boldsymbol{x}(t)$ will stay inside the set $\mathcal{C}$ for all $t > 0$, given the initial condition of $\boldsymbol{x}(0) \in \mathcal{C}$.

The safety concern to be guaranteed is, that the aerial robot stays outside of the obstacle at all time. Here, the obstacles are represented by spheres. Any obstacle that does not resemble a sphere can be represented by a finite number of smaller spheres. A greater number of spheres will result in a more precise approximation of the non-spherical object. However, a greater number of spheres will cause an increased complexity and computational effort. The aerial robot itself is represented by a single sphere. Now, the distance between the center of two spheres is $\|\boldsymbol{x}_{\text{Sphere1}} - \boldsymbol{x}_{\text{Sphere2}}\|_2$. This distance has to be greater than the radii of the spheres, otherwise, they overlap. Therefore,

$$\left\|\boldsymbol{x} - \boldsymbol{x}_{i,\text{o}}\right\|_2 \geq \tilde{r}_i, \tag{4.32}$$

$$\Leftrightarrow \quad \left\|\boldsymbol{x} - \boldsymbol{x}_{i,\text{o}}\right\|_2^2 \geq \tilde{r}_i^2, \tag{4.33}$$

$$\Leftrightarrow \left(\boldsymbol{x} - \boldsymbol{x}_{i,\text{o}}\right)^{\mathsf{T}} \left(\boldsymbol{x} - \boldsymbol{x}_{i,\text{o}}\right) \geq \tilde{r}_i^2, \tag{4.34}$$

is the equivalent constraint. Here $\boldsymbol{x} \in \mathbb{R}^3$ is the robots position, $\boldsymbol{x}_{i,\text{o}} \in \mathbb{R}^3$ is the position of the $i$-th obstacle, more precisely sphere, and $\tilde{r} = r + r_{i,\text{o}} + r_{\text{safety}} \in \mathbb{R}_+$ is the augmented distance between the centers with $r$ as the radius of the sphere around the robot, $r_{i,\text{o}}$ the radius of the sphere around the $i$-th obstacle, and $r_{\text{safety}}$ an additional safety distance.

From that, [WuEtAl16a] derive the constraint function $h_i\colon \mathsf{TSE}(3) \to \mathbb{R}$ for the Control Barrier Function as

$$\hat{h} \coloneqq \gamma\alpha(\hat{g}_i) + \dot{\hat{g}}_i, \tag{4.35}$$

$$= \gamma\alpha\left((\boldsymbol{x} - \boldsymbol{x}_{i,\text{o}})^{\mathsf{T}}(\boldsymbol{x} - \boldsymbol{x}_{i,\text{o}})\right) + 2(\dot{\boldsymbol{x}} - \dot{\boldsymbol{x}}_{i,\text{o}})^{\mathsf{T}}(\boldsymbol{x} - \boldsymbol{x}_{i,\text{o}}) - \beta\dot{b} - \sigma'(s_i)\dot{s}_i, \tag{4.36}$$

where

$$s_i = (\boldsymbol{x} - \boldsymbol{x}_{i,\text{o}})\boldsymbol{R}\boldsymbol{e}_3, \tag{4.37}$$

$$\dot{s}_i = (\dot{\boldsymbol{x}} - \dot{\boldsymbol{x}}_{i,\text{o}})\boldsymbol{R}\boldsymbol{e}_3 + (\boldsymbol{x}_{i,\text{o}} - \boldsymbol{x})\boldsymbol{R}(\dot{\boldsymbol{\varphi}} \times \boldsymbol{e}_3), \tag{4.38}$$

and $\sigma\colon \mathbb{R} \to \mathbb{R}$ is a smooth scalar function satisfying the following conditions

$$\sigma'(s_i) < 0, \tag{4.39}$$

$$|\sigma(s_i)| < \sigma(0) < (\beta - 1)b, \quad \forall \quad s_i \in \mathbb{R}, \tag{4.40}$$

$$2s - \sigma'(s_i) > 0, \quad \forall \quad s_i \in (-\sqrt{\beta}b, 0). \tag{4.41}$$

### 4.3.2 Optimal Control and Quadratic Programming

Enforcing Equation 4.14 on both Equations 4.15 and 4.16, for all $\boldsymbol{u}(t), t > 0$ will guarantee said stability and safety. The constraint can be enforced using a quadratic program (QP) [NocedalEtAl06]. This control scheme is also often referred to as optimal control, as it calculates the optimal input by minimizing the cost function, while enforcing all constraints. Based on the previous calculations, the two quadratic programs can be obtained as

$$\boldsymbol{f}^*_{\text{des}} = \underset{\boldsymbol{f} \in \mathbb{R}^3}{\arg\min} \; \frac{1}{2}\boldsymbol{f}^{\mathsf{T}}\boldsymbol{Q}_{\boldsymbol{x}}\boldsymbol{f} + \boldsymbol{c}_{\boldsymbol{x}}^{\mathsf{T}}\boldsymbol{f}, \tag{4.42}$$

$$\text{subj. to} \quad \dot{V}_{\boldsymbol{x}}(\boldsymbol{f}) + \eta_{\boldsymbol{x}} V_{\boldsymbol{x}} \leq 0 \tag{4.43}$$

and

$$[f_\mathrm{c}^*,\ \boldsymbol{\tau}_\mathrm{c}^*] = \underset{f_\mathrm{c},\delta\in\mathbb{R},\ \boldsymbol{\tau}_\mathrm{c}\in\mathbb{R}^3}{\arg\min}\ \frac{1}{2}q_{\boldsymbol{R},1}\left(f_\mathrm{c}-f_\mathrm{des}\right)^2+\frac{1}{2}\boldsymbol{\tau}_\mathrm{c}^\top\boldsymbol{Q}_{\boldsymbol{R}}^*\boldsymbol{\tau}_\mathrm{c}+\frac{1}{2}q_{\boldsymbol{R},5}\delta^2, \tag{4.44}$$

$$\text{subj. to}\quad \dot{V}_{\boldsymbol{R}}(\boldsymbol{\tau}_\mathrm{c})+\eta_{\boldsymbol{R}}V_{\boldsymbol{R}}\leq\delta, \tag{4.45}$$

$$\dot{B}_j(f_\mathrm{c},\boldsymbol{\tau}_\mathrm{c})\leq\frac{\gamma_j}{B_j},\ j\in\mathcal{I}_s(t). \tag{4.46}$$

The slack variable $\delta$ is introduced, such that the CLF constraint is softened and the controller can violate it in favor of the Control Barrier Function, thus the aerial robot's states stay inside $\mathcal{B}_t$. Moreover, $\boldsymbol{Q}_{\boldsymbol{x}} = \mathrm{diag}(q_{\boldsymbol{x},1}, q_{\boldsymbol{x},2}, q_{\boldsymbol{x},3}) \in \mathbb{R}^{3\times3}$ with $q_{\boldsymbol{x},1}, q_{\boldsymbol{x},2}, q_{\boldsymbol{x},3} > 0$ and $\boldsymbol{Q}_{\boldsymbol{R}}^* = \mathrm{diag}(q_{\boldsymbol{R},2}, q_{\boldsymbol{R},3}, q_{\boldsymbol{R},4}) \in \mathbb{R}^{3\times3}$ and $\boldsymbol{Q}_{\boldsymbol{R}} = \mathrm{diag}(q_{\boldsymbol{R},1}, \boldsymbol{Q}_{\boldsymbol{R}}^*, q_{\boldsymbol{R},5}) \in \mathbb{R}^{5\times5}$ with $q_{\boldsymbol{R},1}, \ldots, q_{\boldsymbol{R},5} > 0$ are the tuning parameters for the quadratic problem. Furthermore, $\eta_{\boldsymbol{x}}, \eta_{\boldsymbol{R}}, \gamma_j \in \mathbb{R}$ are the gain parameters for the constraints, where $\mathcal{I}_s(t)$ denotes the time-varying set of indices, each corresponding to one sphere in particular.

## 4.4 Model Predictive Control

Model Predictive Control [KwonEtAl05] is based on the idea to generate an optimal set of inputs based on a dynamical model and its predicted evolution based on said inputs. While both CLF-CBF-QP and MPC rely on the dynamical system, only the Model Predictive Controller uses the model to predict all following states and optimizing all following inputs. Therefore, it can consider the long-term development of the dynamical model's states, which is advantageous for more effective obstacle avoidance. Now if instead of an infinite time horizon, the time horizon is limited, the states will only be propagated until said horizon is reached.

### 4.4.1 General Notation and Derivation of the Problem Description

Given a dynamical system

$$\dot{\boldsymbol{x}}(t) = f(\boldsymbol{x}(t)) + g(\boldsymbol{x}(t))\boldsymbol{u}(t), \tag{4.47}$$

and its discretization

$$\boldsymbol{x}(t+T_\mathrm{s}) = f_\mathrm{d}(\boldsymbol{x}(t)) + g_\mathrm{d}(\boldsymbol{x}(t))\boldsymbol{u}(t), \tag{4.48}$$

where $T_\mathrm{s}$ is the sampling time. Thus, $t + T_\mathrm{s} \propto k + 1$, and Equation 4.48 can now be written as

$$\boldsymbol{x}_{k+1} = f_\mathrm{d}(\boldsymbol{x}_k) + g_\mathrm{d}(\boldsymbol{x}_k)\boldsymbol{u}_k. \tag{4.49}$$

Here, it is important to distinguish between the measured and predicted states and outputs of the system. The general notation for the predicted state $\boldsymbol{x}_{k+1}$, predicted at time $t$, is $\boldsymbol{x}_{k+1|t}$, based on the measured input state $\boldsymbol{x}_t = \boldsymbol{x}_{t|t}$ and the dynamical model from Equation 4.49. Likewise, $\boldsymbol{u}_{k|t}$ is the control input at time $k$ predicted at time $t$.

A cost function is applied to the optimization, evaluating the controller's performance based on the current and predicted states and outputs. In general, the cost function is defined as

$$J^*_{0\to\infty}(\boldsymbol{x}_0) = \min_{\boldsymbol{u}_0,\dots,\boldsymbol{u}_\infty} \sum_{k=0}^{\infty} h(\boldsymbol{x}_{k|0}, \boldsymbol{u}_{k|0}), \tag{4.50}$$

where $J^*_{0\to\infty}(\boldsymbol{x}_0)$ is the cost function. Given an infinite time horizon, it is evaluated from $t = 0$ to $t \to \infty$. Furthermore, $\boldsymbol{u}_0, \dots, \boldsymbol{u}_\infty$ are the control inputs at each time step, and $h(\boldsymbol{x}_k, \boldsymbol{u}_k)$ in the stage cost for all $k \geq 0$. Therefore, the optimization problem is defined as:

$$J^*_{0\to\infty}(\boldsymbol{x}_0) = \min_{\boldsymbol{u}_{0|0},\dots,\boldsymbol{u}_{\infty|t}} \sum_{k=0}^{\infty} h(\boldsymbol{x}_{k|0}, \boldsymbol{u}_{k|0}), \tag{4.51}$$

$$\text{subj. to}\, \boldsymbol{x}_{k+1|0} = f_\mathrm{d}(\boldsymbol{x}_{k|0}) + g_\mathrm{d}(\boldsymbol{x}_{k|0})\boldsymbol{u}_{k|0}, \quad \forall \quad k \geq 0, \tag{4.52}$$

$$\boldsymbol{x}_{0|0} = \boldsymbol{x}(0), \tag{4.53}$$

$$\boldsymbol{x}_{k|0} \in \mathcal{X}, \quad \forall \quad k \geq 0, \tag{4.54}$$

$$\boldsymbol{u}_{k|0} \in \mathcal{U}, \quad \forall \quad k \geq 0. \tag{4.55}$$

Here, $\mathcal{X} \subseteq \mathbb{R}^{n_x}$ and $\mathcal{U} \subseteq \mathbb{R}^{n_u}$ are the sets of possible states and inputs for a system with $n_x$ states and $n_u$ inputs.

Solving this optimization problem up to $t = \infty$ is only possible in very specific cases and mostly not by numerical operations. Therefore, a limited time horizon $N$, which equals a limited number of time steps for the optimization, is introduced.

With this approach, the optimizer solves the MPC problem for $t, \dots, t + N$, applies the first control input $\boldsymbol{u}_{t|t}$, and starts over, now for $t + T_\mathrm{s}, \dots, t + N + T_\mathrm{s}$. Hence, the new cost function is defined as

$$J^*_{t\to t+N}(\boldsymbol{x}_t) = \min_{\boldsymbol{u}_{t|t},\dots,\boldsymbol{u}_{t+N|t}} \sum_{k=t}^{t+N} h(\boldsymbol{x}_{k|t}, \boldsymbol{u}_{k|t}) + h_\mathrm{f}(\boldsymbol{x}_{t+N+1|t}), \tag{4.56}$$

leading to the Model Predictive Controller as

$$J^*_{t\to t+N}(\boldsymbol{x}_t) = \min_{\boldsymbol{u}_{t|t},\dots,\boldsymbol{u}_{t+N|t}} \sum_{k=t}^{t+N} h(\boldsymbol{x}_{k|t}, \boldsymbol{u}_{k|t}) + h_{\mathrm{f}}(\boldsymbol{x}_{t+N+1|t}), \tag{4.57}$$

$$\text{subj. to}\, \boldsymbol{x}_{k+1|t} = f_{\mathrm{d}}(\boldsymbol{x}_{k|t}) + g_{\mathrm{d}}(\boldsymbol{x}_{k|t})\boldsymbol{u}_{k|t}, \quad \forall \quad k = \{t, \dots, t+N\}, \tag{4.58}$$

$$\boldsymbol{x}_t = \boldsymbol{x}_{t|t}, \tag{4.59}$$

$$\boldsymbol{x}_{k|t} \in \mathcal{X}, \quad \forall \quad k = \{t, \dots, t+N\}, \tag{4.60}$$

$$\boldsymbol{u}_{k|t} \in \mathcal{U}, \quad \forall \quad k = \{t, \dots, t+N\}, \tag{4.61}$$

$$\boldsymbol{x}_{t+N+1|t} \in \mathcal{X}_{\mathrm{f}}. \tag{4.62}$$

Here, $h_{\mathrm{f}}(\boldsymbol{x}_{t+N+1|t})$ is the terminal cost and $\mathcal{X}_{\mathrm{f}}$ the terminal set.

Based on this standard design, the stage cost has to be defined, as well as all constraints needed to adapt the controller to the given problem of obstacle avoidance.

### 4.4.2 Model Predictive Control for Aerial Robots with Obstacle Avoidance

Similar to optimal control, Model Predictive Control is used to calculate the optimal (next) control input, to minimize the overall error between the states and the global reference trajectory, while enforcing all given constraints. While optimal control only takes into account the next step, MPC optimizes over a (finite) time horizon. Optimal control is identical to MPC, if the prediction- and control-horizon are determined for only one time-step. The usefulness of MPC arrives from the fact that it is using an underlying dynamical model of the system to predict the state of the next time step. Based on this information, it can calculate the next control input and therefore the systems movement one step further. But even with the best dynamical model, there will always be a model mismatch between the dynamical model of the controller and the real world system. More importantly, the further the controller propagates the states, the bigger the error gets. However, for each iteration of the controller, a new state sequence will be calculated. Using the new real state information at each start, the error always starts at zero. Using this technique, the controller could predict the complete movement of the system until $t \to \infty$.

This controller is especially useful for obstacle avoidance, as it also takes into account the obstacle's movement. Thus, the MPC can find the optimal trajectory to avoid the obstacles even more efficient than one step solvers. To do so, the MPC only takes the $\mathbb{R}^3$ dynamics into account. Using the full $\mathsf{SE}(3)$ dynamics would just hinder the avoidance performance for the following reasons. First, the size of the optimization problem is proportional to the number on the states and outputs of the system as well as the prediction horizon. Including the $\mathsf{SO}(3)$ dynamics would double the number of states and therefore quadrupling the size of the optimization matrix, without taking into account the change in control inputs. This could potentially also quadruple the solving time. Second, the necessary sampling time would have to increase by a factor of 50. While the positional control loop can generally be run at around $f_{\text{pos}} = 10\,\text{Hz}$, the attitude control loop has to be run with at least $f_{\text{att}} = 250\,\text{Hz}$, better yet, $f_{\text{att}} = 500\,\text{Hz}$. Keeping the prediction horizon the same with regards to the number of steps would therefore reduce the prediction time by a factor of 50. Long term prediction for obstacle avoidance would not be possible. A possible solution could be to create a lookup table for different combinations of states offline [WangEtAl10]. Nevertheless, with the huge combinatorial space especially for continuous states and the varying set of Control Barrier Function constraints, this might not be feasible.

For obstacle avoidance, a constraint will be introduced in subsection 4.4.4, to enforce that the distance between the aerial robot and the obstacles is greater than a safety distance. To enforce this constraint, Control Barrier Functions have been proven to be an excellent choice. Combining an MPC approach with CBFs to enforce a custom constraint is a novel approach. Previous work [WillsEtAl04, WillsEtAl05] only used general barrier functions, which lack the advantages of Control Barrier Functions such as their Lyapunov-like properties, see subsection 4.3.1.

### 4.4.3 Cost Function and Stage Cost

For MPCs with limited prediction horizon, the stage cost is often split into a stage cost for $k = t, \ldots, t+N$ as well as a final stage cost for $k = t+N+1$. The latter results from the last evolution of the dynamical system, given $\boldsymbol{x}_{t+N|t}$ and $\boldsymbol{u}_{t+N|t}$. The stage cost function $h(\boldsymbol{x}_{k|t}, \boldsymbol{u}_{k|t})$ is thus defined as

$$h(\boldsymbol{x}_{k|t}, \boldsymbol{u}_{k|t}) = q(\boldsymbol{x}_{k|t}, \boldsymbol{u}_{k|t}) + p(\boldsymbol{x}_{t+N+1|t}), \quad \forall \quad k = \{t, \ldots, t+N\}, \tag{4.63}$$

where $q(\boldsymbol{x}_{k|t}, \boldsymbol{u}_{k|t})$ is the new stage cost function and $p(\boldsymbol{x}_{t+N+1|t})$ the final stage cost function. Both are often defined as

$$q(\boldsymbol{x}_{k|t}, \boldsymbol{u}_{k|t}) = \boldsymbol{x}_{k|t}^\top \boldsymbol{Q} \boldsymbol{x}_{k|t} + \boldsymbol{u}_{k|t}^\top \boldsymbol{R} \boldsymbol{u}_{k|t}, \quad \forall \quad k = \{t, \ldots, t+N\}, \tag{4.64}$$

$$p(\boldsymbol{x}_{t+N+1|t}) = \boldsymbol{x}_{t+N+1|t}^\top \boldsymbol{Q}_{\text{final}} \boldsymbol{x}_{t+N+1|t} \tag{4.65}$$

where $\boldsymbol{Q}, \boldsymbol{Q}_{\text{final}} \in \mathbb{S}_+^{n_x \times n_x}$, $\boldsymbol{R} \in \mathbb{S}_+^{n_u \times n_u}$ are all positive semi-definite matrices, denoting the stage cost, final stage cost, and input cost respectively.

The overall goal of the aerial robot is not to thrive towards all states equaling zero but to track its desired trajectory. All that while keeping the inputs as low as possible. Thus, $q(\boldsymbol{x}_{k|t}, \boldsymbol{u}_{k|t})$ is augmented to be

$$q(\boldsymbol{x}_{k|t}, \boldsymbol{u}_{k|t}) = (\boldsymbol{x}_{k|t} - \boldsymbol{x}_{\text{ref},k})^{\mathsf{T}} \boldsymbol{Q} (\boldsymbol{x}_{k|t} - \boldsymbol{x}_{\text{ref},k}) + \boldsymbol{u}_{k|t}^{\mathsf{T}} \boldsymbol{R} \boldsymbol{u}_{k|t}, \quad \forall \quad k = \{t, \dots, t+N\}. \tag{4.66}$$

Here, $\boldsymbol{x}_{\text{ref},k}$ is the reference state vector at time $t = k$.

Therefore, the full cost function is

$$\begin{aligned} J^*_{t\rightarrow t+N}(\boldsymbol{x}_t) = \min_{\boldsymbol{u}_{t|t},\dots,\boldsymbol{u}_{t+N|t}} \sum_{k=t}^{t+N} & (\boldsymbol{x}_{k|t} - \boldsymbol{x}_{\text{ref},k})^{\mathsf{T}} \boldsymbol{Q} (\boldsymbol{x}_{k|t} - \boldsymbol{x}_{\text{ref},k}) + \boldsymbol{u}_{k|t}^{\mathsf{T}} \boldsymbol{R} \boldsymbol{u}_{k|t} \\ & + \boldsymbol{x}_{t+N+1|t}^{\mathsf{T}} \boldsymbol{Q}_{\text{final}} \boldsymbol{x}_{t+N+1|t}. \end{aligned} \tag{4.67}$$

### 4.4.4 Constraints

In the following, the constraints of the Model Predictive Controller are listed and discussed, as well as derived where necessary.

**Dynamical Model**

The discrete dynamical model for the Model Predictive Control is based on the model defined in Equation 4.1. However, $f\boldsymbol{R}\boldsymbol{e}_3$ is replaced by the desired thrust $\boldsymbol{f}_{\text{des}}$, as this is the output of the position control and thus input of the attitude control. It can therefore be written as:

$$m\ddot{\boldsymbol{x}} = \boldsymbol{f}_{\text{des}} - m \cdot g\boldsymbol{e}_3. \tag{4.68}$$

Furthermore, the model can also be written as a state-space model as:

$$\dot{\tilde{\boldsymbol{x}}} = \boldsymbol{A}(\tilde{\boldsymbol{x}})\tilde{\boldsymbol{x}} + \boldsymbol{B}(\tilde{\boldsymbol{x}})\boldsymbol{u}, \tag{4.69}$$

$$\Rightarrow \begin{bmatrix} \dot{\boldsymbol{x}} \\ \ddot{\boldsymbol{x}} \end{bmatrix} = \begin{bmatrix} \boldsymbol{I}_3 & \boldsymbol{0}_{3\times3} \\ \boldsymbol{0}_{3\times3} & \boldsymbol{0}_{3\times3} \end{bmatrix} \begin{bmatrix} \boldsymbol{x} \\ \dot{\boldsymbol{x}} \end{bmatrix} + \begin{bmatrix} \boldsymbol{0}_{3\times3} & \boldsymbol{0}_{3\times1} \\ \frac{1}{m}\boldsymbol{I}_3 & -g\boldsymbol{e}_3 \end{bmatrix} \begin{bmatrix} \boldsymbol{f}_{\text{des}} \\ 1 \end{bmatrix}, \tag{4.70}$$

where $\boldsymbol{A} \in \mathbb{R}^{n_x \times n_x}$ is the state matrix with the state vector $\tilde{\boldsymbol{x}} \in \mathbb{R}^{n_x}$ and $\boldsymbol{B} \in \mathbb{R}^{n_x \times n_u}$ is the input matrix with the input vector $\boldsymbol{u} \in \mathbb{R}^{n_u}$. Furthermore, $\boldsymbol{I}_n \in \mathbb{R}^{n\times n}$ is the $n$-dimensional identity matrix and $\boldsymbol{0}_{n\times m} \in \mathbb{R}^{n\times m}$ is the zero matrix of size $n \times m$. Here it is assumed, that the output vector is fully measurably and equals the state vector. This assumption is valid, as the states are the position and velocity of the aerial robot, which are fully measurable, for example using the setup described in subsection 3.2.2.

To discretize the continuous state space model, exact discretization via zero-order hold is used. The result is

$$\boldsymbol{x}_{k+1|t} = \boldsymbol{\Phi}(\boldsymbol{x}_{k|t}, T_{\text{s}})\boldsymbol{x}_{k|t} + \boldsymbol{\Gamma}(\boldsymbol{x}_{k|t}, T_{\text{s}})\boldsymbol{u}_{k|t}, \tag{4.71}$$

where

$$\mathbf{\Phi}(\boldsymbol{x}_{k|t}, T_\mathrm{s}) = \mathrm{e}^{\boldsymbol{A}T_\mathrm{s}} \in \mathbb{R}^{n_x \times n_x}, \tag{4.72}$$

$$\mathbf{\Gamma}(\boldsymbol{x}_{k|t}, T_\mathrm{s}) = \int_0^{T_\mathrm{s}} \mathrm{e}^{\boldsymbol{A}t}\boldsymbol{B}\mathrm{d}t \in \mathbb{R}^{n_x \times n_u} \tag{4.73}$$

are the discrete state matrix and input matrix respectively.

Summarizing, the dynamics constraint is defined as

$$\boldsymbol{x}_{k+1|t} = \mathbf{\Phi}(\boldsymbol{x}_{k|t}, T_\mathrm{s})\boldsymbol{x}_{k|t} + \mathbf{\Gamma}(\boldsymbol{x}_{k|t}, T_\mathrm{s})\boldsymbol{u}_{k|t}, \quad \forall \quad k = \{t, \dots, t+N\}. \tag{4.74}$$

Given Equation 4.70 and a constant sampling time $T_\mathrm{s}$, the equation can be simplified as

$$\boldsymbol{x}_{k+1|t} = \mathbf{\Phi}\boldsymbol{x}_{k|t} + \mathbf{\Gamma}\boldsymbol{u}_{k|t}, \quad \forall \quad k = \{t, \dots, t+N\}. \tag{4.75}$$

With a sampling time of $T_\mathrm{s} = 0.1\,\mathrm{s}$, the discrete state space model would equal

$$\boldsymbol{x}_{k+1|t} = \begin{bmatrix} \boldsymbol{I}_3 & 0.1\boldsymbol{I}_3 \\ \boldsymbol{0}_{3\times3} & \boldsymbol{I}_3 \end{bmatrix} \boldsymbol{x}_{k|t} + \begin{bmatrix} \frac{g}{2\,000}\boldsymbol{I}_3 & -\frac{g}{200}\boldsymbol{e}_3 \\ \frac{g}{100}\boldsymbol{I}_3 & -\frac{g}{10}\boldsymbol{e}_3 \end{bmatrix} \boldsymbol{u}_{k|t}. \tag{4.76}$$

**Input Constraints**

In theory, the output of the position controller, and thus the input of the attitude controller do not have to be bounded. However, for the Model Predictive Controller to predict a feasible trajectory, it has to be aware of the physical limit of the aerial robot's actuation system. Therefore, it shall hold true, that

$$\|\boldsymbol{u}_{k|t}\|_2 \leq f_\mathrm{max}, \quad \forall \quad k \in \{t, \dots, t+N\}, \tag{4.77}$$

$$\Leftrightarrow \boldsymbol{u}_{k|t}^\mathsf{T}\boldsymbol{u}_{k|t} \leq f_\mathrm{max}^2, \quad \forall \quad k \in \{t, \dots, t+N\}. \tag{4.78}$$

Now the physical system proposes another limitation that should be considered via another constraint. The real system cannot reverse its thrust, in other words

$$f_\mathrm{c} \geq 0. \tag{4.79}$$

Now obviously the dynamics predicted by the Model Predictive Controller are in $\mathbb{R}^3$, thus, no attitude information is available. Hence, the direction of thrust could be positive with $-z$. Nonetheless, a “flipped” aerial robot is not stable, also, the desired rotations $\boldsymbol{\varphi}_\mathrm{des}^{(1)}$ and $\boldsymbol{\varphi}_\mathrm{des}^{(2)}$ are almost always close to 0, where $\boldsymbol{\varphi}_\mathrm{des}^{(3)}$ has no effect on the direction of thrust. Combining the given information, it shall also be imposed, that

$$\boldsymbol{u}_{k|t}^{(3)} \geq 0, \quad \forall \quad k \in \{t, \dots, t+N\}. \tag{4.80}$$

Optimizing the inputs for the full prediction horizon $N$ is not only computational expensive but also often unnecessary. Ever so often, it is sufficient to only calculate the inputs for the first part of the prediction horizon, and holding the last calculated input until the end of the prediction horizon. This introduces the control horizon $0 < N_\mathrm{c} \leq N$. Therefore, the input constraints can be written as:

$$\boldsymbol{u}_{k|t}^\mathsf{T}\boldsymbol{u}_{k|t} \leq f_\mathrm{max}^2, \qquad \forall \quad k \in \{t, \dots, t+N_\mathrm{c}\}, \tag{4.81}$$

$$\boldsymbol{u}_{k|t}^{(3)} \geq 0, \qquad \forall \quad k \in \{t, \dots, t+N_\mathrm{c}\}, \tag{4.82}$$

$$\boldsymbol{u}_{k|t} = \boldsymbol{u}_{t+N_\mathrm{c}|t} \qquad \forall \quad k \in \{t+N_\mathrm{c}, \dots, t+N\}. \tag{4.83}$$

To further simplify and linearize the constraint from Equation 4.81, it can be assumed, that each component of $\boldsymbol{u}_{k|t}$ is smaller than $f_\mathrm{max}$. It will be later shown, that for most cases, $\boldsymbol{u}_{k|t}^{(1)}$ and $\boldsymbol{u}_{k|t}^{(2)}$ will actually stay within approximately $-10\,\mathrm{N}$ and $10\,\mathrm{N}$. The linearized variant of Equation 4.81 would be:

$$|\boldsymbol{u}_{k|t}^{(1)}| \leq f_\mathrm{max}, \qquad \forall \quad k \in \{t, \dots, t+N_\mathrm{c}\}, \tag{4.84}$$

$$|\boldsymbol{u}_{k|t}^{(2)}| \leq f_\mathrm{max}, \qquad \forall \quad k \in \{t, \dots, t+N_\mathrm{c}\}, \tag{4.85}$$

$$0 \leq |\boldsymbol{u}_{k|t}^{(3)}| \leq f_\mathrm{max}, \qquad \forall \quad k \in \{t, \dots, t+N_\mathrm{c}\}. \tag{4.86}$$

Lastly, it is obvious from Equation 4.70, that $\boldsymbol{u}_{k|t}^{(4)}$ has to be enforced to be 1 at all times, as it is representing the constant gravitational pull along $e_3$. Thus,

$$\boldsymbol{u}_{k|t}^{(4)} = 1, \qquad \forall \quad k \in \{t, \dots, t+N_\mathrm{c}\}. \tag{4.87}$$

**Input's Rate of Change and Slew Rate**

The actuation system of the aerial robot is not only limited by its maximum thrust but also by the maximum change of thrust. While the time constant of the motor is smaller than the sampling frequency of the Model Predictive Controller, it is still advisable to limit the rate of change. For one, high fluctuations in the thrust will drain the battery faster, due to current peaks. And for another, high fluctuations could also potentially introduce instabilities due to unmodeled effects.

The rate of change can be limited by potentially two ways. Either by a constraint, or by introducing another term in the cost function. Using a constraint will guarantee a maximum rate of change, however, it still allows for fluctuations inside those limits. Otherwise, adding a term to the cost function to minimize the change, will keep it as low as possible, while still allowing high rates of change in cases where necessary. Therefore, the latter was chosen.

The term, which will be added to the cost function is:

$$\sum_{k=t}^{t+N-1} (\boldsymbol{u}_{k|t} - \boldsymbol{u}_{k+1|t})^\mathsf{T} \boldsymbol{R}_\Delta (\boldsymbol{u}_{k|t} - \boldsymbol{u}_{k+1|t}), \tag{4.88}$$

where $\boldsymbol{R}_{\Delta} \in \mathbb{S}_{+}^{n_u \times n_u}$ is the corresponding positive semi-definite weight matrix. By incorporating Equation 4.83, this is equivalent to:

$$\sum_{k=t}^{t+N_c-1} (\boldsymbol{u}_{k|t} - \boldsymbol{u}_{k+1|t})^{\mathsf{T}} \boldsymbol{R}_{\Delta} (\boldsymbol{u}_{k|t} - \boldsymbol{u}_{k+1|t}). \tag{4.89}$$

**Exponential Stability**

To ensure exponential stability, the Control Lyapunov Function from Equation 4.15 can be modified. Also introducing the slack parameter $\delta_k$ to allow the violation of the stability constraint in favor of the obstacle avoidance constraint. Thus,

$$\dot{V}_{\boldsymbol{x}}(\boldsymbol{x}_{k+1|t}, \boldsymbol{u}_{k|t}) + \eta_{\boldsymbol{x}} V_{\boldsymbol{x}}(\boldsymbol{x}_{k+1|t}) \leq \delta_k, \quad \forall \quad k \in \{t, \ldots, t+N\}, \tag{4.90}$$

which is equivalent to

$$\mathcal{L}_f V_{\boldsymbol{x}}(\boldsymbol{x}_{k+1|t}) + \mathcal{L}_g V_{\boldsymbol{x}}(\boldsymbol{x}_{k+1|t}) \boldsymbol{u}_{k|t} + \eta_{\boldsymbol{x}} V_{\boldsymbol{x}}(\boldsymbol{x}_{k+1|t}) \leq \delta_k, \quad \forall \quad k \in \{t, \ldots, t+N\}. \tag{4.91}$$

Now given that the variable $\delta_k$ itself shall be as small as possible, thus reducing the deviation from the stability constraint, $\delta_k$ has to be added to the cost function using the term

$$\sum_{k=t}^{t+N} q \delta_k^2, \tag{4.92}$$

where the influence of $\delta_k$ on the optimization can be controlled via the tuning parameter $q$.

Ensuring exponential stability only for the next time step, as done by the CBF-CLF-QP [WuEtAl16a], is enough to ensure globally exponential stability. Nevertheless, by ensuring exponential stability for all predicted $\boldsymbol{x}_{t+1 \to t+N+1|t}$, the prediction will be closer to the globally stable trajectory.

**Safe Set and Control Barrier Functions**

In subsection 4.3.1, the basics of Control Barrier Functions were derived, and a Control Barrier Function for an aerial robot in $\mathsf{SE}(3)$ was introduced. Contrary to the introduced CBF, the now proposed CBF does not take the orientation of the robot with respect to the obstacle into account. Thus, the Control Barrier Function also applies for a position control in $\mathbb{R}^3$ not only for a system in $\mathsf{SE}(3)$.

From Equation 4.33, a new barrier function can be constructed as

$$g_{k|t,i}(\boldsymbol{x}_{k+1|t}) = \left\| \boldsymbol{x}_{k+1|t} - \boldsymbol{x}_{k+1|t,\mathrm{o},i} \right\|_2^2 - \tilde{r}_i^2 \geq 0. \tag{4.93}$$

The resulting safety region

$$\mathcal{B}_{k|t,i} = \left\{ \boldsymbol{x}_{k+1|t} \colon g_{k|t,i}(\boldsymbol{x}_{k+1|t}) \geq 0 \right\} \subseteq \mathcal{X} \tag{4.94}$$

is the safety region with respect to one sphere,

$$\mathcal{B}_{k|t} = \bigcap_{i=1}^{n_\mathrm{o}} \mathcal{B}_{k|t,i} \subseteq \mathcal{X} \tag{4.95}$$

the intersection for all $n_\mathrm{o}$ spheres, therefore the full safety region to be considered.

The only problem with $g_{k|t,i}(\boldsymbol{x}_{k+1|t})$ is, that this is a function of the configuration space, not the state space, and also does not have a relative degree of 1 [WuEtAl16a]. Consequently, let $h_{k|t,i} \colon \mathsf{TSE}(3) \to \mathbb{R}$ be

$$\begin{aligned} h_{k|t,i}(\boldsymbol{x}_{k+1|t}) &\coloneqq \gamma g_{k|t,i}(\boldsymbol{x}_{k+1|t}) + \dot{g}_{k|t,i}(\boldsymbol{x}_{k+1|t}) && (4.96)\\ &= \gamma(\boldsymbol{x}_{k+1|t} - \boldsymbol{x}_{k+1|t,\mathrm{o},i})^\mathsf{T}(\boldsymbol{x}_{k+1|t} - \boldsymbol{x}_{k+1|t,\mathrm{o},i}) - \gamma\tilde{r}_i^2 \\ &\quad + 2(\dot{\boldsymbol{x}}_{k+1|t} - \dot{\boldsymbol{x}}_{k+1|t,\mathrm{o},i})^\mathsf{T}(\boldsymbol{x}_{k+1|t} - \boldsymbol{x}_{k+1|t,\mathrm{o},i}) - 2\dot{\tilde{r}}_i\tilde{r}_i, && (4.97) \end{aligned}$$

and the corresponding time derivative

$$\begin{aligned} \dot{h}_{k|t,i}(\boldsymbol{x}_{k+1|t}) &= 2\gamma(\dot{\boldsymbol{x}}_{k+1|t} - \dot{\boldsymbol{x}}_{k+1|t,\mathrm{o},i})^\mathsf{T}(\boldsymbol{x}_{k+1|t} - \boldsymbol{x}_{k+1|t,\mathrm{o},i}) - 2\gamma\dot{\tilde{r}}_i\tilde{r}_i \\ &\quad + 2\Big((\dot{\boldsymbol{x}}_{k+1|t} - \dot{\boldsymbol{x}}_{k+1|t,\mathrm{o},i})^\mathsf{T}(\dot{\boldsymbol{x}}_{k+1|t} - \dot{\boldsymbol{x}}_{k+1|t,\mathrm{o},i}) \\ &\qquad + (\ddot{\boldsymbol{x}}_{k+1|t} - \ddot{\boldsymbol{x}}_{k+1|t,\mathrm{o},i})^\mathsf{T}(\boldsymbol{x}_{k+1|t,\mathrm{o},i} - \boldsymbol{x}_{k+1|t,\mathrm{o},i})\Big) \\ &\quad - 2\gamma(\ddot{\tilde{r}}_i\tilde{r}_i + \dot{\tilde{r}}_i\dot{\tilde{r}}_i). && (4.98) \end{aligned}$$

Thus,

$$B_{k|t,i}(\boldsymbol{x}_{k+1|t}) = \frac{1}{h_{k|t,i}(\boldsymbol{x}_{k+1|t})}, \quad \forall \quad i \in \mathcal{I}(t) \tag{4.99}$$

are all Control Barrier Functions, with $h_i(\boldsymbol{x}, t)$ as defined in Equation 4.97, and $\mathcal{I}(t)$ as the time-varying set of indices of all spheres. Moreover, with the new definition of $B_i$ and Equations 4.25 and 4.26, the condition in Equation 4.27, is rewritten as

$$\begin{aligned} \dot{B}_{k|t,i}(\boldsymbol{x}_{k+1|t}, \boldsymbol{u}_{k|t}) &\leq \frac{\gamma_i}{B_{k|t,i}(\boldsymbol{x}_{k+1|t})}, && (4.100)\\ \Leftrightarrow \mathcal{L}_f B_{k|t,i}(\boldsymbol{x}_{k+1|t}) + \mathcal{L}_g B_{k|t,i}(\boldsymbol{x}_{k+1|t})\boldsymbol{u}_{k|t} &\leq \frac{\gamma_i}{B_{k|t,i}(\boldsymbol{x}_{k+1|t})}, && (4.101)\\ \Leftrightarrow \dot{h}_{k|t,i}(\boldsymbol{x}_{k+1|t}) + \gamma h^3_{k|t,i}(\boldsymbol{x}_{k+1|t}) &\geq 0. && (4.102) \end{aligned}$$

This condition will be used to enforce obstacle avoidance, thus safety, on the aerial robot.

**Box Constraints**

The box constraints force the states to stay inside a virtual box. Depending on the flight space, it may be advisable to remove or alter box constraints so that the states of the dynamical system stay inside the flight space. Therefore,

$$\boldsymbol{x}_{k|t} \in \mathcal{F} \qquad \forall \quad k \in \{t, \ldots, t+N\}, \tag{4.103}$$

$$\Leftrightarrow \boldsymbol{x}_{\min} \leq \boldsymbol{x}_{k|t} \leq \boldsymbol{x}_{\max}, \qquad \forall \quad k \in \{t, \ldots, t+N\}, \tag{4.104}$$

where $\mathcal{F} \subseteq \mathcal{X}$ is a safe set defined by the box limits. Given the dimensions of the flight space of the University of California, Berkeley, Equation 4.104 is equivalent to:

$$-2\,\mathrm{m} \leq \boldsymbol{x}_{k|t}^{(1)} \leq 1\,\mathrm{m}, \qquad \forall \quad k \in \{t, \ldots, t+N\}, \tag{4.105}$$

$$-2.5\,\mathrm{m} \leq \boldsymbol{x}_{k|t}^{(2)} \leq 2.5\,\mathrm{m}, \qquad \forall \quad k \in \{t, \ldots, t+N\}, \tag{4.106}$$

$$0\,\mathrm{m} \leq \boldsymbol{x}_{k|t}^{(3)} \leq 6\,\mathrm{m}, \qquad \forall \quad k \in \{t, \ldots, t+N\}. \tag{4.107}$$

### 4.4.5 A novel Model Predictive Control Barrier Function

From combining the cost functions from Equations 4.67, 4.89 and 4.92, with the constraints from Equations 4.75, 4.81, 4.82, 4.83, 4.87, 4.90 and 4.100, and the optional constraint from Equation 4.104, the full novel proposed Model Predictive

Control Barrier Function (MPCBF) derives as:

$$
\begin{aligned}
J^{*}_{t\to t+N}(\boldsymbol{x}_t) = \min_{\substack{\boldsymbol{u}_{t|t},\dots,\boldsymbol{u}_{t+N|t}\\ \delta_t,\dots,\delta_{t+N}}} & \sum_{k=t}^{t+N} (\boldsymbol{x}_{k|t} - \boldsymbol{x}_{\mathrm{ref},k})^{\mathsf{T}} \boldsymbol{Q} (\boldsymbol{x}_{k|t} - \boldsymbol{x}_{\mathrm{ref},k}) + \boldsymbol{u}_{k|t}^{\mathsf{T}} \boldsymbol{R} \boldsymbol{u}_{k|t} + q\delta_k^2 \\
& + \boldsymbol{x}_{t+N+1|t}^{\mathsf{T}} \boldsymbol{Q}_{\mathrm{final}} \boldsymbol{x}_{t+N+1|t} \\
& + \sum_{k=t}^{t+N_{\mathrm{c}}-1} (\boldsymbol{u}_{k|t} - \boldsymbol{u}_{k+1|t})^{\mathsf{T}} \boldsymbol{R}_{\Delta} (\boldsymbol{u}_{k|t} - \boldsymbol{u}_{k+1|t}), && (4.108)\\
\text{subj. to } \; & \boldsymbol{x}_{k+1|t} = \boldsymbol{\Phi} \boldsymbol{x}_{k|t} + \boldsymbol{\Gamma} \boldsymbol{u}_{k|t}, && (4.109)\\
& \boldsymbol{x}_t = \boldsymbol{x}_{t|t}, && (4.110)\\
& \boldsymbol{u}_{k_{\mathrm{c}}|t}^{\mathsf{T}} \boldsymbol{u}_{k_{\mathrm{c}}|t} \le f_{\max}^2, && (4.111)\\
& \boldsymbol{u}_{k_{\mathrm{c}}|t}^{(3)} \ge 0, && (4.112)\\
& \boldsymbol{u}_{k_{\mathrm{c}}|t}^{(4)} = 1, && (4.113)\\
& \boldsymbol{u}_{t+N_{\mathrm{c}}+1\to t+N|t} = \boldsymbol{u}_{t+N_{\mathrm{c}}|t}, && (4.114)\\
& \dot{V}_{\boldsymbol{x}}(\boldsymbol{x}_{k+1|t}, \boldsymbol{u}_{k|t}) + \eta_{\boldsymbol{x}} V_{\boldsymbol{x}}(\boldsymbol{x}_{k+1|t}) \le \delta_k, && (4.115)\\
& \dot{B}_{k|t,i}(\boldsymbol{x}_{k+1|t}, \boldsymbol{u}_{k|t}) \le \frac{\gamma_i}{B_{k|t,i}(\boldsymbol{x}_{k+1|t})}, && (4.116)\\
& \boldsymbol{x}_{\min} \le \boldsymbol{x}_{k|t} \le \boldsymbol{x}_{\max}, && (4.117)\\
& \forall \;\; k \;= \{t,\dots,t+N\}, && (4.118)\\
& \forall \;\; k_{\mathrm{c}} = \{t,\dots,t+N_{\mathrm{c}}\}, && (4.119)\\
& \forall \;\; i \in \mathcal{I}(t). && (4.120)
\end{aligned}
$$

# Chapter 5

# Validation

This chapter contains the validation of the proposed controllers, as well as their comparison. First, the controller are validated on simulation data. Based on the robot's responses in simulation, combined with the corresponding control inputs, first clues can be gathered about the real life performance of each controller. This part will be discussed in section 5.1.

Second, the controllers will be evaluated on the real aerial robot, designed in chapter 3. The CLF-CBF-QP is not be tested on the real robot for reasons explained in section 5.2.

## 5.1 Simulations

Testing Euler Angle controller, CLF-CBF-QP and MPCBF in simulation is the first step for the validation of their performance. Obviously, the Euler Angle controller is not able to avoid obstacles in any way. It is the first controller implemented on the aerial robot. Therefore, it shall also be validated in the simulation for stability to compare the results with the real experiment data. This could potentially give clues on whether or not the dynamical model is approximating the real world well enough or if it is oversimplified. Also, as it is used as the attitude controller for the MPCBF, validating its stability is necessary to validate the stability of the MPCBF. Therefore, the controllers to be validated are the primitive Euler Angle controller, the optimized CLF-CBF-QP, as well as the MPCBF.

### Simulation Parameter

The used simulation parameters, such as sampling times, weights, and others, are listed in Table 5.1.

Table 5.1: Parameters for the simulation.

| | | | |
|---|---|---|---|
| Quadrotor | | | |
| $m$ | $1.02\,\mathrm{kg}$ | Frame | F330 (330 mm in diagonal) |
| $\boldsymbol{J}$ | $\mathrm{diag}(5, 5, 9.8) \cdot 10^{-3}\,\mathrm{kg\,m^2}$ | Rotors | $8 \times 4.5$ (8 in diameter) |
| $f_{\mathrm{max}}$ | $28.51\,\mathrm{N}$ | $f_{\mathrm{pos}}$ | $10\,\mathrm{Hz}$ |
| $\boldsymbol{\tau}_{\mathrm{max}}$ | $[1.66, 1.66, 0.21]^{\mathsf{T}}\mathrm{N\,m}$ | $f_{\mathrm{att}}$ | $500\,\mathrm{Hz}$ |
| Euler Angle (EA) | | | |
| $\boldsymbol{k}$ | $[4, 4, 4]^{\mathsf{T}}$ | $\boldsymbol{k}_{\mathrm{p}}$ | $[0.28, 0.28, 0.17]^{\mathsf{T}}$ |
| $\boldsymbol{k}_{\mathrm{v}}$ | $[3, 3, 3]^{\mathsf{T}}$ | $\boldsymbol{k}_{\mathrm{i}}$ | $[0, 0, 0]^{\mathsf{T}}$ |
| | | $\boldsymbol{k}_{\mathrm{d}}$ | $[0.07, 0.08, 0.02]^{\mathsf{T}}$ |
| Position-QP (CLF) | | | |
| $\boldsymbol{k}_{\boldsymbol{x}}$ | 8 | $\boldsymbol{Q}_{\boldsymbol{x}}$ | $\mathrm{diag}(5, 5, 25)$ |
| $\varepsilon_{\boldsymbol{x}}$ | 2 | $\boldsymbol{c}_{\boldsymbol{x}}$ | $\mathbf{0}_{3\times1}$ |
| | | $\eta_{\boldsymbol{x}}$ | 2.5 |
| Attitude-QP (CBF) | | | |
| $\boldsymbol{K}_{\boldsymbol{J}}$ | $\mathrm{diag}(1, 1, 0.1)$ | $\boldsymbol{Q}_{\boldsymbol{R}}$ | $\mathrm{diag}(25, 10, 10, 10, 250)$ |
| $\boldsymbol{E}_{\boldsymbol{R}}$ | $4 \cdot \boldsymbol{I}_3$ | $\boldsymbol{c}_{\boldsymbol{R}}$ | $\mathbf{0}_{5\times1}$ |
| $\boldsymbol{C}_{\boldsymbol{R}}$ | $\mathrm{diag}(30, 30, 3)$ | $\eta_{\boldsymbol{R}}$ | 150 |
| | | $\gamma_i$ | 5 |
| Model Predictive Control Barrier Function (MPCBF) | | | |
| $T_{\mathrm{s}}$ | $0.1\,\mathrm{s}$ | $\boldsymbol{Q}$ | $\mathrm{diag}(5, 5, 2.5, 1, 1, 0.5)$ |
| $T$ | $2.5\,\mathrm{s}$ | $\boldsymbol{Q}_{\mathrm{final}}$ | $\mathrm{diag}(5, 5, 2.5, 1, 1, 0.5)$ |
| $T_{\mathrm{c}}$ | $0.5\,\mathrm{s}$ | $\boldsymbol{R}$ | $\mathrm{diag}(2, 2, 0.1, 1 \cdot 10^{-12})$ |
| $N$ | 25 | $\boldsymbol{R}_{\Delta}$ | $\mathrm{diag}(0.2, 0.2, 0.01, 1 \cdot 10^{-13})$ |
| $N_{\mathrm{c}}$ | 5 | $q$ | 250 |

### Trajectory

Compared to a step input, either with a constant position $\boldsymbol{x}$, velocity $\dot{\boldsymbol{x}}$ or acceleration $\ddot{\boldsymbol{x}}$, a circular trajectory is far more demanding on the controller. Especially the position controller as it has to constantly adapt to a changing setpoint. The counterclockwise circular trajectory is defined as

$$\boldsymbol{x}_{\mathrm{ref}} = r_{\mathrm{c}} \begin{bmatrix} \cos\frac{2\pi t}{T} \\ \sin\frac{2\pi t}{T} \\ 0 \end{bmatrix} + \boldsymbol{x}_{\mathrm{c}}, \tag{5.1}$$

$$\frac{\mathrm{d}\boldsymbol{x}_{\mathrm{ref}}}{\mathrm{d}t} = \frac{2\pi r_{\mathrm{c}}}{T} \begin{bmatrix} -\sin\frac{2\pi t}{T} \\ \cos\frac{2\pi t}{T} \\ 0 \end{bmatrix} + \frac{\mathrm{d}\boldsymbol{x}_{\mathrm{c}}}{\mathrm{d}t}, \tag{5.2}$$

$$\frac{\mathrm{d}^2\boldsymbol{x}_{\mathrm{ref}}}{\mathrm{d}t^2} = \frac{4\pi^2 r_{\mathrm{c}}}{T^2} \begin{bmatrix} -\cos\frac{2\pi t}{T} \\ -\sin\frac{2\pi t}{T} \\ 0 \end{bmatrix} + \frac{\mathrm{d}^2\boldsymbol{x}_{\mathrm{c}}}{\mathrm{d}t^2}, \tag{5.3}$$

$$\frac{\mathrm{d}^3\boldsymbol{x}_{\mathrm{ref}}}{\mathrm{d}t^3} = \frac{8\pi^3 r_{\mathrm{c}}}{T^3} \begin{bmatrix} \sin\frac{2\pi t}{T} \\ -\cos\frac{2\pi t}{T} \\ 0 \end{bmatrix} + \frac{\mathrm{d}^3\boldsymbol{x}_{\mathrm{c}}}{\mathrm{d}t^3}, \tag{5.4}$$

$$\frac{\mathrm{d}^4\boldsymbol{x}_{\mathrm{ref}}}{\mathrm{d}t^4} = \frac{16\pi^4 r_{\mathrm{c}}}{T^4} \begin{bmatrix} \cos\frac{2\pi t}{T} \\ \sin\frac{2\pi t}{T} \\ 0 \end{bmatrix} + \frac{\mathrm{d}^4\boldsymbol{x}_{\mathrm{c}}}{\mathrm{d}t^4}, \tag{5.5}$$

where $r_{\mathrm{c}} > 0$ denotes the (constant) radius of the circle, $t > 0$ the current time since start, $T > 0$ the period, $\boldsymbol{x}_{\mathrm{c}} \in \mathbb{R}^3$ the center of circle, and the derivatives $\frac{\mathrm{d}^3\boldsymbol{x}}{\mathrm{d}t^3}$ and $\frac{\mathrm{d}^4\boldsymbol{x}}{\mathrm{d}t^4}$ the jerk and snap respectively. Using the third and fourth time-derivative, it is possible to generate a differential flat trajectory. For differential flat trajectories, the system's states and inputs are a function of the flat outputs and a finite number of their time derivatives. Here, the flat outputs are given by $\boldsymbol{x} = [x, y, z]^{\mathsf{T}}$. Based on the flat outputs, it can be guaranteed, that the resulting trajectory is smooth and that the underactuated aerial robot can follow this flat trajectory [MellingerEtAl11, SreenathEtAl13b].

### 5.1.1 Stability

Here, the stability of the different controllers is be investigated. Especially, as the proposed MPCBF is a pure position controller and lacks its attitude counterpart. Due to the cascaded nature of the control architecture, any stable attitude controller suits the MPCBF. As the MPCBF is avoiding obstacles on a higher level then the CLF-CBF-QP, there are no special requirements for the attitude controller. Therefore, it shall be determined, whether the simple proposed Euler-Angle attitude controller is suitable for the MPCBF, making it a MPCBF-EA. Otherwise, the orientation Lyapunov function can be used to create an attitude CLF controller, thus a MPCBF-CLF. Here, it is interesting to note, that the Control Barrier Function is now inside the position controller, and not the attitude controller, as opposed to the CLF-CBF-QP. The advantages will be clear, once the performance of the MPCBF and the CLF-CBF-QP will be compared.

### Attitude Controller

First, the behavior of the Euler Angle controller and the CLF-CBF-QP are compared for high initial attitude disturbances, ignoring the translatory motion of the system. The results from an initial disturbance of $\boldsymbol{\varphi}_0 = [-179°, 0°, 179°]^\mathsf{T}$ are depicted in 5.1. The values were chosen to be as close as possible to the maximum disturbance of $\boldsymbol{\varphi}_0 = [180°, 0°, 180°]^\mathsf{T}$, while not creating a gimbal lock problem, where the definition of the Euler Angle may break down, rendering it impossible for the controllers to even calculate the correct orientation. Contrary to [WuEtAl16a] and [Mueller18], the error decays way faster using the Euler Angle controller. In [WuEtAl16a], it is proven, that the CLF-CBF-QP is stable even for high angular errors. However, they do not mention the time needed for the error to decay completely. In this case, the CLF-CBF-QP needs roughly 30 s to settle.

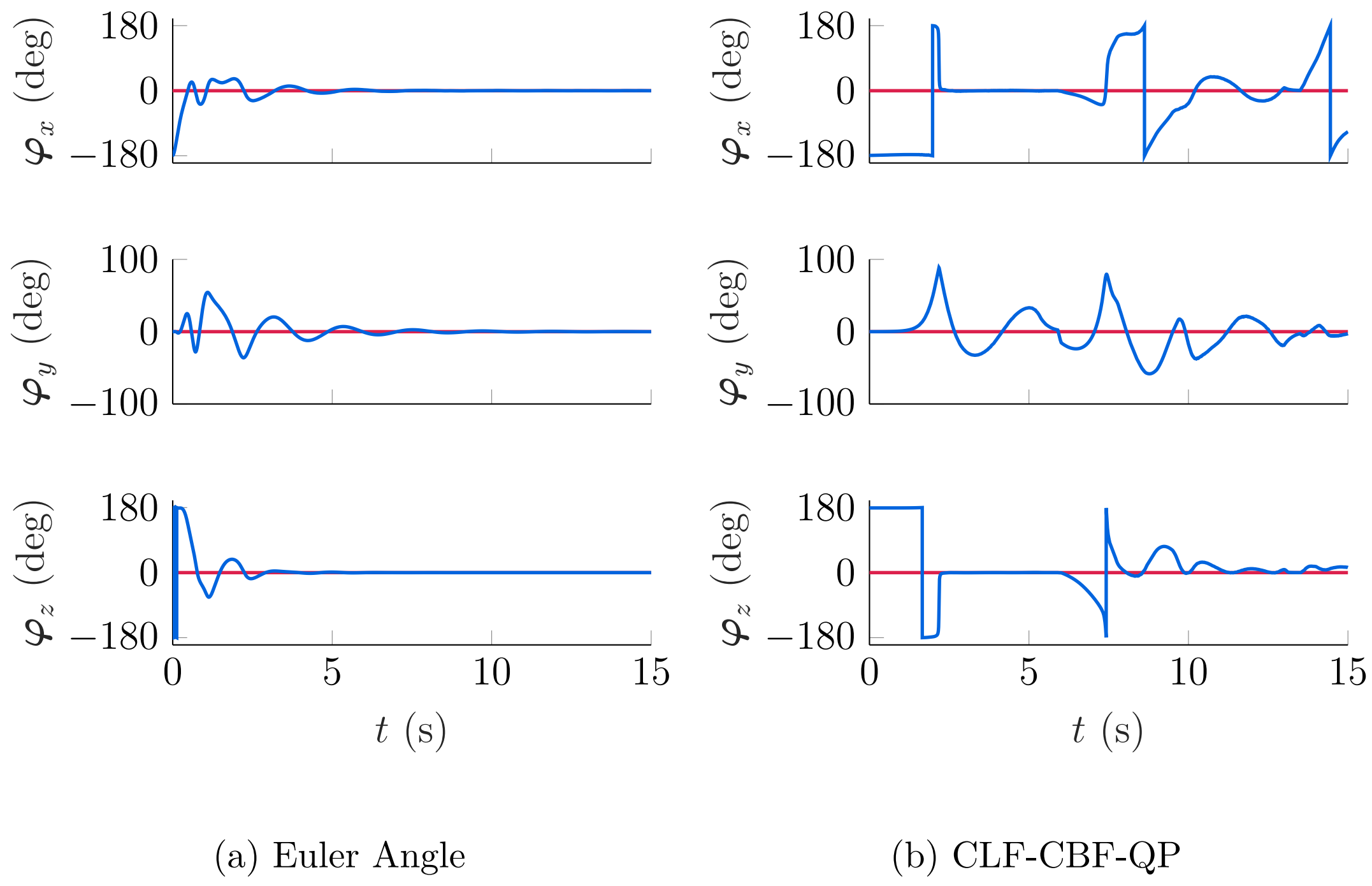

Figure 5.1: Performance of the Euler-Angle controller compared to the CLF-CBF-QP for high attitude errors.

In Figure 5.2b a very high fluctuation for the control outputs using the CLF-CBF-QP is visible. This results from the fact, that the controller is designed to optimize the flight time, hence reducing the energy consumption of the motors. As the energy consumption is proportional to the integral of torques squared over the whole time and each motor, minimizing the integral maximizes the flight time. The real world motors are not able to follow that trajectory. First, the time constant of the motors is higher then the sampling time of the attitude controller. Second, this "bang-bang" like behavior fo the controller requires the thrust and moments to be applied precisely when computed. Due to inevitable delays in the real system, the control outputs are always applied a little too late, roughly 10 ms to 50 ms. For that reason, the CLF-CBF-QP is not stable in real world behaving systems. A possible augmentation would be adding the time derivative of the control output to the optimization problem, like done by MPCBF.

In comparison, the control output of the Euler Angle controller is rather smooth. The high frequency fluctuations arise from discrete behavior of the position control. In between the update rate of $\boldsymbol{f}_{\text{des}}$, $\boldsymbol{\tau}$ is smooth, but for every update of $\boldsymbol{f}_{\text{des}}$ a discontinuity in $\boldsymbol{\tau}$ occurs.

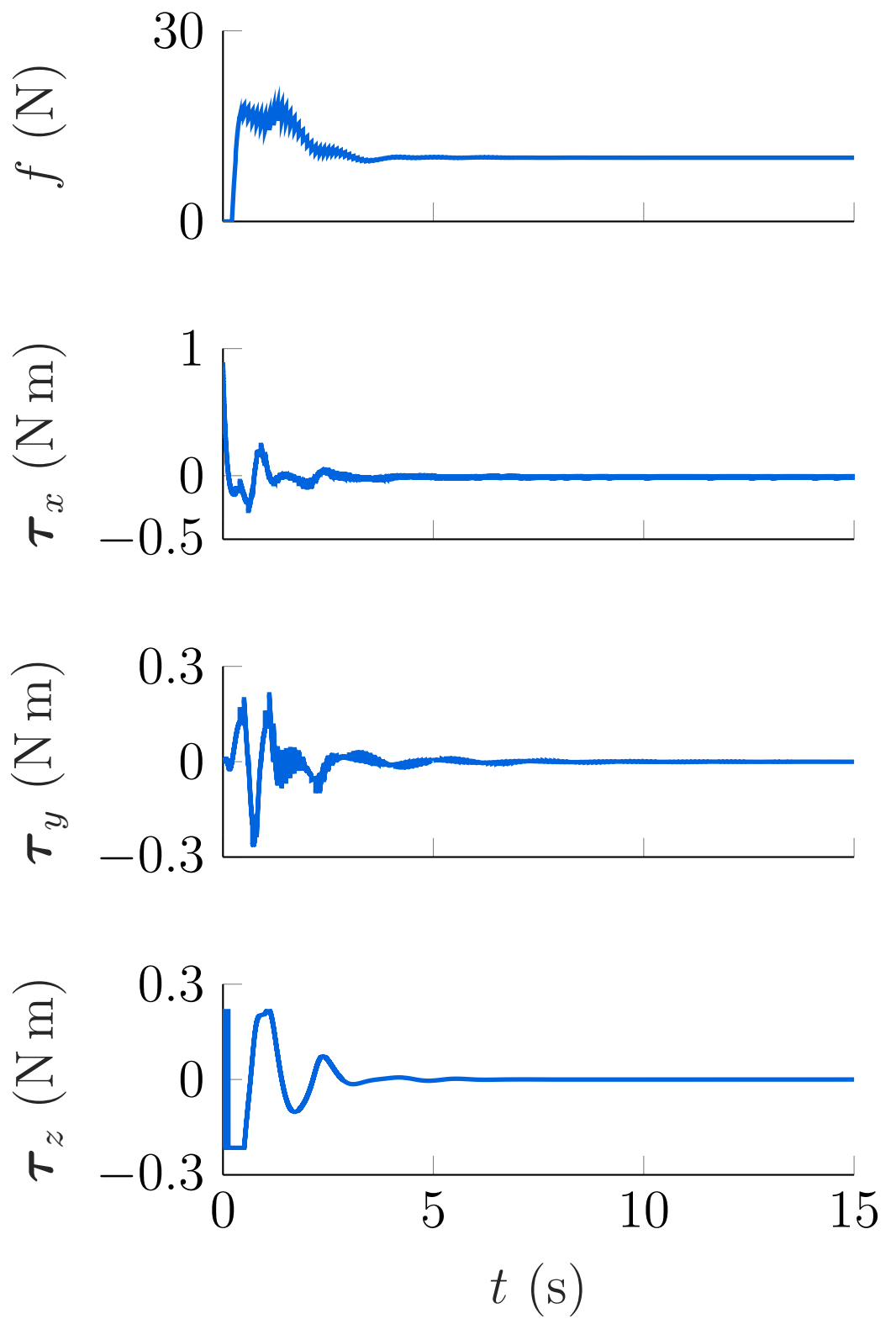

(a) Euler Angle

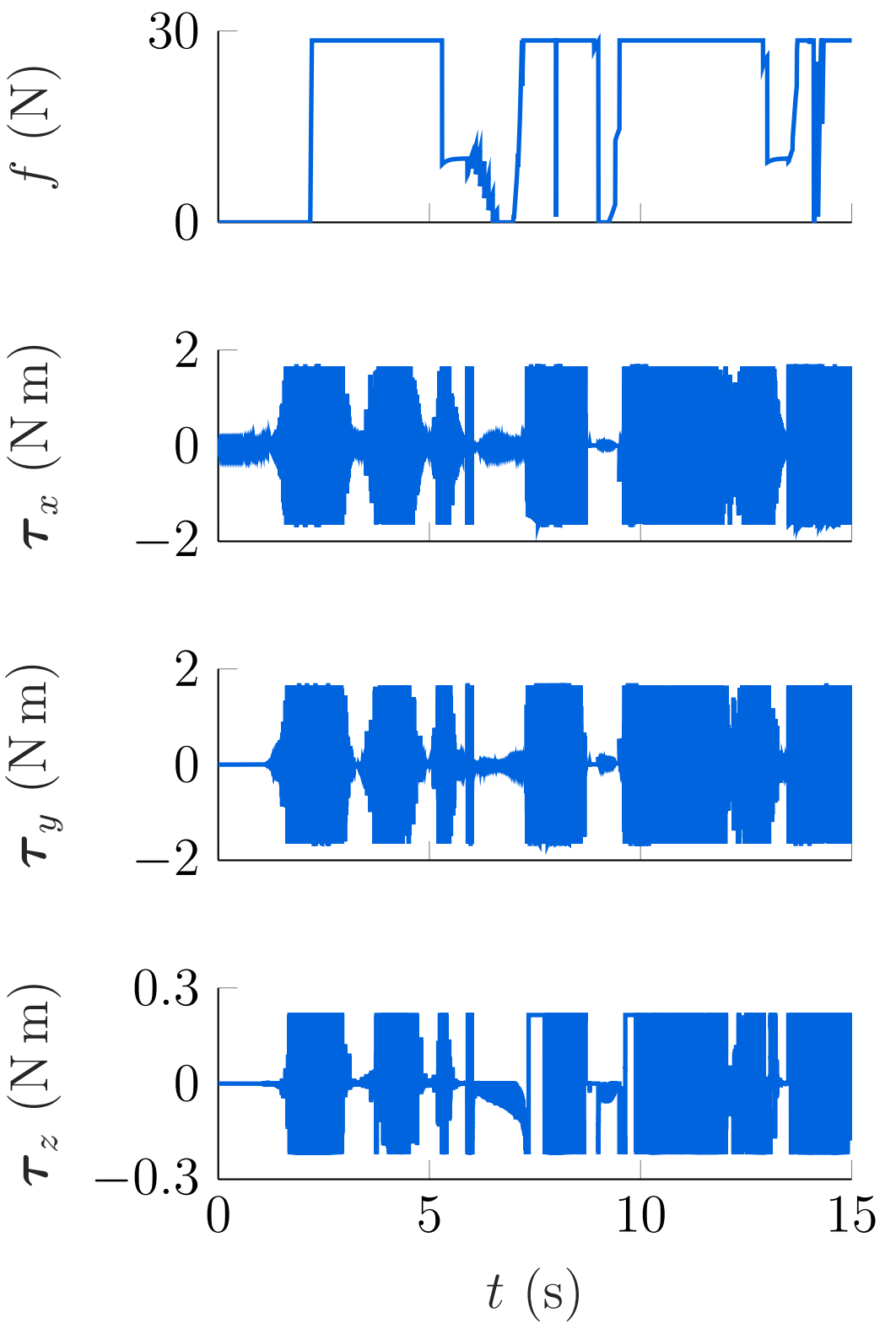

(b) CLF-CBF-QP

Figure 5.2: Comparison of the control outputs from the Euler-Angle controller and the CLF-CBF-QP for high attitude errors.

For smaller input disturbances with $\boldsymbol{\varphi}_0 = [30°, 30°, 30°]^\mathsf{T}$, both controllers behave similarly, as depicted in Figure 5.3. Both the Euler Angle controller and the CLF-CBF-QP can stabilize the robot within approximately 5 s resulting in roughly the same performance. Here, the settling times for the Euler Angle controller are 1.8 s, 5.3 s, and 1.7 s for $\varphi_x$, $\varphi_y$, and $\varphi_z$, respectively. For the CLF-CBF-QP, the settling times are 2.9 s, 5.8 s, and 1.6 s. Nevertheless, the control output of the CLF-CBF-QP is similar to Figure 5.2b as it again has the same "bang-bang" like behavior.

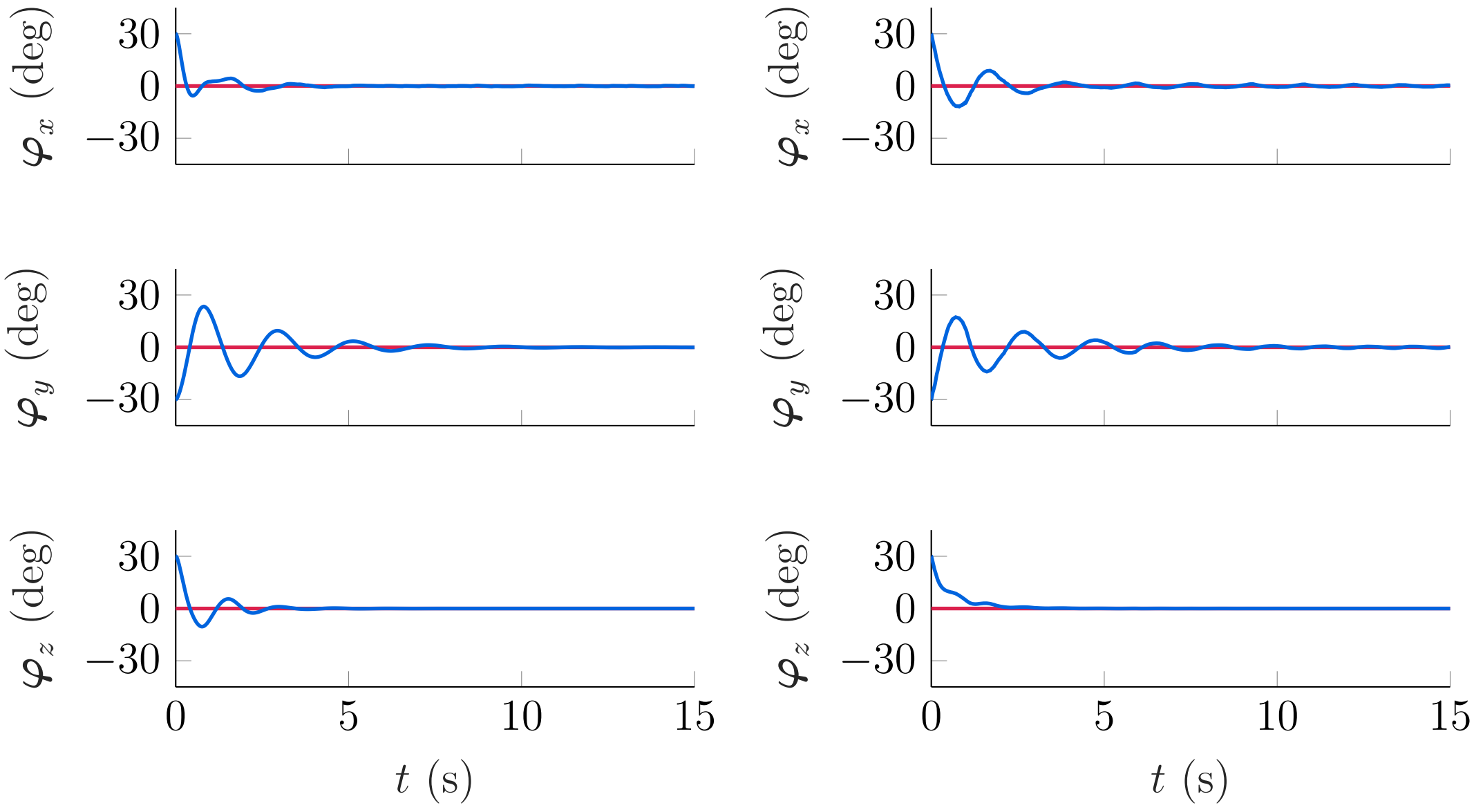

(a) Euler Angle (b) CLF-CBF-QP

Figure 5.3: Performance of the Euler-Angle controller compared to the CLF-CBF-QP for small attitude errors.

To compare the expected energy usage of each controller, the integral over the square of all individual thrusts is used. The reason being, that the thrust is directly proportional to the the squared rotation speed of each motor, $\omega_i^2$, and the rotation speed itself is proportional to the electrical current, given a fixed electrical voltage across the motor. Thus, the electrical power, which is current times voltage, integrated over the flight time, yields the total energy usage for each individual rotor. Therefore,

$$E^* \propto \sum_{i=1}^{4} \int_0^{t_\text{end}} f_i^2(t)\mathrm{d}t, \tag{5.6}$$

where the $E^*$ is the total expected energy usage over the duration of flight. Given a discrete sampling time of the controller, this equals

$$E^* \propto \frac{1}{f_{\text{att}}} \sum_{i=1}^{4} \sum_{t=0}^{t_{\text{end}}} f_i^2(t). \tag{5.7}$$

The thrust at each motor is a combination of the total thrust and all three moments. Here, the thrust is equally spread across all four motors and the moments either increase or decrease the thrust, depending on the motor. For example, for motor 3, the front left motor, all moments contribute positive towards the thrust. Roughly speaking, the three moments are scaled by the maximum possible moments to $-1$ to 1, added together, limited, added to the scaled thrust (0 to 1), and then limited to 0 to 1. The result is the scaled thrust at each motor, which then gets converted back by PX4 to a PWM signal for the ESCs. The reason being, that for a maximum moment, say along the yaw axis, the motors 3, front left, and 4, back right, would have to spin up to the maximum speed, and motors 1 and 2, front right and back left respectively, would need to almost stop spinning. However, this would obviously also introduce a thrust of 50 % of the maximum thrust. Thus, if the thrust is only 30 % of the maximum thrust, the maximum achievable yaw moment is only 60 % of the maximum possible yaw moment. So the total moment is limited to $-\min(f_{i,\text{s}}, |1 - f_{i,\text{s}}|)$ to $\min(f_{i,\text{s}}, |1 - f_{i,\text{s}}|)$, where $f_{i,\text{s}}$ is the scaled thrust of the $i$-th motor. Using the results from Figure 5.3, the total expected energies are proportional to 7 651.2 and 9 460.6 for the Euler Angle controller and the CLF-CBF-QP, respectively. While, based on the numbers themselves, there is no possibility to predict the total maximum flight time, their relation shows, that the expected maximum flight time for the Euler Angle controller actually is almost 25 % longer. Part of the reason may be the aggressive behavior of the controller, and weighing attitude errors higher than control inputs. With further tuning it might be possible to increase the expected flight time. Still, the aggressive control outputs are hard to follow for the motor and possible effects from high frequency changes are completely disregarded. Those could potentially further decrease the performance of the motors or the flight time. It is also to be noted, that at full throttle, each motor draws approximately 170 W. Compared to that, the power draw of 5 W to 10 W for the rest of the hardware, mainly the Jetson TX2, is neglectable.

Due to the superior behavior of the Euler Angle controller, it is chosen as the primary attitude controller for the MPCBF, making it a MPCBF-EA. A reason to use a attitude CLF controller with the MPCBF would be to guarantee exponential stability for the attitude. However, the proposed Lyapunov function would introduce the non desirable "bang-bang" like behavior of the control output. While the Euler Angle controller does not guarantee exponentially stability by design, it can stabilize the robot as good as the attitude CLF.

### Position Controller

The position controller is not as critical to the global stability as the attitude controller. As previously mentioned, a faulty or non-existing position controller will not hinder the attitude controller from stabilizing the robot. But it does greatly impact the reference tracking ability of the robot. Hence, a good position controller is needed to minimize the deviation from the reference trajectory. In the case of the MPCBF, the position controller is also responsible for the obstacle avoidance.

The three position controllers, MPCBF, CLF-CBF-QP, and Euler-Angle, were tasked to follow a circular reference trajectory, see section 5.1. With a radius of $r_\mathrm{c} = 1.5\,\mathrm{m}$ and a period of as low as $T = 4\,\mathrm{s}$, the average speed was up to approximately $2.4\,\frac{\mathrm{m}}{\mathrm{s}}$. The results for $T = 8\,\mathrm{s}$ and $T = 4\,\mathrm{s}$ are shown in Figure 5.4a and Figure 5.4b, respectively. To compare the average deviation, the mean error $e_\mathrm{mean}$ as

$$e_\mathrm{mean} = \frac{1}{T \cdot f_\mathrm{att}} \sum_{t=0}^{T} \|\boldsymbol{x}_t - \boldsymbol{x}_{t,\mathrm{ref}}\|_2 \tag{5.8}$$

is used. For $T = 8\,\mathrm{s}$, all three controllers where able to follow the reference trajectory smoothly with a mean error of $e_\mathrm{mean,MPCBF} = 0.255\,\mathrm{m}$, $e_\mathrm{mean,CLF-CBF-QP} = 0.330\,\mathrm{m}$, and $e_\mathrm{mean,EA} = 0.393\,\mathrm{m}$. But by halving the period to $T = 4\,\mathrm{s}$, the tracking problem is significant harder for all three controllers. The mean errors are now $e_\mathrm{mean,MPCBF} = 0.597\,\mathrm{m}$, $e_\mathrm{mean,CLF-CBF-QP} = 1.118\,\mathrm{m}$, and $e_\mathrm{mean,EA} = 1.190\,\mathrm{m}$. In the case fo the CLF-CBF-QP more than three times as high. Due to the model predictions in the MPCBF it can handle these situations better compared to the CLF-CBF-QP. Based on the results, it can be concluded, that at least for slow flights all three controllers can track a reference trajectory with the MPCBF following it the closest.

## 5.1.2 Obstacle Avoidance

Since the goal is to design a controller for obstacle avoidance, the two proposed controllers, namely the MPCBF(-EA) and the CLF-CBF-QP, have to be tested in simulations with obstacles. Avoiding a single very far away obstacle with a low relative speed is a fairly easy task for both. In this case, the controller has enough time to adapt and prepare for the encounter. The hard problems for both controllers are environments with either very fast moving, suddenly appearing, or many obstacles. In the first two cases, the controller has to react fast to drastic changes, in the latter case, it has to factor in multiple obstacles, which can significantly decrease the safe set $\mathcal{B}_{k|t}$. Assuming that both controllers are able to avoid obstacles, the first comparison will be of an obstacle flying close by, but not colliding with the robot as long as it stays at its reference position.

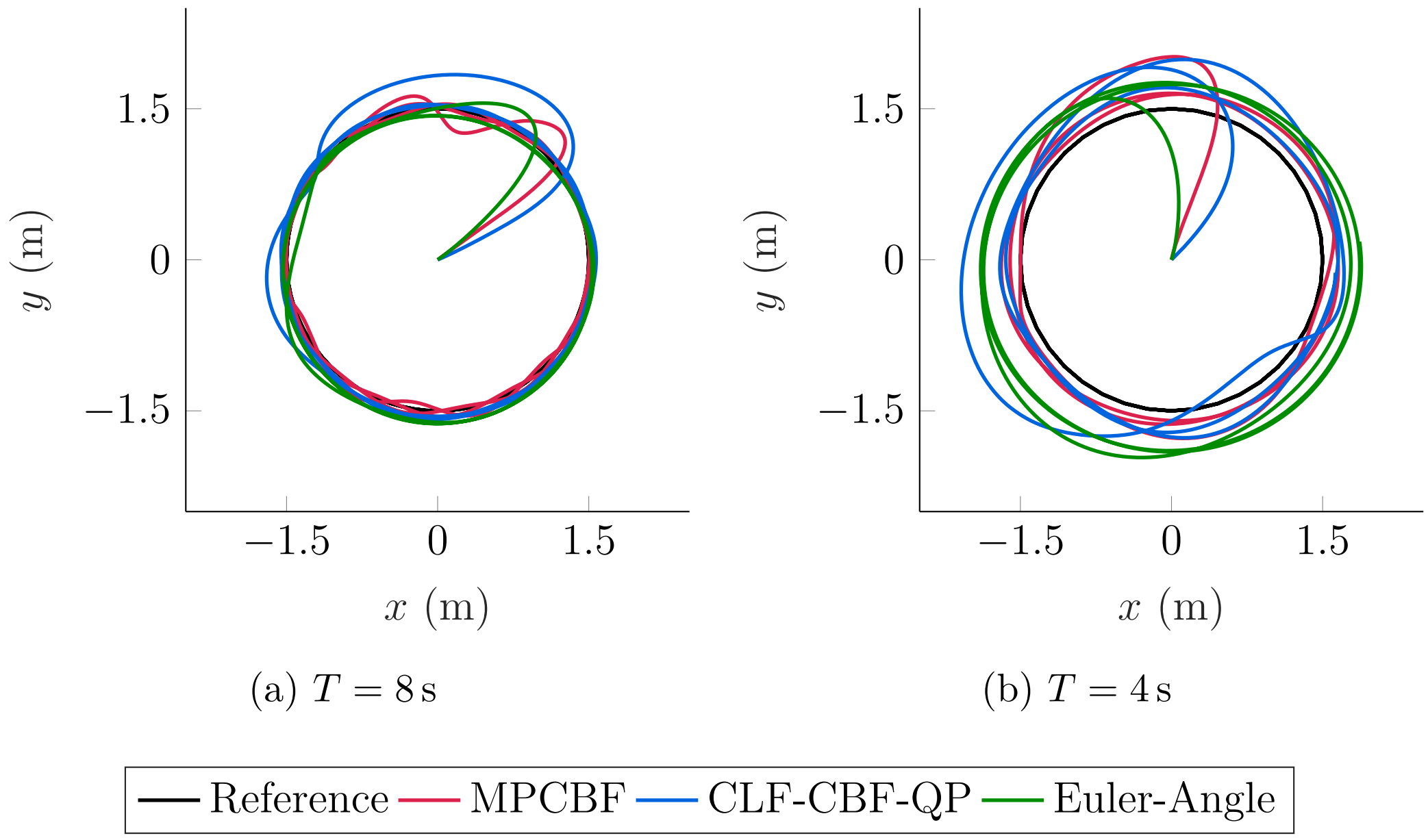

Figure 5.4: Top view of the aerial robot's trajectory.

Secondly, an obstacle will be shot with a high relative velocity onto the robot, thus evaluating the reaction time. In the first two tests, it is expected, that the MPCBF will perform better, but at least as good as the CLF-CBF-QP. The reason is, that due to the ability to predict its own position and that of the obstacle, it can form a more sophisticated decision, whether or not and in what direction to avoid the obstacle. Lastly, both controllers will participate in a flight through a highly cluttered environment. For all tests, the position, velocity, and acceleration of all obstacles is assumed to be known at all times.

### Close-by obstacles

In the following, the behavior of both controllers shall be investigated, with obstacles flying close to the robot, but not in the direct way of it. According to Table 5.1, the robot's frame has a length of 330 mm and the rotor's diameter is 8 in, or roughly 200 mm. Therefore, a sphere around the robot would have an radius of $r = 265\,\text{mm}$, ignoring slight offsets and conversion errors, without any safety measures. A sphere with a radius of $\boldsymbol{r}_\text{o} = 1\,\text{m}$ could therefore come as close as 1.265 m measuring the distance from both centers. Given some small margin of errors, for a sphere whose center is as close as 1.28 m, even if the obstacle sphere is going at high speeds, for example at least $2\,\frac{\text{m}}{\text{s}}$, the controller does not have to move the robot in $\mathbb{R}^3$. Nevertheless, determining if the obstacle is too close to the obstacle and therefore the aerial robot has to be moved, can be a potentially hard problem. As a result, the behavior in such a case shall be investigated. Based on the previous assumptions, the obstacle's position is given as

$$\boldsymbol{x}_\mathrm{o}(t) = \begin{bmatrix} 1.28\,\mathrm{m} \\ 0\,\mathrm{m} \\ -5\,\mathrm{m} \end{bmatrix} + \begin{bmatrix} 0\,\frac{\mathrm{m}}{\mathrm{s}} \\ 0\,\frac{\mathrm{m}}{\mathrm{s}} \\ 2\,\frac{\mathrm{m}}{\mathrm{s}} \end{bmatrix} t, \quad t > 0\,\mathrm{s}. \tag{5.9}$$

The resulting behavior of both the MPCBF and the CLF-CBF-QP is depicted in Figure 5.5. As a first result, both avoid the obstacle without problem. However, while the MPCBF moves ever so slightly, the CLF-CBF-QP tries to avoid the obstacle by moving further from the reference setpoint. Again, to compare the controller's performance, the mean error, see Equation 5.8, is used. In an ideal situation, the mean error would be zero for both controllers as they start exactly on the reference setpoint and the obstacle misses the aerial robot, even if only by a few cm. Yet, the mean errors are $e_{\mathrm{mean,MPCBF}} = 0.09\,\mathrm{m}$ and $e_{\mathrm{mean,CBF}} = 2.96\,\mathrm{m}$ for the MPCBF and the CLF-CBF-QP respectively. It is clearly observable, that the MPCBF does in fact not move by much around the reference trajectory. Only a little bit to the back, presumably to create some artificial, not explicitly defined safety margin. The CLF-CBF-QP however, flies up a lot as it tries to increase the distance between the aerial robot and the obstacle. Due to the constraints involved, it not only tries to maximize the distance but also minimize the relative velocity between the obstacle and the aerial robot. It would have been more effective to move back along $x$ to increase the distance instead of moving up along $z$. Furthermore, due to the high error along $z$ the CLF-CBF-QP has a hard time decreasing the error along $z$ again and stabilizing the aerial robot.

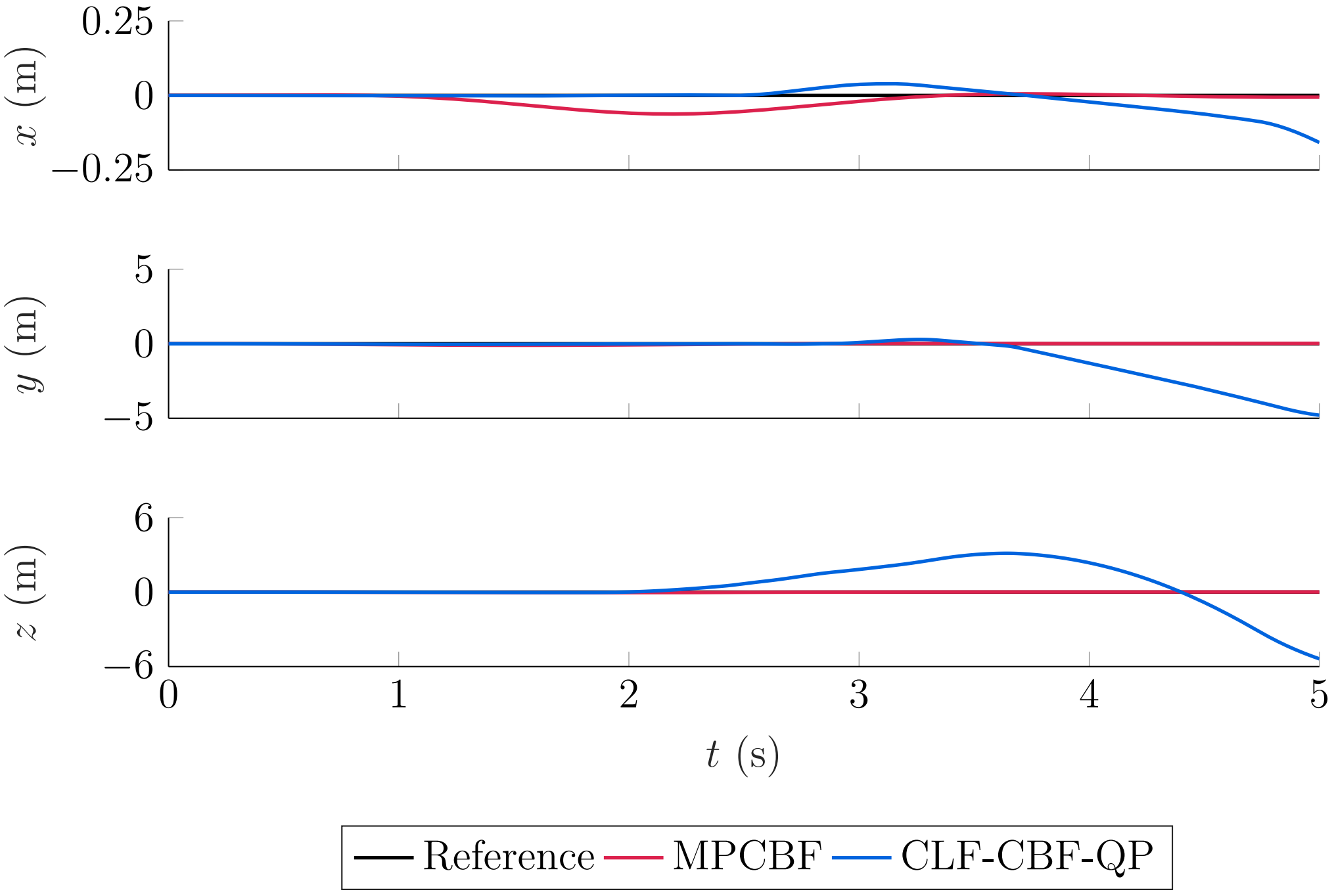

Figure 5.5: Behavior of the MPCBF and the CLF-CBF-QP given a close-by obstacle.

### High-Speed and -Performance Obstacle Avoidance

One possible scenario is a relative motion between the aerial robot and an obstacle that would inevitably lead to a collision, if not counteracted on. This does also include two aerial robots flying towards each other without intercommunication or where one is not equipped with any obstacle avoidance technology. Validating and evaluating the performance of the controllers in such a scenario can help to better understand the limits of the controller. Suppose the aerial robot is not flying close to its physical limits, then there is no significant difference to whether the aerial robot moves towards a static obstacle or an obstacles moves towards the aerial robot, which again is tracking a fixed reference point. Therefore, the factors which do have an impact on the performance are the relative speed, the initial distance, and the radius of the obstacle. Here, the first two factors roughly translate to one, which is the time the controller has to act on the approaching obstacle. As one possible avoidance strategy would be to always give full thrust, if state constraints allow for a movement in the $z$ coordinate of the body frame. It comes down to the time the controller has to clear the space the obstacle will occupy once it reaches the current position of the aerial robot.

The first simulation will be conducted with the aerial robot at its desired position, the origin, and an obstacle with $\boldsymbol{x}_\mathrm{o}(0) = [4\,\mathrm{m}, 0\,\mathrm{m}, 0\,\mathrm{m}]^\mathsf{T}$, $\dot{\boldsymbol{x}}_\mathrm{o}(0) = [-2\,\frac{\mathrm{m}}{\mathrm{s}}, 0\,\frac{\mathrm{m}}{\mathrm{s}}, 0\,\frac{\mathrm{m}}{\mathrm{s}}]^\mathsf{T}$, and $r_\mathrm{o} = 0.5\,\mathrm{m}$. The relative speed is fairly low and the controller had 2 s until predicted collision. In this scenario, both controllers can avoid the obstacle as seen in Figure 5.6. Looking at the predicted movement of the MPCBF, marked in blue, clearly shows its anticipation to move down in order to avoid the incoming obstacle. It also shows, that after the avoidance maneuver, the MPCBF flies back towards the reference trajectory. Moreover, comparing the three states, in Figures 5.6a, 5.6b and 5.6c, shows that the aerial robot follows the predicted trajectory at $t = 1\,\mathrm{s}$ closely. However, the MPCBF acts sooner onto the approaching obstacle. This does not mean that this is the better behavior in any case, it very much depends on the objective of safety versus reference tracking. Between $t = 0\,\mathrm{s}$ and $t = 2\,\mathrm{s}$, the mean error of the MPCBF is in fact higher than the CLF-CBF-QP's, $e_{\mathrm{mean,MPCBF}} = 1.25\,\mathrm{m}$ compared to $e_{\mathrm{mean,CBF}} = 0.48\,\mathrm{m}$. Nevertheless, for higher speeds the behavior of the CLF-CBF-QP might not the optimal behavior. From an optimization standpoint, the CLF-CBF-QP clearly does a better job of minimizing the position error, but from a safety standpoint, the MPCBF's behavior might still be favored. Especially for higher speeds.

This advantage for the CLF-CBF-QP quickly turns into a disadvantage for higher velocities, see Figure 5.7. For the second simulation, with an obstacle at $\boldsymbol{x}_\mathrm{o}(0) = [5\,\mathrm{m}, 0\,\mathrm{m}, 0\,\mathrm{m}]^\mathsf{T}$, $\dot{\boldsymbol{x}}_\mathrm{o}(0) = [-5\,\frac{\mathrm{m}}{\mathrm{s}}, 0\,\frac{\mathrm{m}}{\mathrm{s}}, 0\,\frac{\mathrm{m}}{\mathrm{s}}]^\mathsf{T}$, and $r_\mathrm{o} = 0.75\,\mathrm{m}$, the CLF-CBF-QP is no longer able to avoid the obstacle, see Figure 5.7. Here, a bigger obstacle was deliberately chosen to increase the difficulty for the controllers. Likewise, an even higher speed or shorter initial distance could have been chosen. Now [WuEtAl16a] proves that, given a feasible solution, $\mathcal{C}$ is always forward invariant and therefore, the aerial robot's state will always stay within $\mathcal{B}_{k|t}$. Thus, the CLF-CBF-QP would avoid the obstacle. Nevertheless, this is based on the assumption that there always is a feasible solution to the optimization problem. This assumption does not hold true for example if the controller acts too slowly and already collided with the obstacle and is thus inside the virtual sphere. The constraints can also turn infeasible if the obstacle is too close or too fast and there is no control input that would allow for a successful avoiding maneuver. In such a case, there are two main options to handle the situation. Either calculate no new control outputs, keeping the last ones or setting all to zero, or removing the infeasible constraint. The last one would allow for a potential collision with the obstacle but would still allow the flight to continue. Thus, the latter option was chosen. A third and rather trivial solution would be to abort the simulation and thus the flight. Regardless, this is not possible in a real experiment.

As already mentioned, the CLF-CBF-QP is not able to avoid the obstacle in the second scenario. Clearly, this is not due to the physical limits but rather to the fact that the controller acts too late on the approaching obstacle. What seemed to be a disadvantage of the MPCBF for slower velocities turned into the reason the MPCBF is able to avoid the obstacle. The problem is, that the CLF-CBF-QP does not take the relative speed into account enough and weights the distance higher. Thus it reacts later, which, consequently, leads to a collision.

The MPCBF correctly predicts the further movement of the aerial robot at $t = 0.5\,\mathrm{s}$, see Figure 5.7a. Nevertheless, the predicted overshoot of $0.5\,\mathrm{m}$ to $0.75\,\mathrm{m}$ is considerably high. It is, however, drastically reduced at $t = 1\,\mathrm{s}$, see Figure 5.7c. This clearly depicts, that, while $N_\mathrm{c} < N$, the predicted trajectory will still converge towards the reference trajectory as time goes on. It also shows, that the MPCBF can predict the optimal trajectory from early on, which is an obvious advantage compared to the CLF-CBF-QP. Nevertheless, in the real flight tests, see section 5.2, there will be a model mismatch and thus the predicted trajectory will not necessarily align with the real trajectory.

Based on the simulations alone, the MPCBF could avoid the same obstacle even if it was moving towards the aerial robot with $30\,\frac{\mathrm{m}}{\mathrm{s}}$. Any faster and even the MPCBF would collide due to the physical limits of the aerial robot. If the aerial robot was moving towards a stationary obstacle, the maximum speed at which the MPCBF could avoid the obstacle is roughly $50\,\frac{\mathrm{m}}{\mathrm{s}}$. Both cases with the assumption, that the controller has 2.5 s, the prediction horizon, to react. At those speeds other unmodeled effects like the drag force have a significant impact on the performance. Moreover, it is not discussed how the aerial robot would even reach speeds this high. Those results are purely based on the underlying simulation, which are built on the assumption of slow speeds. To make any valid statement on such high speeds, a different model would have to be implemented into the simulation.

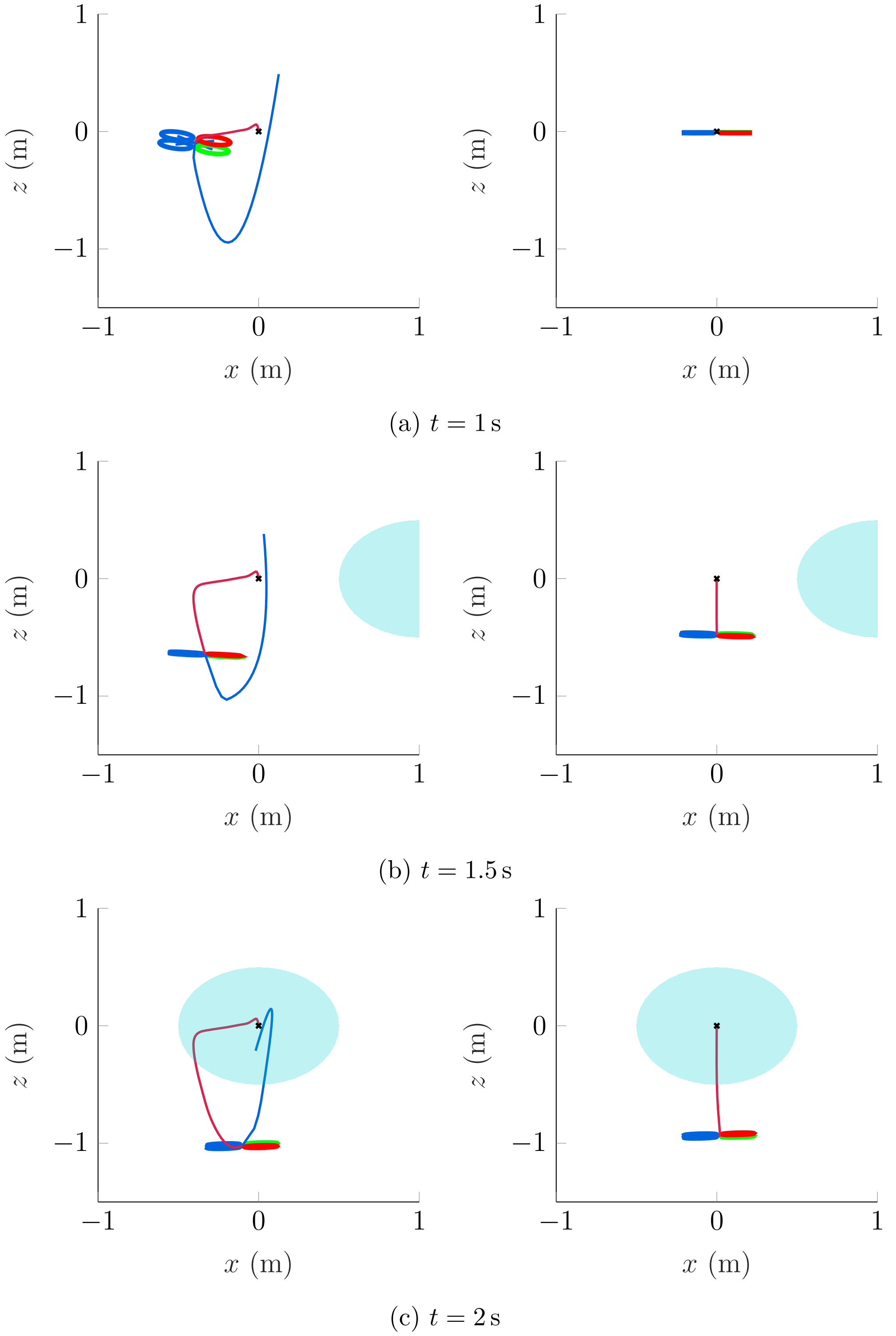

Figure 5.6: Different screenshots of the MPCBF (left) and the CLF-CBF-QP (right) at different times, with $\boldsymbol{x}_\text{o}(0) = [4\,\text{m}, 0\,\text{m}, 0\,\text{m}]^\mathsf{T}$, $\dot{\boldsymbol{x}}_\text{o}(0) = [-2\,\frac{\text{m}}{\text{s}}, 0\,\frac{\text{m}}{\text{s}}, 0\,\frac{\text{m}}{\text{s}}]^\mathsf{T}$. The black $\times$ marks the desired position, blue the predicted path and red the previous path.

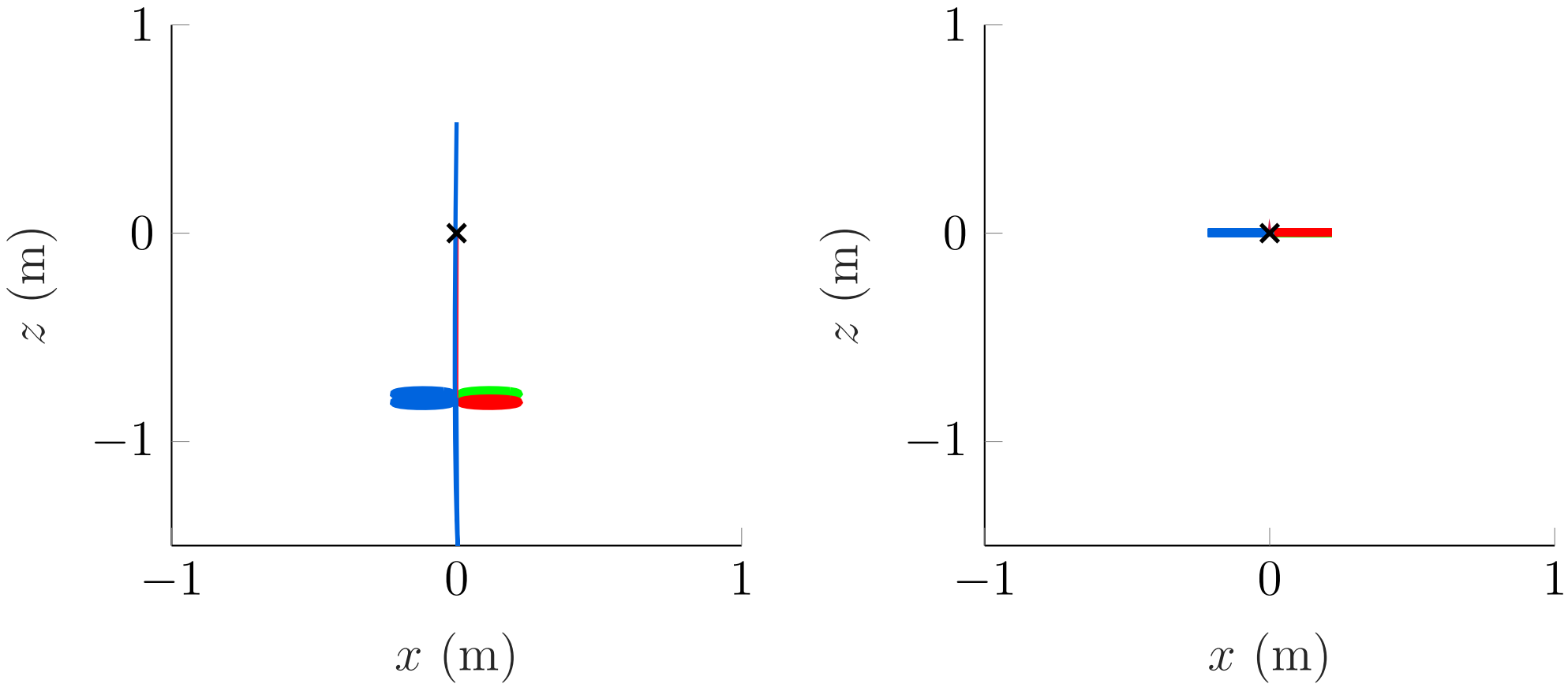

(a) $t = 0.5\,\mathrm{s}$

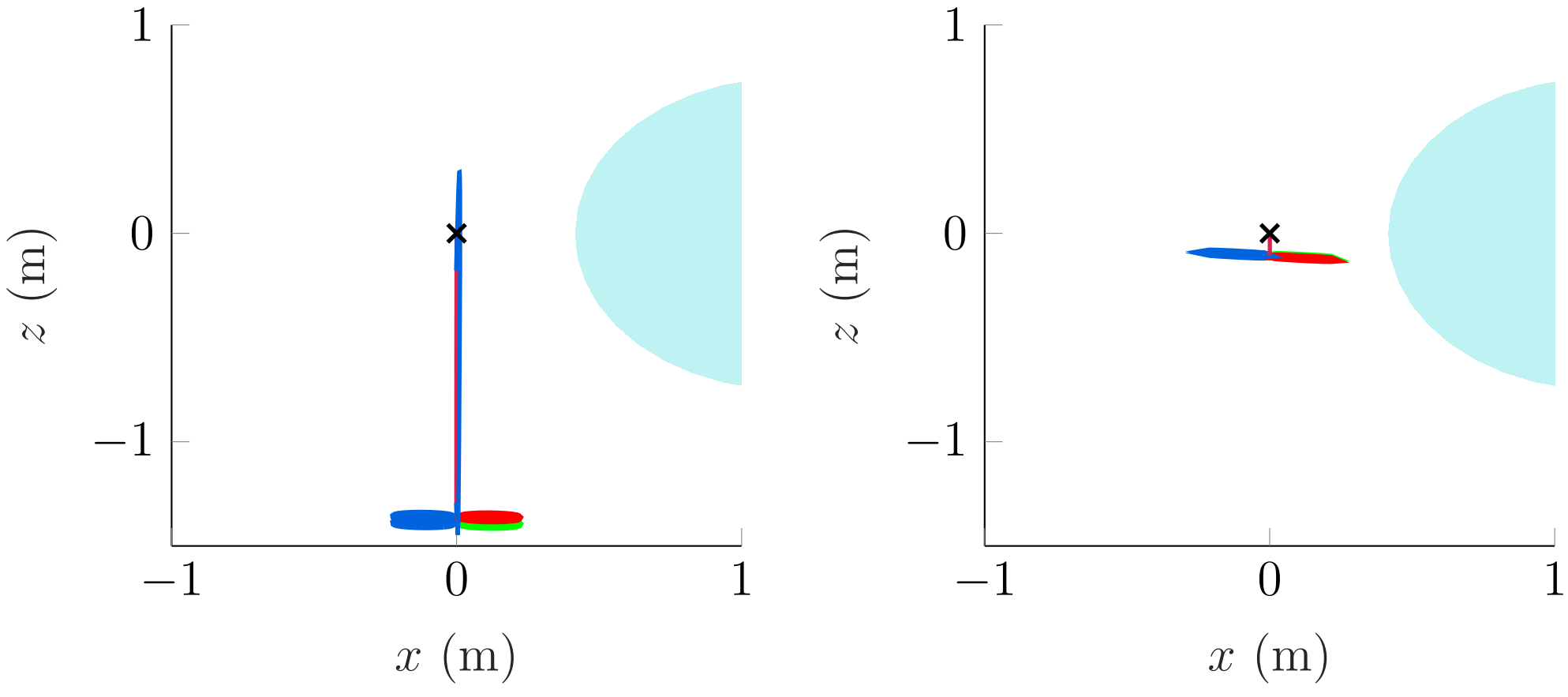

(b) $t = 0.75\,\mathrm{s}$

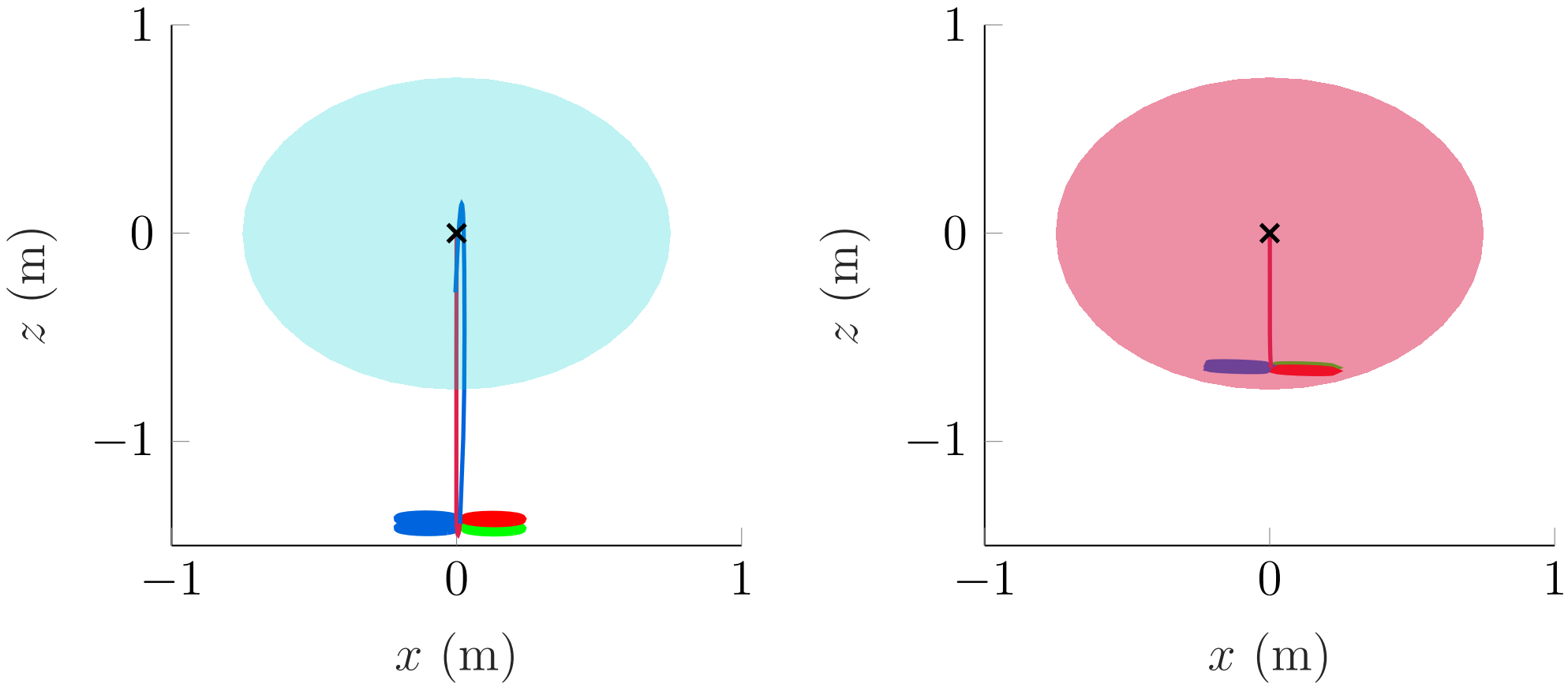

(c) $t = 1\,\mathrm{s}$

Figure 5.7: Different screenshots of the MPCBF (left) and the CLF-CBF-QP (right) at different times, with $\boldsymbol{x}_{\mathrm{o}}(0) = [5\,\mathrm{m}, 0\,\mathrm{m}, 0\,\mathrm{m}]^{\mathsf{T}}$, $\dot{\boldsymbol{x}}_{\mathrm{o}}(0) = [-5\,\frac{\mathrm{m}}{\mathrm{s}}, 0\,\frac{\mathrm{m}}{\mathrm{s}}, 0\,\frac{\mathrm{m}}{\mathrm{s}}]^{\mathsf{T}}$. The black $\times$ marks the desired position and red the previous path.

### Highly Cluttered Environments

The last test for performance comparison is a flight through a highly cluttered environment. Such scenarios could be found in a forest, but also in a classic indoors environment, with highly cluttered obstacles. Those would then need multiple spheres to be approximately correctly. Each obstacle adds one constraint to the optimization problem, therefore, reducing $\mathcal{B}_{k|t}$. Consequently, the number of possible control outputs shrinks. Here, it is vital for the controller not to maneuver itself into a situation, where there exists no feasible solution to the optimization problem. Doing so would most likely result in a crash and potentially loss of the robot. Nevertheless, the CLF-CBF-QP was again affected by infeasible solutions in flight. While those were also removed to continue the flight, it often results in violation of $\mathcal{B}_{k|t}$, thus, a virtual collision with a sphere. It is possible, that the CLF-CBF-QP loses control due to infeasible constraints. Recovering from such situations is often a hard problem which results in recovery maneuvers creating huge errors.

The simulations were done with 40 static obstacles randomly positioned inside a box of $x = -4\,\text{m}$ to $4\,\text{m}$, $y = -4\,\text{m}$ to $4\,\text{m}$, $z = -2\,\text{m}$ to $2\,\text{m}$, see Figure 5.8. Here, obstacles which intersect with the desired trajectory are shown in red, all others in a light shade of green. Only three obstacles intersect with the trajectory, a circle with radius $r_\text{c} = 1.5\,\text{m}$, but a few more are in the direct flight space of the aerial robot flying along the circle. The obstacles around and within the circle increase the difficulty for the controller to avoid the obstacles which are in its direct way, without moving too far from the desired trajectory. Furthermore, the box constraint was set to the box around the obstacles.

Running the simulation for $60\,\text{s}$, which equals five full orbits at $T = 12\,\text{s}$, the MPCBF converged to a stable trajectory after one successful orbit Figure 5.9. The CLF-CBF-QP however, does not converge, with the orange parts of the plot denoting the parts of the trajectory where the aerial robot would have collided with the obstacle. If anything, the trajectory seems to be unstable from time to time. For example at $t = 12\,\text{s}$ to $24\,\text{s}$ and $t = 38\,\text{s}$ to $54\,\text{s}$. Note that this is roughly the same time interval with respect to $T$, still, in both cases the trajectory is different and the third orbit is rather stable and close to the desired trajectory.

Looking at a top view in Figure 5.10 of both trajectories fortifies this discrepancy even more. The MPCBF's trajectory is always close to the circle, the CLF-CBF-QP however drifts away multiple meters. There is still no pattern that could predict how the CLF-CBF-QP would behave on the next orbit. The reason behind this behavior has to do with infeasible solutions. As mentioned before, whenever there occurred an infeasible solution, or more precisely an infeasible constraint, this constraint was removed and the optimizer was run again. Now this is clearly not a desired scenario in real flights, nevertheless it allows the simulation to continue and to observe the recovering behavior of the CLF-CBF-QP. Evidently, recovering from a situation in which the CLF-CBF-QP does collide with an obstacle, appears to not be an easy task for the CLF-CBF-QP. Looking at Figure 5.9, where a collision can be observed at approximately $t = 40\,\text{s}$, the recovery maneuver brings the aerial robot down to almost $z = -50\,\text{m}$. As the behavior of the controller for a situation where $\boldsymbol{x} \notin \mathcal{C}$ is not defined, reaching the border of $\mathcal{C}$ or even crossing it, will leave the controller in an unstable situation. This is, what can be observed throughout this chapter whenever the aerial robot is inevitably colliding or has already collided with an obstacle using the CLF-CBF-QP.

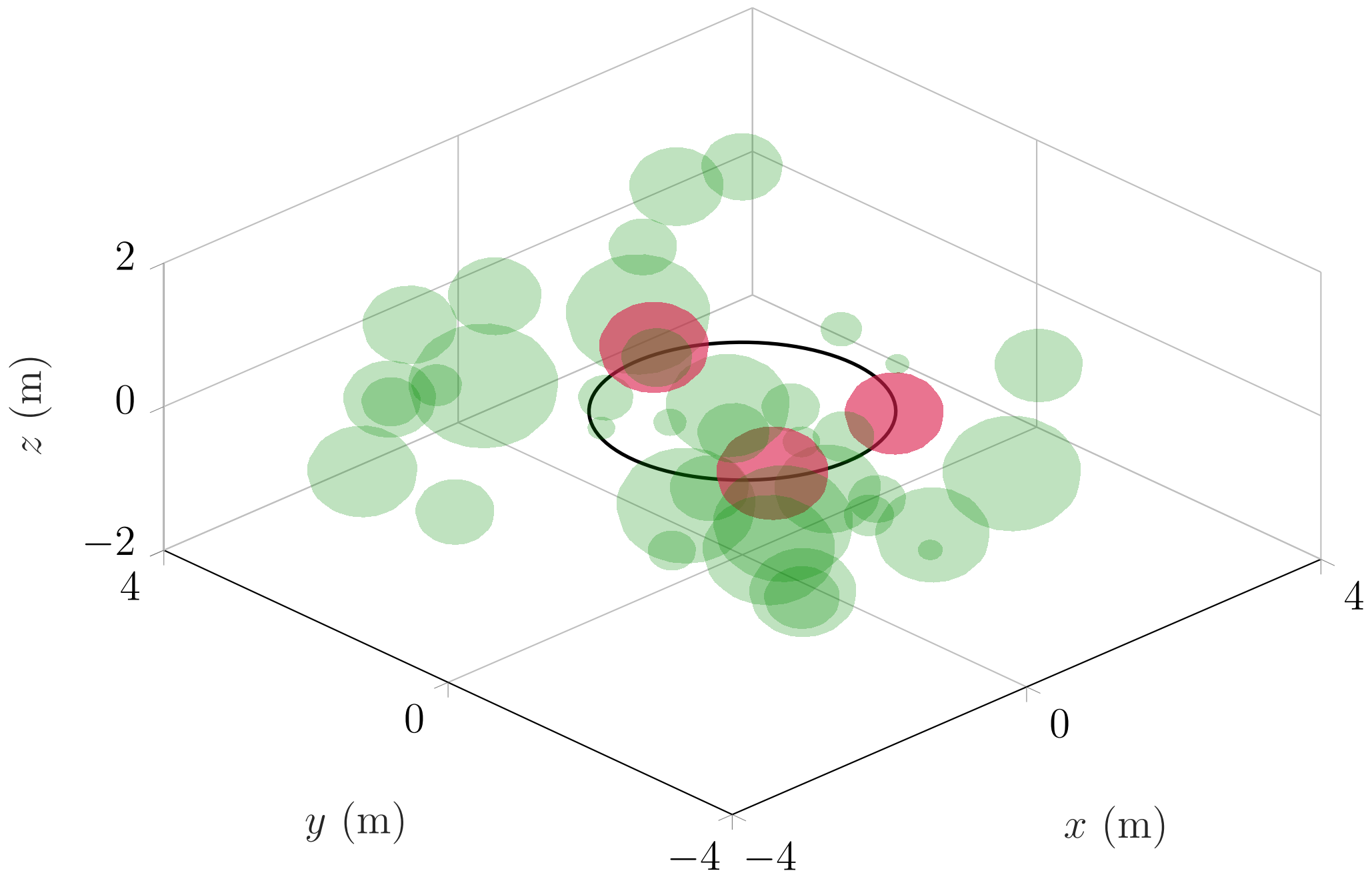

Figure 5.8: Position of the 40 static obstacles with the desired trajectory.

The total mean error, Equation 5.8 for the MPCBF adds up to $e_{\text{mean,MPCBF}} = 1.33\,\text{m}$, compared tot the total mean error for the CLF-CBF-QP of $e_{\text{mean,CBF}} = 12.23\,\text{m}$, which is greater than $e_{\text{mean,MPCBF}}$ by a factor of 9.22. Comparing the first 12 s, the first orbit, reduces the difference to a factor of 3.66, with $e_{\text{mean,MPCBF}} = 1.34\,\text{m}$ and $e_{\text{mean,CBF}} = 4.92\,\text{m}$. Similar for the third round, $t = [24\,\text{s}, 36\,\text{s}]$, the difference is $e_{\text{mean,MPCBF}} = 1.32\,\text{m}$ compared to $e_{\text{mean,CBF}} = 2.76\,\text{m}$, a factor of 2.08. From that, it is clear that if the CLF-CBF-QP does find a feasible solution, it is not that far off compared to the MPCBF. However, the problem is that it does not always find a feasible solution. As discussed, this can have multiple roots. Mostly, that it is already too close to an obstacle or that the relative velocity is too high, thus an avoidance maneuver is no longer possible.

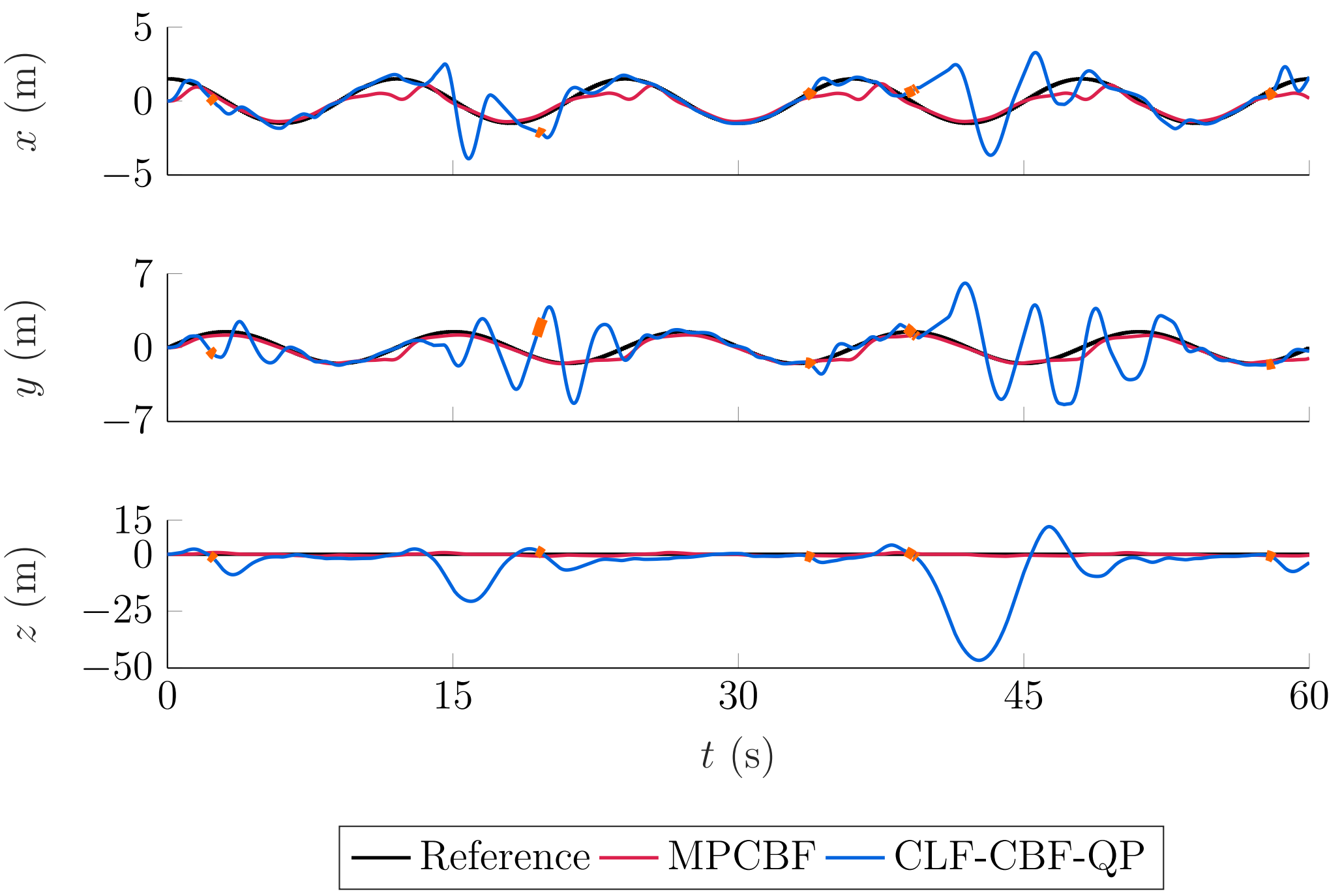

Figure 5.9: Position of the aerial robot for a circular trajectory in a highly cluttered environment, collisions are marked in orange.

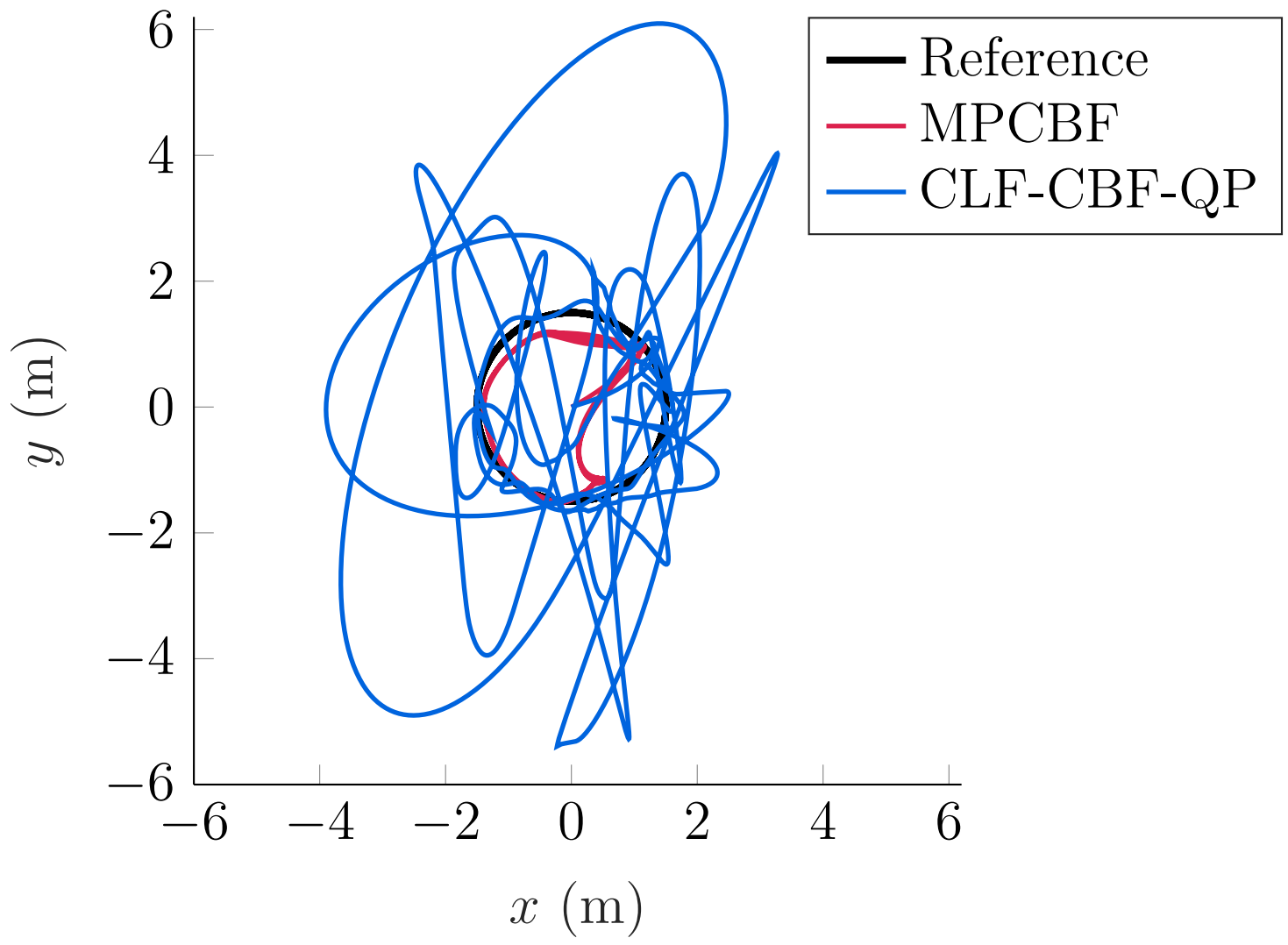

Figure 5.10: Top view of the aerial robot's trajectory.

Looking at screenshots from the flight at different times for the controllers reveals the differences between the controllers. In general, the MPCBF does follow the reference trajectory closely, whichclearly visible by the predicted trajectory in Figure 5.11c, where the predicted trajectory is shown in blue. Moreover, the screenshot from $t = 1.3\,\mathrm{s}$, see Figure 5.12a, shows the superior behavior of the MPCBF as it can predict its trajectory around the obstacle and therefore act on it earlier, compared to the CLF-CBF-QP. It also shows, that while the MPCBF avoids the obstacle, its predicted trajectory converges towards the reference trajectory as soon as it passes the obstacle. Nevertheless, at certain positions and times, it is trapped, see Figure 5.11b. With a longer prediction horizon, the controller would realize, that being trapped and hovering at a certain sport is by far not the optimal trajectory. However, for that brief moment with a limited prediction horizon it seems to be the trajectory which minimizes the distance to $x_{t,\mathrm{ref}}$ and also the control outputs.

The CLF-CBF-QP on the other hand clearly violates its objective at $t = 2.4\,\mathrm{s}$, see Figure 5.12a, and $t = 19.8\,\mathrm{s}$, see Figure 5.12c. At these time steps, the aerial robot clearly collides, or did so, with the surrounding obstacles. However it only does so, because the infeasible constraints where removed. As those constraints belong to the obstacles the aerial robot collided with, a violation of the objective is the only possibility to regain a feasible solution. At $t = 11\,\mathrm{s}$, see Figure 5.12b, it is visible, that the controller does a better job flying around the obstacles compared to the MPCBF, see Figure 5.11b. But only shortly after that at $t = 19.8\,\mathrm{s}$, the CLF-CBF-QP appears to lose control over the aerial robot, as it has to compensate for a huge disturbance of unknown cause.

Summarizing, the MPCBF is able to find a local optimal solution that is closer to the global optimum and is, in general, more reliably avoid the obstacles in its path. Nevertheless, this could already have been conjectured, as the MPCBF optimizes over a longer time horizon, namely more than the one step which is CLF-CBF-QP's time horizon. Regardless, [WuEtAl16a] proofed that the CLF-CBF-QP does indeed promise to stay within $\mathcal{B}_{k|t}$. The problem arising here, is that it is fairly simple to create situations that do not allow for a feasible solution of the optimization problem and therefore forward invariance of $\mathcal{C}$. Being able to always find a feasible solution was the premise, [WuEtAl16a] built their proof upon. Finally, the shown situations, while not taken from real experiments, in fact represent real world scenarios and are not specifically constructed to disprove [WuEtAl16a].

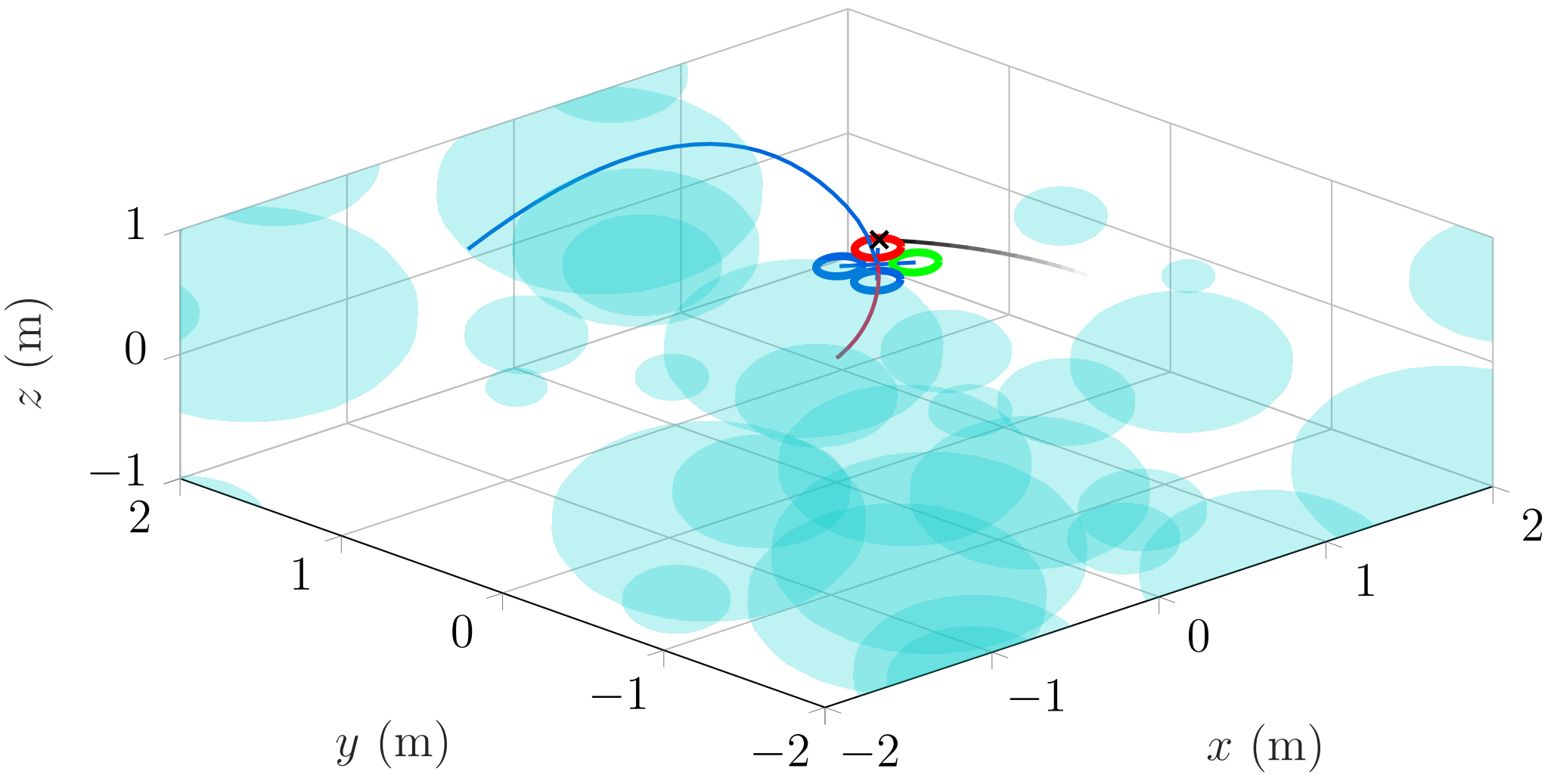

(a) $t = 1.3\,\mathrm{s}$

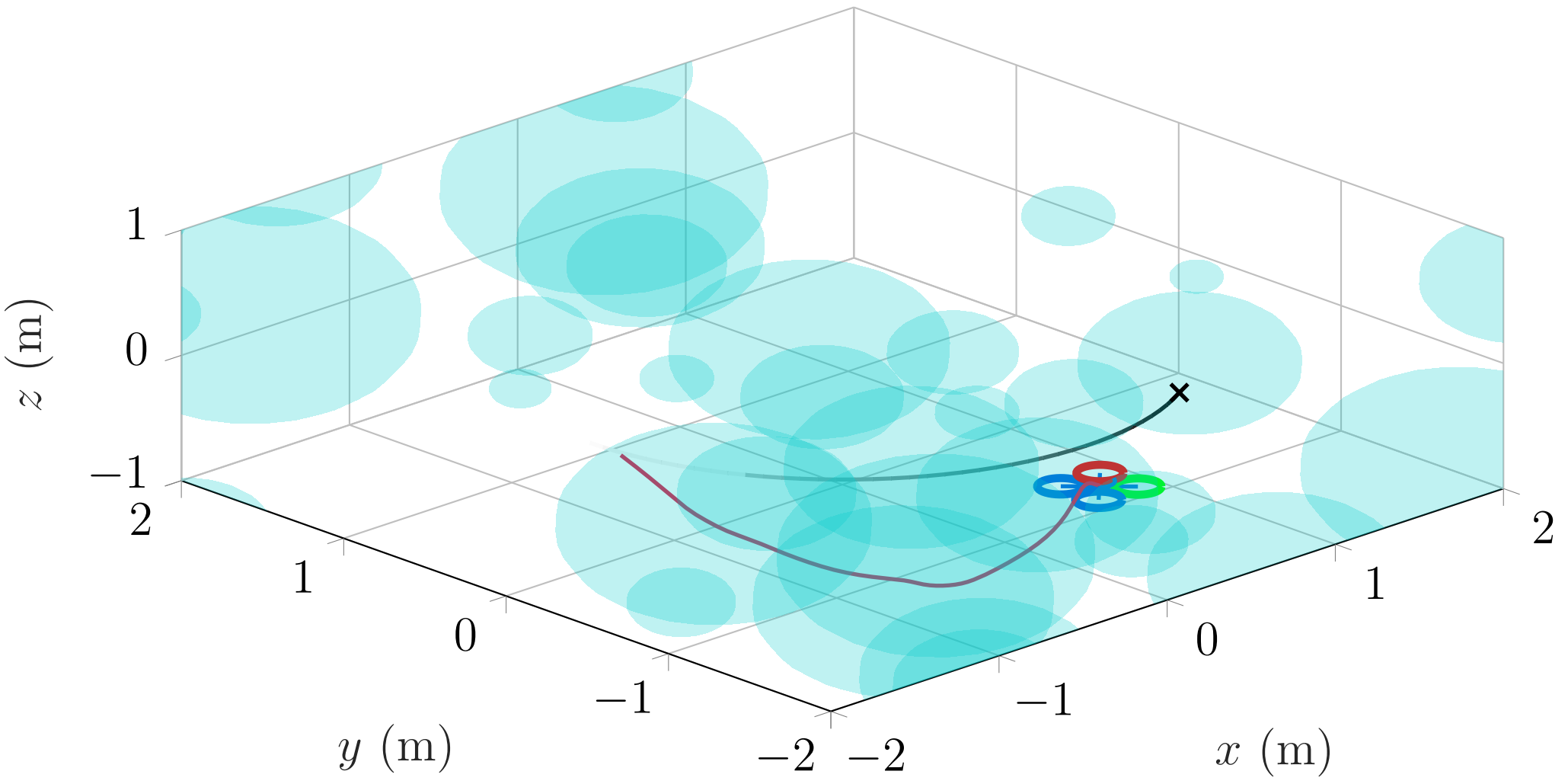

(b) $t = 10\,\mathrm{s}$

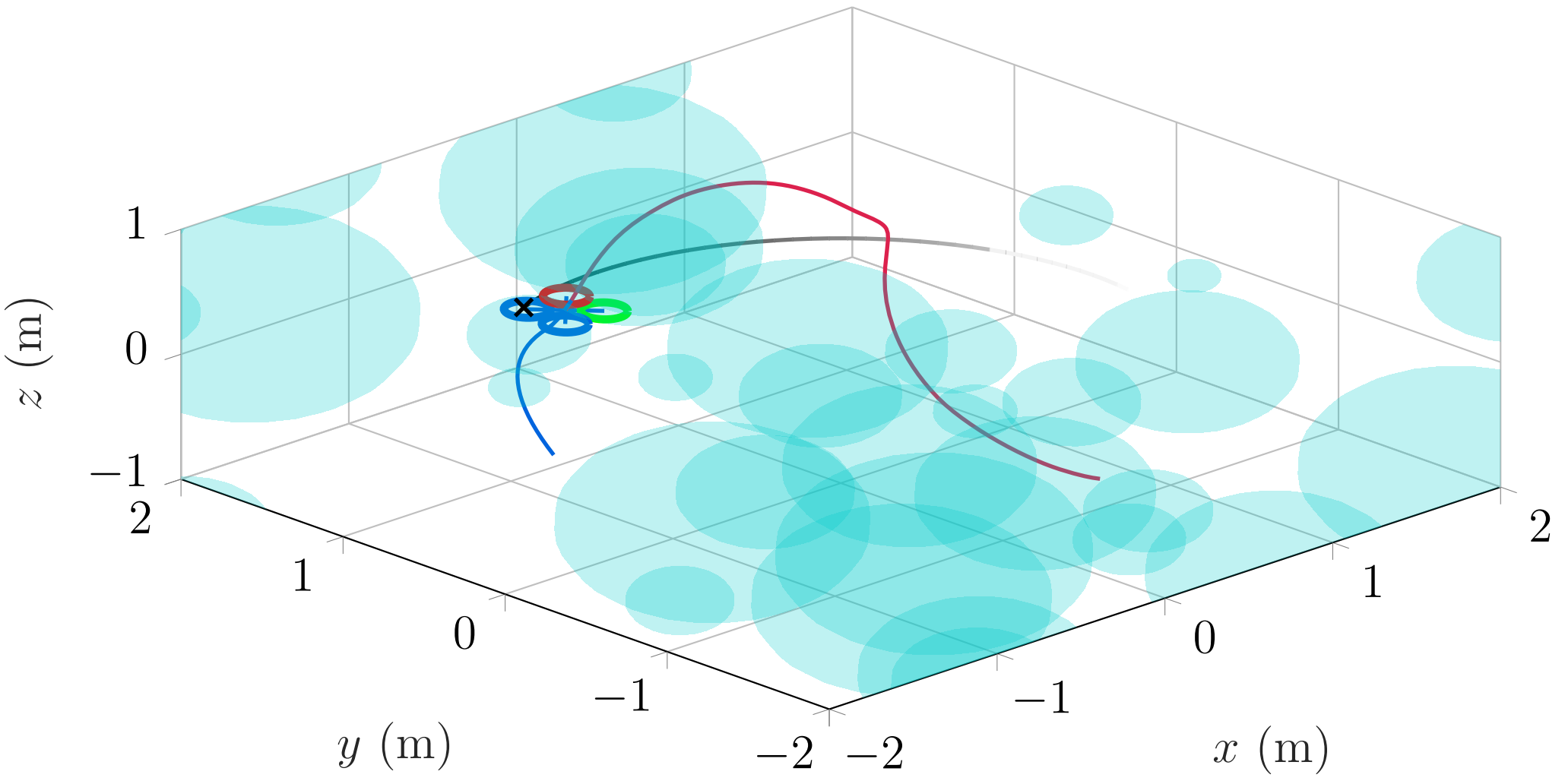

(c) $t = 15.7\,\mathrm{s}$

Figure 5.11: Different screenshots of the MPCBF.

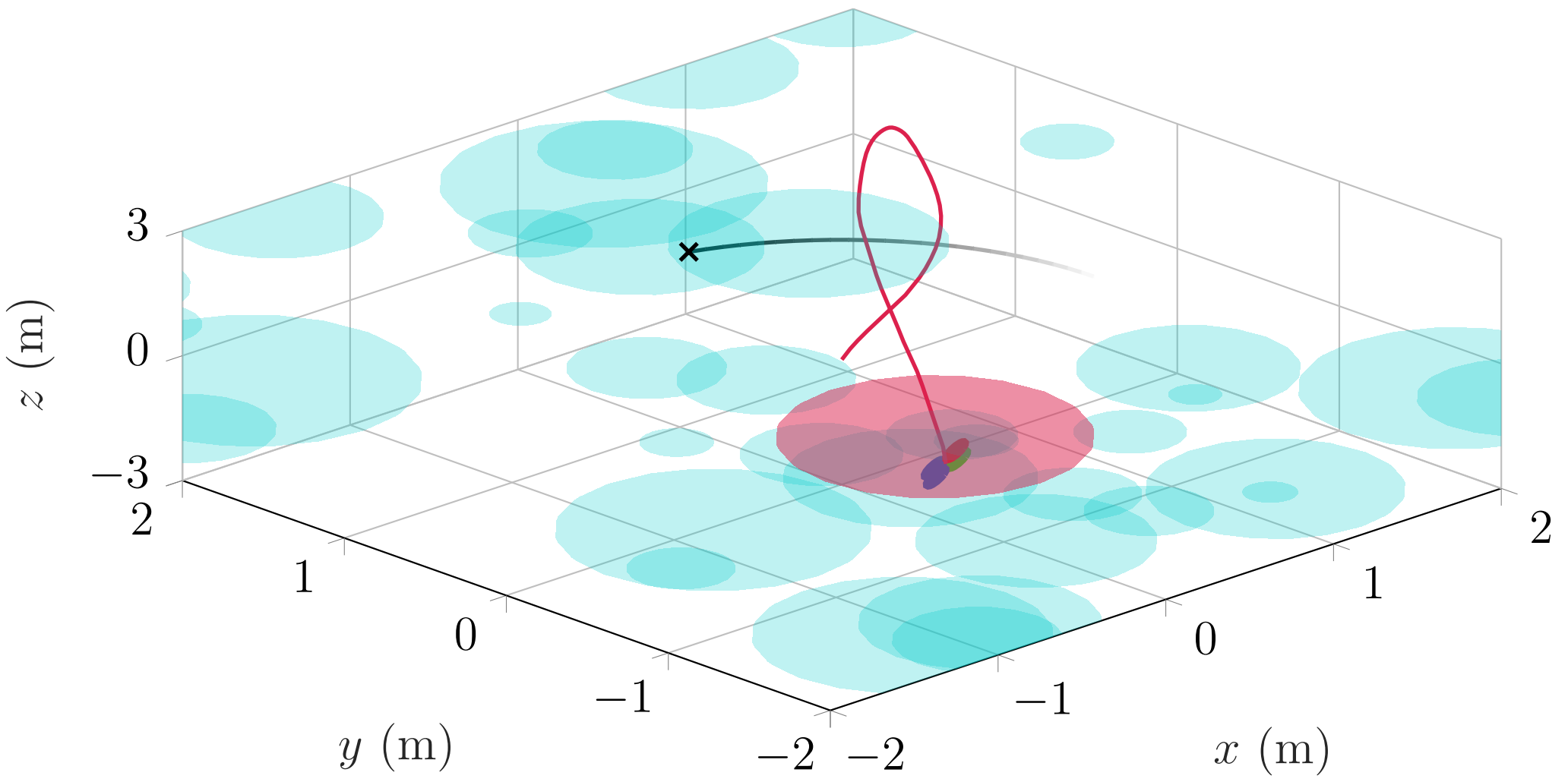

(a) $t = 2.4\,\mathrm{s}$

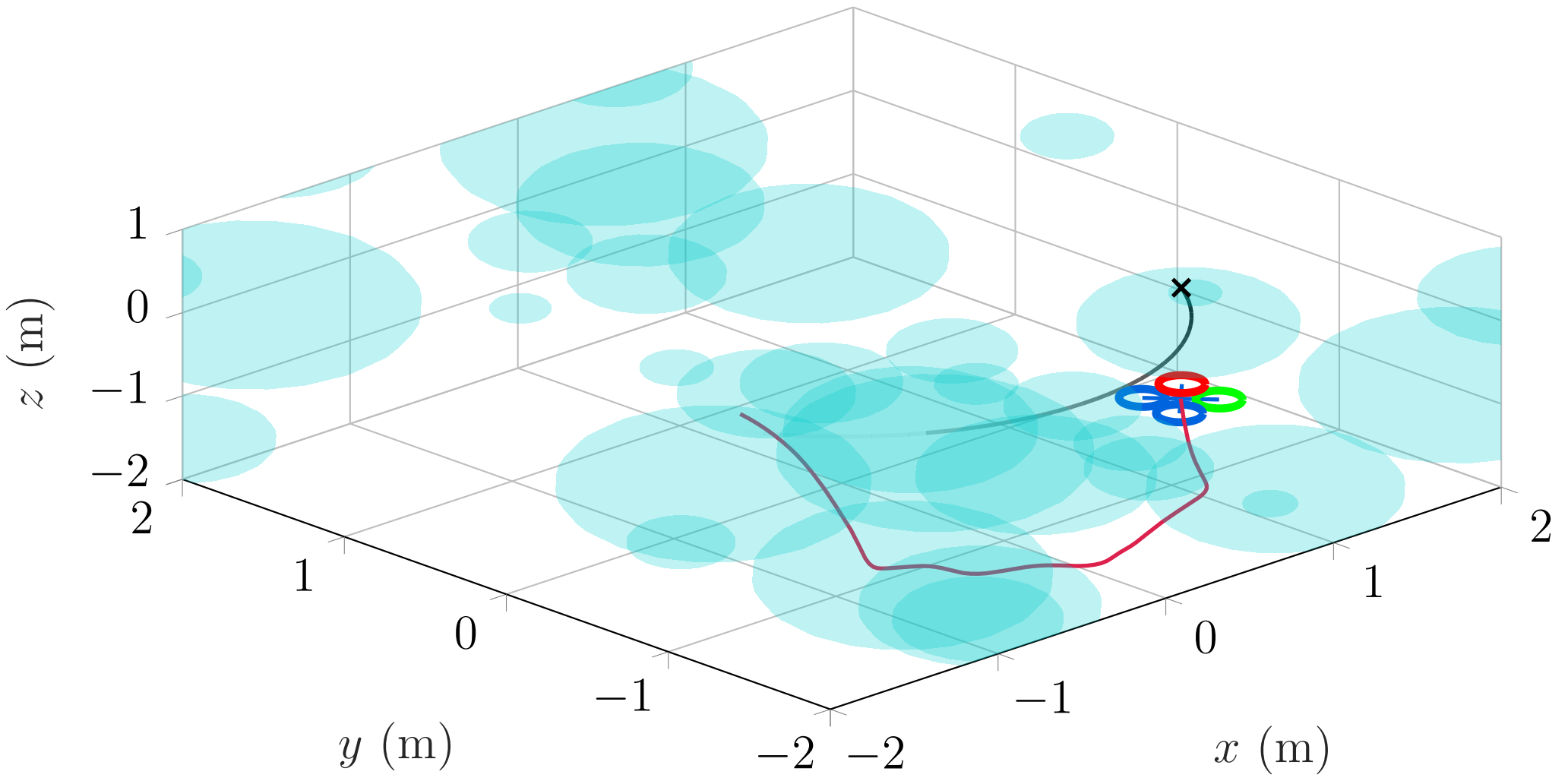

(b) $t = 11\,\mathrm{s}$

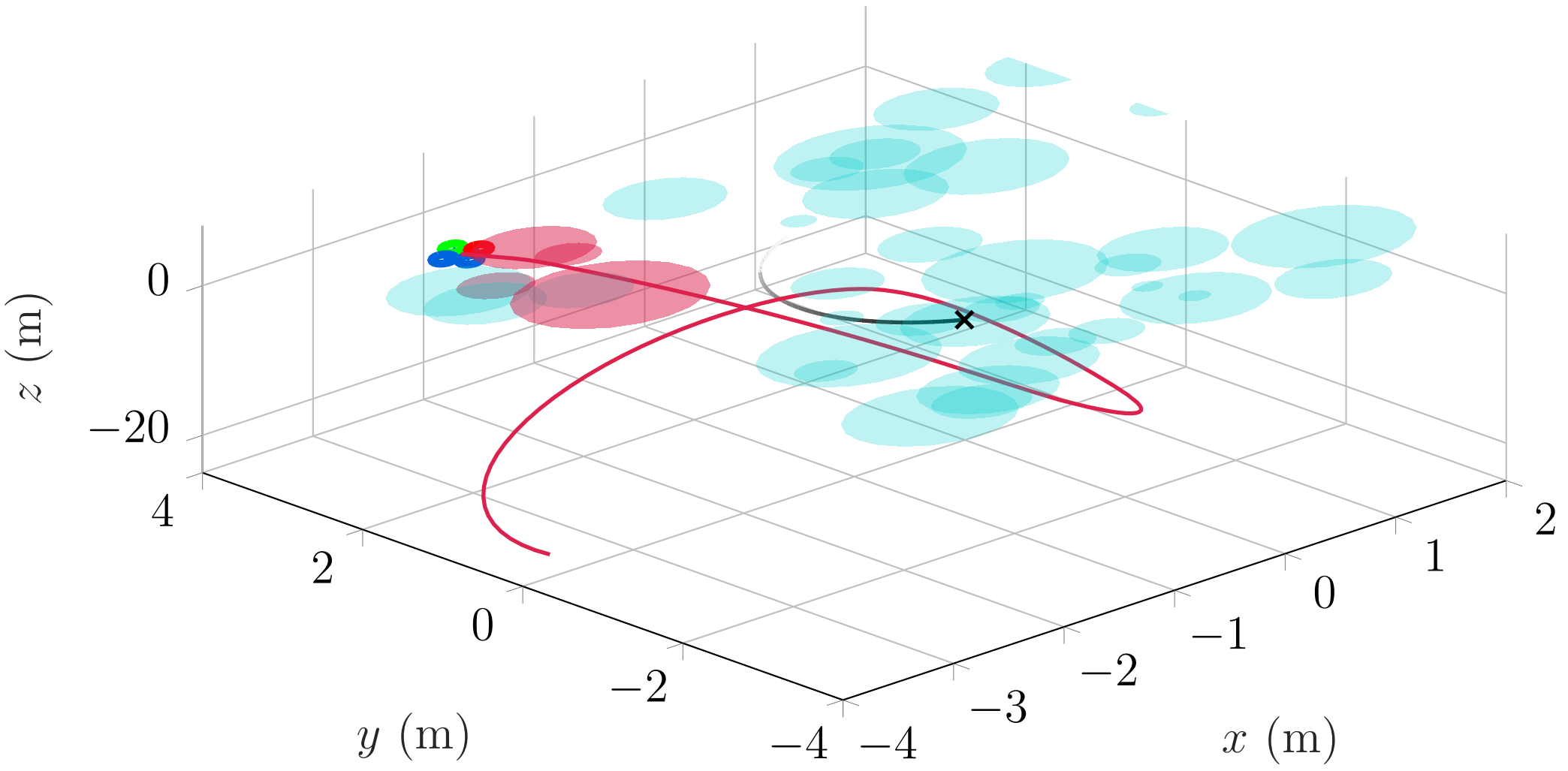

(c) $t = 19.8\,\mathrm{s}$

Figure 5.12: Different screenshots of the CLF-CBF-QP.

## 5.2 Flight Tests

The flight tests were conducted in Prof. Mark Mueller's flight space at the University of California, Berkeley. As mentioned in chapter 3, the dimensions of the flight space are approximately $3\,\mathrm{m} \times 5\,\mathrm{m} \times 6\,\mathrm{m}$ ($x \times y \times z$). To first validate the overall performance of the aerial robot, the Euler Angle controller is used. The results of the robot's hovering are discussed in subsection 5.2.1. Tests using the CLF-CBF-QP were conducted as well. However, the aerial robot was not stable and would crash after a maximum of around 5 s in-flight. The high fluctuations of the control output made it impossible for the motor to follow the reference signal. Furthermore, the MPCBF's behavior in both simple tracking missions, as well as for obstacle avoidance, are both discussed in subsection 5.2.2. The same gains were used as in the simulation, see Table 5.1.

### 5.2.1 Euler Angle Control

The first experiment was conducted to validate the system itself and the Euler Angle controller. A constant setpoint of $\boldsymbol{x}_{\mathrm{ref}} = [0\,\mathrm{m}, 0\,\mathrm{m}, 1.25\,\mathrm{m}]^{\mathsf{T}}$ was commanded to the position controller and a zero Euler Angle setpoint for the attitude controller. Starting at $\boldsymbol{x}_0 = [0.34\,\mathrm{m}, -1.54\,\mathrm{m}, 0.45\,\mathrm{m}]^{\mathsf{T}}$, the controller is able to stabilize the system but with a constant error. This is expected, as the controller lacks I-gains. Nevertheless, the results shown in Figure 5.13 clearly validate that the designed and built system is capable of flying and the Euler Angle position and attitude controller is able to stabilize the system. While the $x$- and $y$-positions have a constant error, the $z$-position starts off at exactly the reference position, but soon looses height. This is due to the way the PWM signal is calculated. There exists a precise mapping from thrust to PWM signal, which is depended on the battery level. The corresponding errors for $t = 10\,\mathrm{s}$ to $250\,\mathrm{s}$ are $e_{\mathrm{mean,EA}} = 0.284\,\mathrm{m}$ with the mean error for each coordinate being $e^{(x)}_{\mathrm{mean,EA}} = 0.173\,\mathrm{m}$, $e^{(y)}_{\mathrm{mean,EA}} = 0.106\,\mathrm{m}$, $e^{(z)}_{\mathrm{mean,EA}} = 0.176\,\mathrm{m}$. If the voltage of the battery is too low, it cannot provide the necessary power, thus, through slip, the rotor speed is slower than anticipated. Nevertheless, the mapping was only created for a specific battery level, thus, for all other lower levels, the error is non-zero. To fix this problem, there are three possible solutions. First, create a mapping for each voltage level of the battery. Second, add an active feedback from the motor with an additional control loop. Lastly, an integrative term in the position controller can also fix the problem.

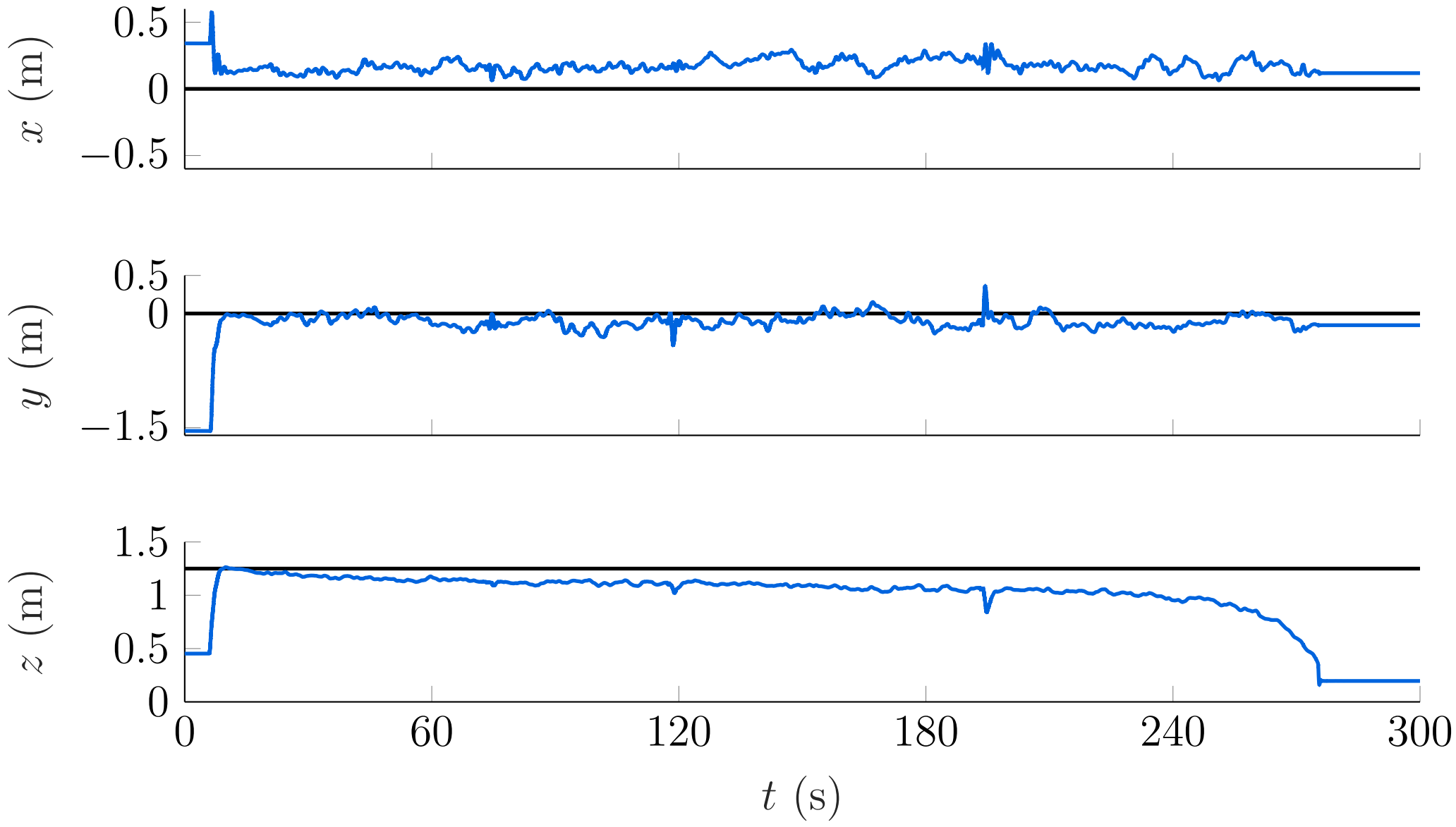

Figure 5.13: Experimental result for the Euler Angle controller given a constant reference setpoint.

### 5.2.2 MPCBF

Given the validated platform, the MPCBF is tested for its general stability and reference tracking, as well as for its obstacle avoidance behavior. Similar to the experiments with the Euler Angle controller, the MPCBF is first tested with a constant setpoint $\boldsymbol{x}_{\text{ref}} = [0\,\text{m}, -1\,\text{m}, 1.5\,\text{m}]^\mathsf{T}$. The results are shown in Figure 5.14. Again, there is a constant drop in $z$ which correlates to the battery voltage and therefore the remaining charge. Although all experiments were started with a fully charged battery, the charge was quickly dropping due to the high power consumption of the motors, a general problem for aerial robots. Overall, the mean error for $t = 20\,\text{s}$ to $380\,\text{s}$ is $e_{\text{mean,MPCBF}} = 0.109\,\text{m}$ with the mean error for each coordinate being $e^{(x)}_{\text{mean,MPCBF}} = 0.055\,\text{m}$, $e^{(y)}_{\text{mean,MPCBF}} = 0.040\,\text{m}$, $e^{(z)}_{\text{mean,MPCBF}} = 0.070\,\text{m}$. Comparing the experiments from Figure 5.13 and Figure 5.14, it follows, that the MPCBF in fact performs better than the Euler-Angle. The mean error for all three axis are less than half of that from the EA. Therefore, the predictive controller's performance is superior over the EA-controller's performance. Furthermore, Figure 5.14 shows, that the MPCBF can cope with the disturbances in the real system. It does not react too aggressively which could lead to instabilities, and still fast enough to reduce the disturbances and bring the system back closer to the reference value. The global exponential stability of the CLF-constraint, see Equation 4.90, cannot be verified from the results. Judging from the results, the problem seems to be the inevitably occurring dis-turbances, which hinder the global stability. Still, the CLF-constraint contributes to the overall stability and the lower mean error compared to the Euler-Angle controller.

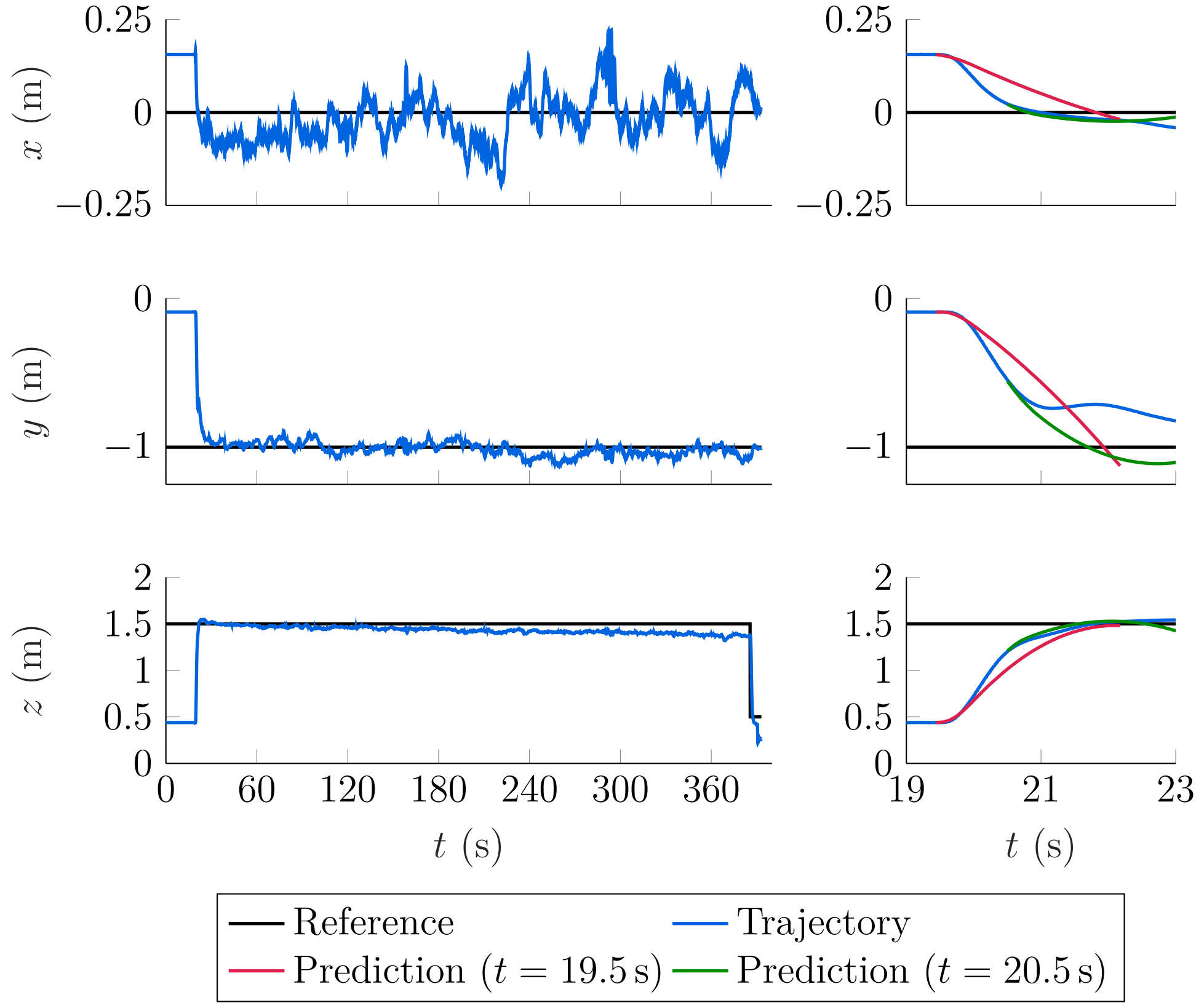

Figure 5.14: Experimental result for the MPCBF given a constant reference set-point without any obstacles.

The right hand side of Figure 5.14 also shows the prediction at two different time steps. The first is when the aerial robot is armed and starts its flight, at around $t = 19.5\,\mathrm{s}$, and the other one is one second later, at $t = 20.5\,\mathrm{s}$, around halfway before settling. Focusing only on the $x$ and $y$ plots, there is a mismatch between the predicted trajectory and the real trajectory. The mismatch is most likely due to a model-mismatch, but an insufficient tuning could also contribute. Especially, as the second prediction for $x$ is already fairly close to the real trajectory. For $y$, the second prediction is close to the desired behavior, however, the trajectory shows a slight disturbance which happens at around $t = 21\,\mathrm{s}$. For $z$, both predictions are in fact not far off compared to the real trajectory, as both clearly converge towards the reference trajectory. In general, it seems like with the first prediction, the MPCBF tries to resist the change in position, because of the high weight on $\dot{\boldsymbol{x}}$. However, for smaller weights on the velocity, the dampening of the MPCBF is not sufficient to stabilize the aerial robot. For later stages of the experiment, the predicted trajectory closely resembled a low-pass behavior towards the reference trajectory. With the ever increasing error along $z$, the desired thrust also increased. While the MPCBF was was able to further reduce the error compared to the Euler-Angle, it still was not able to fully compensate for the battery drainage. Again, the observed mean error along the $z$-axis is $e^{(z)}_{\text{mean,MPCBF}} = 0.070\,\text{m}$ for the MPCBF compared $e^{(z)}_{\text{mean,EA}} = 0.176\,\text{m}$ for the EA. But as the MPCBF has no integrative behavior and no model for the battery drainage, it cannot compensate for the real thrust being lower than the desired thrust.

For the second experiment, a circular trajectory was commanded with $r_\text{c} = 1\,\text{m}$ and $T = 30\,\text{s}$, the experiment was run up until around $t = 160\,\text{s}$ at which the aerial robot was disarmed and safely fell into the safety net. The center of the circle was set to $\boldsymbol{x}_\text{c} = [0\,\text{m}, 0\,\text{m}, 1\,\text{m}]^\mathsf{T}$ and changed at $t = 85\,\text{s}$ to $\boldsymbol{x}^{(z)}_\text{c} = 1.5\,\text{m}$. Moreover, a fixed virtual obstacle was added at $\boldsymbol{x}_\text{obs} = [0.87\,\text{m}, 0.5\,\text{m}, \boldsymbol{x}^{(z)}_\text{c}]^\mathsf{T}$, with $r_\text{obs} = 0.25\,\text{m}$. Thus, the obstacle would always stay at the same height as the aerial robot is supposed to be. Not only did the MPCBF achieve close reference tracking, but it also avoided the obstacle every time.

In Figure 5.15 it is clearly visible, that the MPCBF deviates from the reference trajectory to avoid the obstacle. It can also be seen that there are always three plummets in the $x$ and $y$ plot, where the aerial robot deviates from its previous trajectory but quickly recovers. The general appearance of those is similar for each avoidance maneuver which allows the conclusion, that those plummets are systematic. As it turns out, with the average speed along the trajectory, the time needed to pass the diameter of the obstacle is roughly 2.4 s and the time to fly exactly along the obstacle is 7.7 s. Thus, the MPCBF cannot predict the complete trajectory around it in the beginning. The short prediction horizon is most likely the reason for the plummets.

To further visualize the result, the trajectory of the robot is plotted in 3D in 5.16. Here, only the time range of $t = [24.5\,\text{s}, 144.5\,\text{s}]$ is shown, when the aerial robot flew four full rotations. The two spherical obstacles are in fact the same obstacle at different times in the experiment and therefore different heights to match the reference height of the aerial robot. For the most part, the aerial robot follows the trajectory closely, however, while safely avoiding the obstacle, there is a huge deviation from the trajectory. The average of the maximum distance between the aerial robot and the obstacle for each maneuver is 1.2 m. With $r_\text{qr} = 0.265\,\text{m}$, $r_\text{safety} = 0\,\text{m}$, and $r_\text{obs} = 0.25\,\text{m}$, this is approximately 0.69 m too far off. The deviation can again be influenced with more precise tuning of the CBF constraint's parameter $\gamma$, see Equation 4.100.

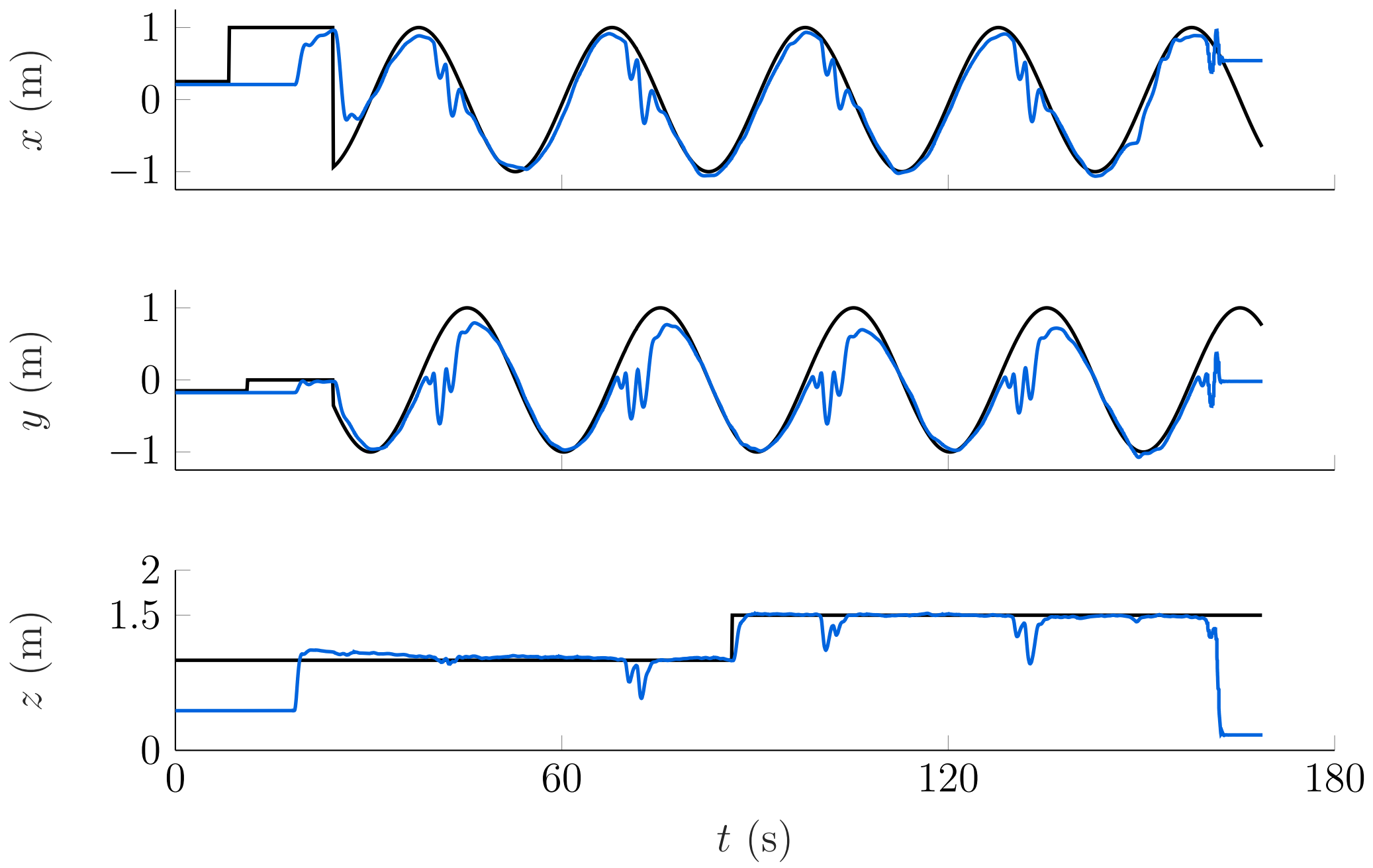

Figure 5.15: Experimental result for the MPCBF given a circular trajectory with an obstacle with fixed $x$ and $y$ position and the $z$ position equal to $\boldsymbol{x}_{\text{ref}}^{(z)}$.

The third experiment was performed with a faster circular trajectory with $T = 15\,\text{s}$ and $r_{\text{c}} = 1\,\text{m}$, $\boldsymbol{x}_{\text{c}} = [0\,\text{m}, 0\,\text{m}, 2\,\text{m}]^{\mathsf{T}}$, still counterclockwise. This time, the obstacle follows the exact same circular trajectory but instead of going counterclockwise, it moves clockwise around $\boldsymbol{x}_{\text{c}}$. Thus, both the aerial robot and the obstacle meet twice for each full circumvolution. Moreover, the obstacle's radius was reduced to $r_{\text{obs}} = 0.125\,\text{m}$. By doing so, the prediction horizon can predict a trajectory, which avoids the obstacle, as the time it would take the aerial robot to fly around it is closer to the prediction horizon. Before, with $T = 30\,\text{s}$ and $r_{\text{obs}} = 0.25\,\text{m}$, the shortest path around the obstacle is $1.62\,\text{m}$ long, with an average speed of $0.21\,\frac{\text{m}}{\text{s}}$, this would roughly take $7.7\,\text{s}$. Now, the shortest path around the obstacle is reduced to $1.23\,\text{m}$ and with an increased average speed of $0.42\,\frac{\text{m}}{\text{s}}$ the time it takes to fly around the obstacle is now $2.93\,\text{s}$. Thus, it should be easier for the MPCBF to find a collision-free trajectory.

The resulting trajectory is shown in Figure 5.17. Again, the two instances along the circle where the aerial robot has to avoid the obstacle are clearly visible. Interesting enough, the avoidance maneuver on the right hand side of the plot shows, that the aerial robot has to fly backwards quite a bit. Whereas the maneuver on the left hand side appears to be smoother. For this maneuver, the MPCBF avoids the obstacle safely at each fly-by. For the other encounter, namely on the right hand side, it does not. Each encounter results in a collision, which is shown for different time steps in Figure 5.18. What happens is, that the MPCBF controls the aerial robot in such a way, that it at first collides with the obstacle, from which it then recovers, resulting in the backwards flight. Therefore, the controller has to again avoid the obstacle, which it safely does on the second try. In Figure 5.18a for $t = 66\,\mathrm{s}$, the MPCBF predicts a trajectory, that deviates ever so slightly to the right to avoid the obstacle. But only 0.5 s later, in Figure 5.18b, the aerial robot collides with the obstacle. It appears as if the MPCBF misjudged the obstacle's relative velocity and is not deviating enough. As it collides with the obstacle, the MPCBF will still try to increase its distance to the obstacle and thus accelerate away from the obstacle as fast as possible, leading to the observed backwards motion of the aerial robot. Since the obstacle is not a real physical object, the collision is purely virtual, hence, the MPCBF can continue its flight. In Figure 5.18c, yet another 0.5 s later, it appears as if the aerial robot recovered from the collision and is now predicting a trajectory that would guide it safely around the obstacle. While it does avoid the obstacle, judging from the sequential three plots, Figures 5.18d, 5.18e and 5.18f, it is visible, that the MPCBF rapidly changes its predicted trajectory and, therefore, far off from the closed-loop trajectory. The pattern is repeated for all five encounters of the aerial robot with the obstacle. Here, the predicted trajectory in Figure Figure 5.18d for $t = 67.5\,\mathrm{s}$ appears to steer the aerial robot even further backward to avoid the obstacle.

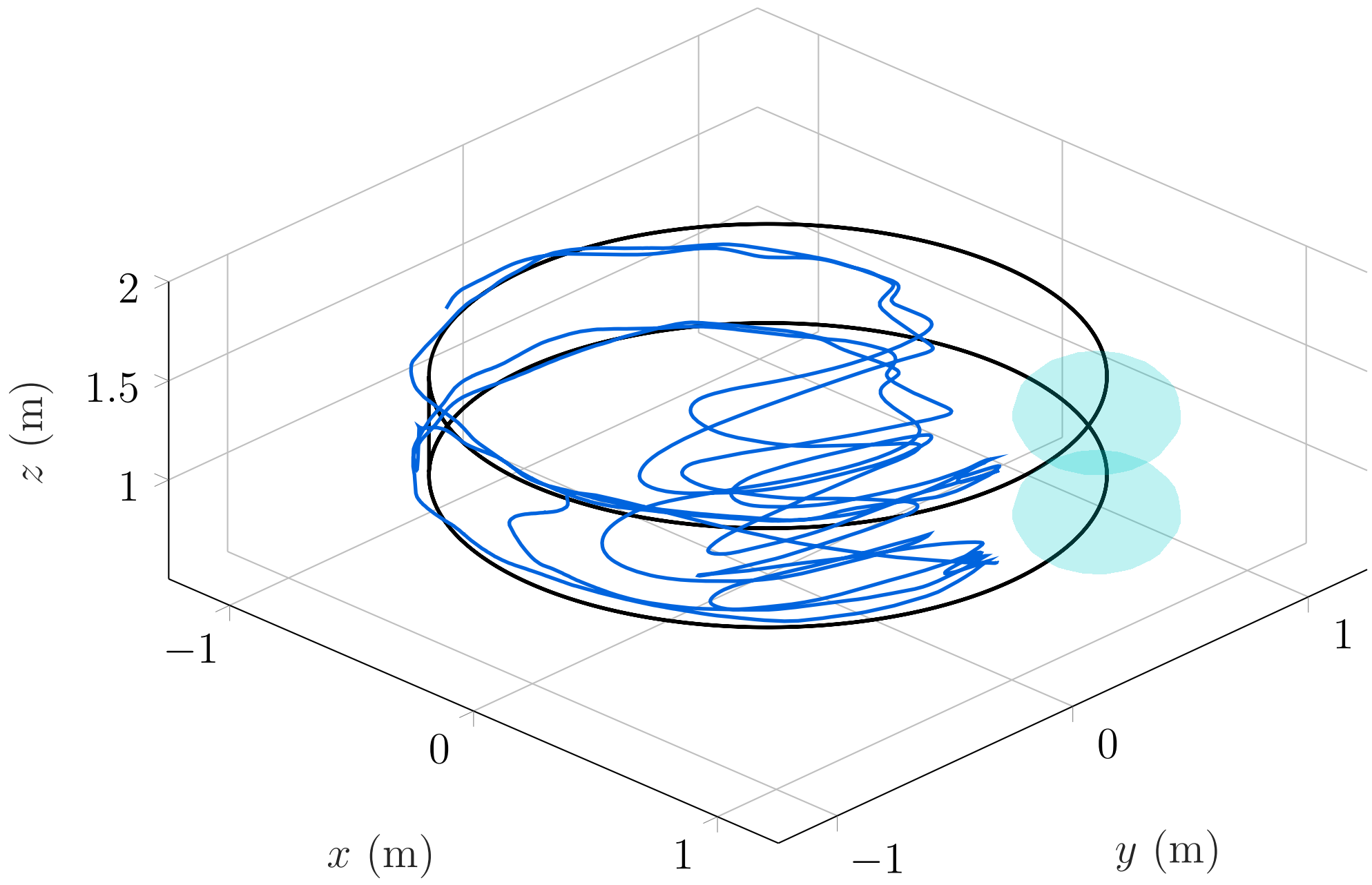

Figure 5.16: 3D view of Figure 5.15 for $t = [24.5\,\text{s}, 144.5\,\text{s}]$.

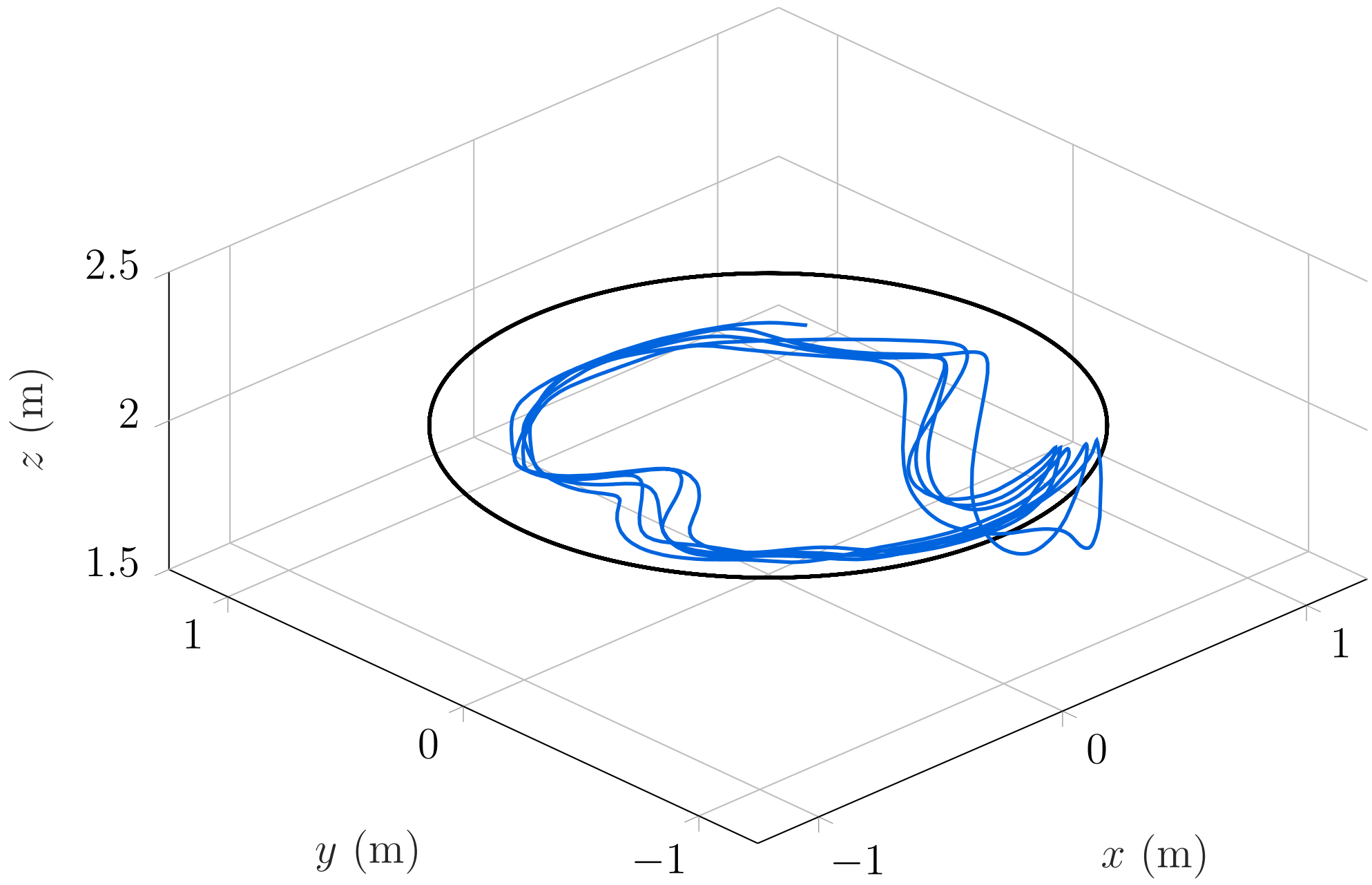

Figure 5.17: 3D view of the aerial robots trajectory for $t = [55\,\text{s}, 130\,\text{s}]$.

At this point, it is not clear, as to why the avoidance algorithm fails, especially in such a repetitive way. Oddly, the initial condition for both the left hand side and right hand side encounter are similar, just rotated by 180°. The only direct asymmetry in the MPCBF and its constraints is the box constraint, but as the reference trajectory fully lies inside the box and the controller can always deviate towards the center, it cannot be the cause of the problem. It could be, however, that the control horizon $N_\mathrm{c}$ is too small, but again, in the experiments before and for the other encounter, the MPCBF works fine. While the predicted trajectory is optimized over 2.5 s, the control output is only optimized over 0.5 s after which the last control output is held for the rest of the prediction. This way, the computational requirements can be drastically reduced. With a longer control horizon, it is potentially easier for the MPCBF to optimize the flight path, especially around an obstacle. But an increased control horizon increases the solving time and with a larger number of obstacles, even Cvxgen does not run fast enough. Yet another reason might be the choice of $\gamma_i$. The higher $\gamma_i$ is, the more aggressive the CBF constraint is. Nevertheless, even with a very small but positive $\gamma_i$, the CBF constraint should always force the controller to avoid the obstacle. And as $\gamma_i$ is constant for the flight, it does not explain the asymmetric behavior. Independent of the cause of the problem, the MPCBF should not collide with the obstacle due to the strong CBF constraint. But again, as before in the simulations, the constraint appears to cause infeasibility in special circumstances resulting in the violation of the assumption for the forward invariance of $\mathcal{C}$. As described, the CBF only guarantees safety, as long as $\mathcal{C}$ is forward invariant, which is equal to the optimizer finding a solution to the optimization problem at all times.

Another discrepancy is, that the constant error between the flight path and the reference trajectory is greater compared to the previous experiment. A similar behavior has been observed in Figure 5.4, where a reduction of the time period lead to an increase in the constant error. Nevertheless, the cause for the rising error for the current situation is suboptimal tuning. Due to the smaller radius and time period, the average velocity increases and thus the acceleration, which in turn results in higher necessary forces to keep the aerial robot on track. As the MPCBF optimizes both, the error and the control output, the weights on the error would have to be increased and the weights on the control outputs decreased. However, the new set of parameters might result in worse reference tracking for other missions.

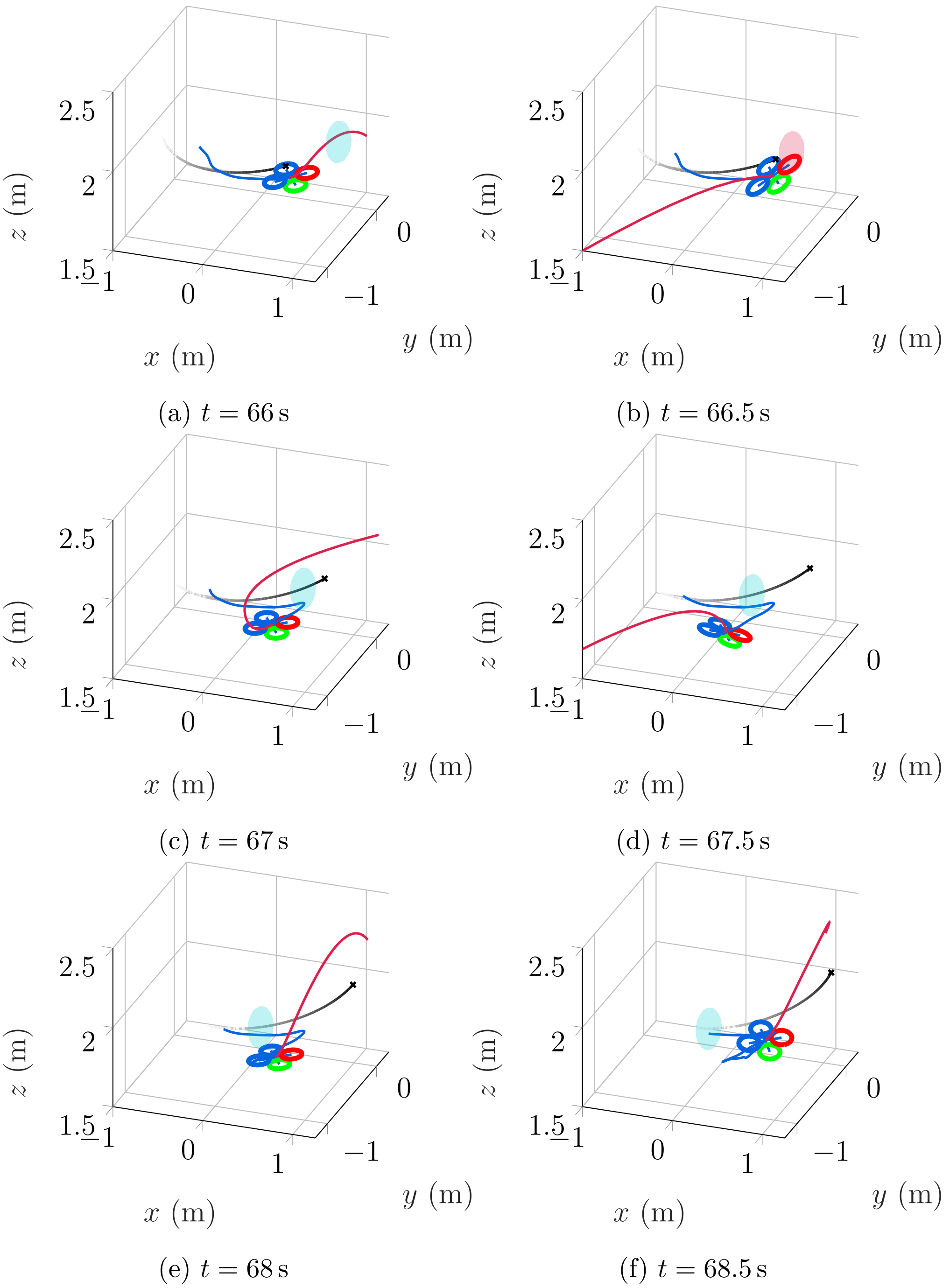

(a) $t = 66\,\mathrm{s}$ (b) $t = 66.5\,\mathrm{s}$ (c) $t = 67\,\mathrm{s}$ (d) $t = 67.5\,\mathrm{s}$ (e) $t = 68\,\mathrm{s}$ (f) $t = 68.5\,\mathrm{s}$

Figure 5.18: Exemplaric avoidance maneuver with the MPCBF.

# Chapter 6

# Summary

This chapter summarizes the work in this thesis. It points out what has been achieved but also the problems which have not been solved. First, the work in this thesis is summed up. Secondly, the contributions of this thesis are highlighted and lastly, an outlook is given.

## 6.1 Conclusion

In this work, an aerial robot platform for obstacle avoidance has been developed, with computationally expensive controllers in mind. The platform was built around a Jetson TX2, providing enough computational power for optimal control strategies. The most important contribution, however, is the proposed MPCBF controller. It combines the obstacle avoidance algorithm of a CBF with the ability to predict ones future states from a MPC. Both the platform itself and the MPCBF have been tested extensively. The results proof that the MPCBF is able to avoid obstacles while tracking a reference trajectory, all while running in real-time on the platform.

In the following, the conclusions for the different topics of this thesis as well as solved and unsolved problems are summarized. Whenever applicable, a possible solution is presented. This can be the starting point of future work.

**Software** The controllers in this thesis are written in C++ for ROS. Due to the modularity of both C++ and ROS, the controllers can be easily adapted to other systems. Conversely, other controllers, written for ROS, can also easily be adapted for the aerial robot platform. To solve the optimization problem, CVXGEN was used. CVXGEN proved to be faster than general solvers and solves the problem in real time. However, it relies on a generated C-code for each problem definition and the capabilities of Cvxgen to generate said code are limited, as it is purely able to solve linear convex problems. Also, a longer prediction horizon might not be possible, if the code generation does not succeed. A longer prediction horizon increases the problem size, as the Cvxgen code can only be generated through their web portal and the maximum time allowed for code generation is 10 min, with an increasing problem size, the time needed rises above the servers limitations. Another problem is, that the maximum number of obstacles to be taken into account has to be defined beforehand. Still, the code can be generated for more obstacles than are present. A possible solution to regain flexibility might be to revert back to a general solver. As these can be slow compared to Cvxgen, it cannot be guaranteed, that general solvers can solve the problem fast enough. Nevertheless, the GPU on the Jetson TX2 could be used to solve the optimization problem. Due to their architecture, GPUs are faster at solving matrix operations compared to CPUs, opening up the possibility to even more computationally demanding optimization problems to be solved on the Jetson TX2's GPU.

**Hardware** To test the proposed MPCBF, the aerial robot platform was designed. It is equipped with a Jetson TX2 for flight control and to provide computation power, an Aerocore 2 as the interconnection between the Jetson TX2 and the ESCs, as well as a camera which could be used for pose information and obstacle detection. The platform is designed around the DJI Flamewheel F330 frame to provide multiple mounting points for all the equipment. All this makes the platform future proof and easy to expand at a later time. While not the first of its kind, the platform provides high computational power in a light-weight design.

The current design of the aerial robot is only equipped with one camera. Therefore, obstacle detection will vary depending on the orientation of the aerial robot. Not all obstacles can be detected at all times and as only the detected obstacles can be avoided, safety might not be guaranteed if the aerial robot is not moving in the direction of the camera or obstacles are moving towards the aerial robot while not in the cameras field-of-view. To eliminate this problem, multiple approaches can be chosen. The aerial robot could, for example, turn around its yaw axis at a constant speed to capture its surroundings. Using a tracking algorithm for the detected obstacles, it might be able to predict their position up until they are back inside the cameras field-of-view. This will, however, reduce the degrees of freedom of the aerial robot, which therefore also imposes a new constraint on the controller. Another solution might be to use multiple cameras. If the sum of the horizontal field-of-view is greater than 360°, the aerial robot can detect obstacles in every direction in its $x$-$y$-plane. The disadvantage would be a higher mass, increased power consumption, and, possibly, a higher strain on the Jetson TX2, depending on the used camera. However, smaller camera designs are available, which are lighter and require significantly less space and power.

One other problem is the dependency of the thrust from the battery charge. While the physical relationship cannot be altered, the future version of the aerial robot platform shall have a mechanism or algorithm to compensate for the lower thrust. Still, for controllers with integrative behavior this is not necessary, but neither the CLF-CBF-QP nor the MPCBF have an integrative behavior.

As of right now, the aerial robot is not completely autonomous. While all control signal are computed on-board, the pose estimation still relies on the motion-capture system. Some early tests promised sufficient pose estimation using the Intel D435i or T265, however, the pose data from the motion capture is more precise and reliable. Also, the camera's depth information has not been used for obstacle detection. But even with these restrictions, the proof of concept and proof of work are achieved.

**CLF-CBF-QP** It has been shown, that the CLF-CBF-QP produces outputs which cannot be followed by the real system. To use the CLF-CBF-QP in a real flight, it has to be modified, to smoothen its outputs. A new and improved Control Lyapunov Function candidate shall include the previous control output and minimize the difference between the previous and the current control output. Using this method, the MPCBF is able to smoothen its control output. The CLF-CBF-QP is not reliable for demanding obstacle avoidance due to the limited prediction horizon of one time-step. Here, the MPCBF can outperform the CLF-CBF-QP due to its longer, yet finite, optimization horizon.

**MPCBF** Throughout this thesis, a new controller, the Model Predictive Control Barrier Function, has been derived. This novel controller combines the forward-invariant safety of Control-Barrier-Functions with the idea of Model Predictive Control, namely to optimize over a finite time horizon instead of optimizing only for the next time step. The resulting trajectory is in general closer to the reference trajectory and the avoidance maneuver results in smaller deviations, compared to the CLF-CBF-QP. The outstanding capabilities of the MPCBF have been proven both in simulation and in real flight tests.

Not only did the MPCBF outperform the CLF-CBF-QP in the simulations, the CLF-CBF-QP is not even able to control the aerial robot platform in real flight tests, whereas the MPCBF is. In some scenarios, the performance of the MPCBF can still be improved, for example for trajectory tracking with higher velocities and accelerations, as in Figures 5.4 and 5.16. Adapting the weights of the optimization problem is the most obvious approach. A more sophisticated solution might be an adaptive-MPCBF where the weights would change depending on the current state of the aerial robot, but other solutions might exist.

The MPCBF did not always perform as well in the flight tests as it appeared in the simulations. In one of the experiments, the MPCBF consistently violated the CBF constraint, always in the same situation with the same repetitive outcome. As already mentioned, the problem might be a too short control horizon compared to the prediction horizon. While the MPCBF performed well in the simulations with a control horizon of 0.5 s compared to the prediction horizon of 2.5 s, the MPCBF could benefit from a longer control horizon in the experiments.

Possible ways forward for the controller include tests with multiple obstacles, as well as the implementation of the on-board camera for both pose estimation and obstacle detection. Another interesting topic would be, to modify the CBF constraint to achieve different goals, one could be to follow an object, at a fixed distance through an environment while avoiding all other obstacles.

While the MPCBF meets the requirements for an obstacle avoidance-driven controller for safety-critical aerial robots, there is still potential to further improve the controller and its obstacle avoidance capabilities.

## 6.2 Contributions

This thesis proposes the Model Predictive Control Barrier Function (MPCBF), a new controller designed to achieve reliable reference tracking, while avoiding all obstacles. The MPCBF combines the ideas of MPC with the strict safety constraints of a CBF. We derived the concept of a CBF and proved that it will guarantee safety for all time, thus avoiding all obstacles. Enforcing the CBF constraint for all times, we can guarantee, that the states of the aerial robot will stay inside a safe set. Since we are using an MPC, we predict a certain number of future states, for which we also enforce the CBF constraint. Therefore, we can find the optimal obstacle avoiding trajectory for all predicted states. Furthermore, we augmented the controller with different constraints, for example a CLF constraint to ensure exponential stability.

Optimal control is computationally more expensive than ordinary control laws. As the commonly available flight controllers are not powerful enough to solve the optimization problem in real time, we developed an aerial robot platform for optimal controllers. The new aerial robot is using a Jetson TX2 for increased computational power, as well as to provide the interface for additional equipment, such as a camera. The platform, while not necessary the first of its kind, is equipped with high computational power in a light-weight design. The integrated camera can be used for both obstacle detection and pose estimation for future work. We also used Ros, Mavlink with Mavros, and Px4 for the on-board intercommunication as well as the communication with an optional purely monitoring ground station. The developed system is fully autonomous in terms of control, it does not rely on another system to solve the optimization problem and calculate the control outputs.

To validate the MPCBF and compare it to another obstacle avoidance controller, namely the CLF-CBF-QP, we used multiple simulations. In the simulated scenarios, we observed that the MPCBF performs well above the capabilities of the CLF-CBF-QP. In the event, that the CLF-CBF-QP was infeasible, the MPCBF was not and could still avoid the obstacles. Judging by the predicted trajectories of the MPCBF, we concluded, that its advantages arises from its ability to optimize over a longer time period compared to the CLF-CBF-QP. As we applied the CBF constraint at all time steps, the MPCBF can choose a trajectory, results in future feasible state. The CLF-CBF-QP is incapable of doing so, as it only optimizes for the next time step. Thus, it sometimes reaches states with infeasible solutions. This is a clear benefit of the MPCBF.

In a series of experiments, we first proved the general flight capabilities of the aerial robot platform and the MPCBF. Afterwards, we used a single static obstacle to observe the MPCBF's behavior for obstacle avoidance. Using a circular trajectory, we ensured, that the aerial robot will cross paths with the obstacle multiple times. With this setup, we were able to confirm the avoidance performance of the MPCBF. We were able to prove, that the MPCBF safely avoids stationary obstacles. The final step was a relative motion between the aerial robot and the obstacle. In this scenario, however, the MPCBF did not always avoid the obstacle. We showed, there exists a repeating pattern of certain situations, in which the MPCBF collided with the obstacle. Afterwards, it would fly backwards, away from the obstacle it slightly touched, and successfully avoided it on the second try.

Summarizing the experiments, we showed that the MPCBF is well capable of obstacle avoidance, even if there are still some problems with the tuning for certain situations. We can conclude from the experiments and simulations, that the MPCBF should be able to safely navigate through denser environments, as long as the optimization problem can be solved in real-time. Therefore, the MPCBF meets our requirements for an obstacle avoidance-driven controller for safety-critical aerial robots. A proof of concept and proof of work are both achieved.

## 6.3 Outlook

As mentioned above, the MPCBF can fail to avoid obstacles reliable. This problem has to be further investigated. Afterwards, the MPCBF can be tested with multiple obstacles at once, static as well as dynamic obstacles. Parallel to that, the weights of the MPCBF have to be refined via further tuning. Since it appears that different weights are required for different flight situations, an Adaptive-MPCBF is a promising augmentation to adapt the MPCBF to different environments and situations. To overcome the constant loss of height, either an integrative term can be added to the MPCBF or a secondary control loop, which, based on the battery's voltage and feedback from the ESC, compensates for the decreasing real thrust.

Lastly, to achieve a completely autonomous flight of the aerial robot, the camera system has to be adapted to provide reliable pose and depth information. To receive additional global position information, a GPS module can be included in the design. Which is straightforward, since the Aerocore 2 already has the necessary connections and Px4 is able to stream the resulting position to the Jetson TX2. Moreover, a solution to the limited field-of-view of only one camera has to be developed. If all of the above is achieved, the aerial robot can navigate itself fully autonomously through an outdoor environment.

# Bibliography

[AcevedoEtAl13] Acevedo, J.J.; Arrue, B.C.; Maza, I.; Ollero, A.: Cooperative Large Area Surveillance with a Team of Aerial Mobile Robots for Long Endurance Missions. Journal of Intelligent & Robotic Systems, Vol. 70, No. 1-4, pp. 329–345, 2013.

[AlejoEtAl09] Alejo, D.; Conde, R.; Cobano, J.A.; Ollero, A.: Multi-UAV collision avoidance with separation assurance under uncertainties. In 2009 IEEE International Conference on Mechatronics, pp. 1–6, IEEE, 14.04.2009 - 17.04.2009.

[AmesEtAl19] Ames, A.D.; Coogan, S.; Egerstedt, M.; Notomista, G.; Sreenath, K.; Tabuada, P.: Control Barrier Functions: Theory and Applications. arXiv, 2019.

[AmesEtAl14a] Ames, A.D.; Galloway, K.; Sreenath, K.; Grizzle, J.W.: Rapidly Exponentially Stabilizing Control Lyapunov Functions and Hybrid Zero Dynamics. IEEE Transactions on Automatic Control, Vol. 59, No. 4, pp. 876–891, 2014.

[AmesEtAl14b] Ames, A.D.; Grizzle, J.W.; Tabuada, P.: Control barrier function based quadratic programs with application to adaptive cruise control. In 53rd IEEE Conference on Decision and Control, pp. 6271–6278, IEEE, 2014.

[ArduPilot09] ArduPilot: ArduPilot Open Source Autopilot: `http://ardupilot.org/`, 2009.

[BanguraEtAl14] Bangura, M.; Mahony, R.: Real-time Model Predictive Control for Quadrotors. IFAC Proceedings Volumes, Vol. 47, No. 3, pp. 11773–11780, 2014.

[BeardEtAl03] Beard, R.W.; McLain, T.W.: Multiple UAV cooperative search under collision avoidance and limited range communication constraints. In 42nd IEEE International Conference on Decision and Control (IEEE Cat. No.03CH37475), pp. 25–30, IEEE, Dec. 9-12, 2003.

[BorrmannEtAl15] Borrmann, U.; Wang, L.; Ames, A.D.; Egerstedt, M.: Control Barrier Certificates for Safe Swarm Behavior. IFAC-PapersOnLine, Vol. 48, No. 27, pp. 68–73, 2015.

[BovbelEtAl19] Bovbel, P.; Kotaru, P.V.; Jülg, C.; Palmerius, K.L.; Lange, J.: JohannLange/vrpn_client_ros: Release 1.0, 2019.

[BoydEtAl10] Boyd, S.; Vandenberghe, L.: Convex optimization. Cambridge: Cambridge Univ. Press, 8. printing. Edn., 2010.

[BuijsEtAl02] Buijs, J.; Ludlage, J.; van Brempt, W.; Moor, B.D.: Quadratic Programming in Model Predictive Control for Large Scale Systems. IFAC Proceedings Volumes, Vol. 35, No. 1, pp. 301–306, 2002.

[CaiEtAl14] Cai, G.; Dias, J.; Seneviratne, L.: A Survey of Small-Scale Unmanned Aerial Vehicles: Recent Advances and Future Development Trends. Unmanned Systems, Vol. 02, No. 02, pp. 175–199, 2014.

[CarloniEtAl13] Carloni, R.; Lippiello, V.; D'Auria, M.; Fumagalli, M.; Mersha, A.Y.; Stramigioli, S.; Siciliano, B.: Robot Vision: Obstacle-Avoidance Techniques for Unmanned Aerial Vehicles. IEEE Robotics & Automation Magazine, Vol. 20, No. 4, pp. 22–31, 2013.

[D'AndreaEtAl13] D'Andrea, R.; Hehn, M.; Brescianini, D.: Nonlinear Quadrocopter Attitude Control: Technical Report. ETH Zurich, 2013.

[DJI12] DJI: DJI releases "Flame Wheel" F330 ARF Kit: `https://www.dji.com/newsroom/news/dji-releases-flamewheel-f330-arf-kit/`, 10.02.1012.

[Emax18] Emax: EMAX MT2208 Multirotor Motor - Cooling Series: `https://emaxmodel.com/catalog/product/view/id/1584/s/emax-cooling-series-multicopter-motor-mt2208/`, 2018.

[GhamryEtAl17] Ghamry, K.A.; Kamel, M.A.; Zhang, Y.: Multiple UAVs in forest fire fighting mission using particle swarm optimization. In 2017 International Conference on Unmanned Aircraft Systems (ICUAS), pp. 1404–1409, IEEE, 13.06.2017 - 16.06.2017.

[GhandhariEtAl01] Ghandhari, M.; Andersson, G.; Hiskens, I.A.: Control Lyapunov functions for controllable series devices. IEEE Transactions on Power Systems, Vol. 16, No. 4, pp. 689–694, 2001.

[GiernackiEtAl17] Giernacki, W.; Skwierczynski, M.; Witwicki, W.; Wronski, P.; Kozierski, P.: Crazyflie 2.0 quadrotor as a platform for research and education in robotics and control engineering. In 2017 22nd International Con-ference on Methods and Models in Automation and Robotics (MMAR), pp. 37–42, IEEE, 28.08.2017 - 31.08.2017.

[GrimmEtAl05] Grimm, G.; Messina, M.J.; Tuna, S.E.; Teel, A.R.: Model predictive control: for want of a local control Lyapunov function, all is not lost. IEEE Transactions on Automatic Control, Vol. 50, No. 5, pp. 546–558, 2005.

[Gumstix18] Gumstix, I.: Introducting the AeroCore 2 - Launch your Drone, Robotics or IoT Device!: `https://www.gumstix.com/aerocore-2/`, 2018.

[Hall15] Hall, B.C.: Lie Groups, Lie Algebras, and Representations, Vol. 222. Cham: Springer International Publishing, 2. Edn., 2015.

[HehnEtAl11] Hehn, M.; D'Andrea, R.: A flying inverted pendulum. In A. Bicchi (Ed.) 2011 IEEE International Conference on Robotics and Automation, pp. 763–770, Piscataway, NJ: IEEE, 2011.

[Hildreth57] Hildreth, C.: A quadratic programming procedure. Naval Research Logistics Quarterly, Vol. 4, No. 1, pp. 79–85, 1957.

[HrabarEtAl10] Hrabar, S.; Merz, T.; Frousheger, D.: Development of an autonomous helicopter for aerial powerline inspections. In 2010 1st International Conference on Applied Robotics for the Power Industry (CARPI 2010), pp. 1–6, IEEE, 05.10.2010 - 07.10.2010.

[Hrabar08] Hrabar, S.: 3D path planning and stereo-based obstacle avoidance for rotorcraft UAVs. In 2008 IEEE/RSJ International Conference on Intelligent Robots and Systems, pp. 807–814, IEEE, 22.09.2008 - 26.09.2008.

[Intel18] Intel RealSense Technology: Intel® RealSense™ Depth Camera D435i: `https://www.intelrealsense.com/depth-camera-d435i/`, 2018.

[Intel19] Intel RealSense Technology: Intel® RealSense™ Tracking Camera T265: `https://www.intelrealsense.com/tracking-camera-t265/`, 2019.

[JiEtAl17] Ji, J.; Khajepour, A.; Melek, W.W.; Huang, Y.: Path Planning and Tracking for Vehicle Collision Avoidance Based on Model Predictive Control With Multiconstraints. IEEE Transactions on Vehicular Technology, Vol. 66, No. 2, pp. 952–964, 2017.

[Khalil02] Khalil, H.K.: Nonlinear systems. Upper Saddle River, NJ: Prentice Hall, 3. ed. Edn., 2002.

[KoyasuEtAl02] Koyasu, H.; Miura, J.; Shirai, Y.: Recognizing Moving Obstacles for Robot Navigation using Real-time Omnidirectional Stereo Vision. Journal of Robotics and Mechatronics, Vol. 14, No. 2, pp. 147–156, 2002.

[KoyasuEtAl03] Koyasu, H.; Miura, J.; Shirai, Y.: Mobile robot navigation in dynamic environments using omnidirectional stereo. In 2003 IEEE International Conference on Robotics and Automation, pp. 893–898, Piscataway, N.J: IEEE, 2003.

[KushleyevEtAl13] Kushleyev, A.; Mellinger, D.; Powers, C.; Kumar, V.: Towards a swarm of agile micro quadrotors. Autonomous Robots, Vol. 35, No. 4, pp. 287–300, 2013.

[KwonEtAl05] Kwon, W.H.; Han, S.H.: Receding Horizon Control. London: Springer-Verlag, 2005.

[LaiEtAl16] Lai, S.; Wang, K.; Qin, H.; Cui, J.Q.; Chen, B.M.: A robust online path planning approach in cluttered environments for micro rotorcraft drones. Control Theory and Technology, Vol. 14, No. 1, pp. 83–96, 2016.

[LangeEtAl19] Lange, J.; Meier, L.; Agar, D.; Küng, B.; Gubler, T.; Oes, J.; Babushkin, A.; Charlebois, M.; Bapst, R.; Mannhart, D.; Antener, A.D.; Goppert, J.; Goppert, A.; Riseborough, P.; Grob, M.; Whitehorn, M.; Wilks, S.; Mohammed, K.; Smeets, S.; Marques, N.; Kirienko, P.; Jansen, J.; Rivizzigno, M.; Gagne, D.; Siesta, B.; Guscetti, S.; de Souza, J.R.; Achermann, F.; Tobler, C.; Lecoeur, J.: JohannLange/-Firmware_Private: Release for Hybrid-Robotics-Lab based on v1.8.2, 2019.

[Lange19] Lange, J.: JohannLange/cbf_clf: Release v1.3, 2019.

[LeeEtAl10a] Lee, T.; Leok, M.; McClamroch, N.H.: Control of Complex Maneuvers for a Quadrotor UAV using Geometric Methods on SE(3). arXiv, 2010.

[LeeEtAl10b] Lee, T.; Leok, M.; McClamroch, N.H.: Geometric tracking control of a quadrotor UAV on SE(3). In 2010 49th IEEE Conference on Decision and Control, pp. 5420–5425, Piscataway, NJ: IEEE, 2010.

[MattingleyEtAl11] Mattingley, J.; Wang, Y.; Boyd, S.: Receding Horizon Control. IEEE Control Systems, Vol. 31, No. 3, pp. 52–65, 2011.

[MattingleyEtAl12] Mattingley, J.; Boyd, S.: CVXGEN: a code generator for embedded convex optimization. Optimization and Engineering, Vol. 13, No. 1, pp. 1–27, 2012.

[MeierEtAl15] Meier, L.; Honegger, D.; Pollefeys, M.: PX4: A node-based multithreaded open source robotics framework for deeply embedded platforms. In 2015 IEEE International Conference on Robotics and Automation (ICRA), pp. 6235–6240, IEEE, 26.05.2015 - 30.05.2015.

[MeierEtAl18] Meier, L.; Agar, D.; Küng, B.; Gubler, T.; Sidrane, D.; Oes, J.; Babushkin, A.; Charlebois, M.; Bapst, R.; Mannhart, D.; Antener, A.D.; Goppert, J.; Tridgell, A.; Riseborough, P.; Grob, M.; Whitehorn, M.; Wilks, S.; Mohammed, K.; Smeets, S.; Kirienko, P.; Marques, N.; Tobler, C.; Jansen, J.; Rivizzigno, M.; Gagne, D.; Siesta, B.; de Souza, J.R.; Achermann, F.; Lecoeur, J.: PX4/Firmware: v1.8.2 Stable Release, 2018.

[MellingerEtAl11] Mellinger, D.; Kumar, V.: Minimum snap trajectory generation and control for quadrotors. In A. Bicchi (Ed.) 2011 IEEE International Conference on Robotics and Automation, pp. 2520–2525, Piscataway, NJ: IEEE, 2011.

[MoeslundEtAl01] Moeslund, T.B.; Granum, E.: A Survey of Computer Vision-Based Human Motion Capture. Computer Vision and Image Understanding, Vol. 81, No. 3, pp. 231–268, 2001.

[MoriEtAl13] Mori, T.; Scherer, S.: First results in detecting and avoiding frontal obstacles from a monocular camera for micro unmanned aerial vehicles. In 2013 IEEE International Conference on Robotics and Automation, pp. 1750–1757, IEEE, 06.05.2013 - 10.05.2013.

[MuellerEtAl13] Mueller, M.W.; D'Andrea, R.: A model predictive controller for quadrocopter state interception. In 2013 European Control Conference (ECC), pp. 1383–1389, IEEE, 17.07.2013 - 19.07.2013.

[Mueller18] Mueller, M.W.: Multicopter attitude control for recovery from large disturbances. arXiv, 2018.

[Murphy12] Murphy, R.R.: A decade of rescue robots. In 2012 IEEE/RSJ International Conference on Intelligent Robots and Systems, pp. 5448–5449, IEEE, 07.10.2012 - 12.10.2012.

[MurrayEtAl94] Murray, R.M.; Li, Z.; Sastry, S.S.: A mathematical introduction to robotic manipulation. Boca Raton, Fla.: CRC Press, 1994.

[NocedalEtAl06] Nocedal, J.; Wright, S.J.: Numerical Optimization. Springer New York, 2006.

[NVIDIA19] NVIDIA Autonomous Machines: Jetson TX2 Module: `https://developer.nvidia.com/embedded/jetson-tx2/`, 2019.

[PalatEtAl05] Palat, R.C.; Annamalai, A.; Reed, J.H.: Cooperative Relaying for Ad-Hoc Ground Networks Using Swarm UAVs. In MILCOM 2005 - 2005 IEEE Military Communications Conference, pp. 1–7, IEEE, 17-20 Oct. 2005.

[PepyEtAl06] Pepy, R.; Lambert, A.; Mounier, H.: Path Planning using a Dynamic Vehicle Model. In 2nd information and communication technologies, 2006, pp. 781–786, Damascus: NOSSTIA, 2006.

[pixhawk18] pixhawk: Pixhawk: `http://pixhawk.org/`, 2018.

[PoundsEtAl06] Pounds, P.; Mahony, R.; Corke, P.: Modelling and control of a quad-rotor robot. In Australian Robotics and Automation Association Inc. (Ed.) Australasian Conference on Robotics and Automation 2006, Auckland, New Zealand: Australian Robotics and Automation Association Inc., 2006.

[PullenEtAl02] Pullen, K.; Bregler, C.: Motion capture assisted animation. In T. Appolloni (Ed.) Proceedings of the 29th annual conference on Computer graphics and interactive techniques - SIGGRAPH '02, p. 501, New York, New York, USA: ACM Press, 2002.

[QuigleyEtAl09] Quigley, M.; Conley, K.; P Gerkey, B.; Faust, J.; Foote, T.; Leibs, J.; Wheeler, R.; Y Ng, A.: ROS: an open-source Robot Operating System. ICRA Workshop on Open Source Software, Vol. 3, 2009.

[RöbenackEtAl14] Röbenack, K.; Winkler, J.; Franke, M.: Nonlinear control of complex systems using algorithmic differentiation. In Technische Universität Ilmenau (Ed.) 58th Ilmenau Scientific Colloquium, 2014.

[ROSflight19] ROSflight: A lean, open-source autopilot system built by researchers, for researchers: `http://rosflight.org/`, 2019.

[ShenEtAl15] Shen, C.; Shi, Y.; Buckham, B.: Model predictive control for an AUV with dynamic path planning. In 2015 54th annual conference of the Society of Instrument and Control Engineers of Japan (SICE), pp. 475–480, Piscataway, NJ: IEEE, 2015.

[SreenathEtAl13a] Sreenath, K.; Kumar, V.: Dynamics, Control and Planning for Cooperative Manipulation of Payloads Suspended by Cables from Multiple Quadrotor Robots. In Robotics: Science and Systems IX, Robotics: Science and Systems Foundation, June 24-28, 2013.

[SreenathEtAl13b] Sreenath, K.; Michael, N.; Kumar, V.: Trajectory generation and control of a quadrotor with a cable-suspended load - A differentially-flat hybrid system. In 2013 IEEE International Conference on Robotics and Automation, pp. 4888–4895, IEEE, 06.05.2013 - 10.05.2013.

[Stereolabs19] Stereolabs Inc.: Meet ZED Mini, the world's first camera for mixed-reality: `https://www.stereolabs.com/zed-mini/`, 2019.

[TagliabueEtAl19] Tagliabue, A.; Wu, X.; Mueller, M.W.: Model-free Online Motion Adaptation for Optimal Range and Endurance of Multicopters. In 2019 IEEE International Conference on Robotics and Automation, Vol. 2019, IEEE, 2019.

[TeeEtAl09a] Tee, K.P.; Ge, S.S.: Control of nonlinear systems with full state constraint using a Barrier Lyapunov Function. In Proceedings of the 48h IEEE Conference on Decision and Control (CDC) held jointly with 2009 28th Chinese Control Conference, pp. 8618–8623, IEEE, 15.12.2009 - 18.12.2009.

[TeeEtAl09b] Tee, K.P.; Ge, S.S.; Tay, E.H.: Barrier Lyapunov Functions for the control of output-constrained nonlinear systems. Automatica, Vol. 45, No. 4, pp. 918–927, 2009.

[ThomasEtAl13] Thomas, J.; Polin, J.; Sreenath, K.; Kumar, V.: Avian-Inspired Grasping for Quadrotor Micro UAVs. In Volume 6A: 37th Mechanisms and Robotics Conference, p. V06AT07A014, ASME, Sunday 4 August 2013.

[ThomasEtAl14] Thomas, J.; Loianno, G.; Polin, J.; Sreenath, K.; Kumar, V.: Toward autonomous avian-inspired grasping for micro aerial vehicles. Bioinspiration & biomimetics, Vol. 9, No. 2, p. 025010, 2014.

[TomicEtAl12] Tomic, T.; Schmid, K.; Lutz, P.; Domel, A.; Kassecker, M.; Mair, E.; Grixa, I.; Ruess, F.; Suppa, M.; Burschka, D.: Toward a Fully Autonomous UAV: Research Platform for Indoor and Outdoor Urban Search and Rescue. IEEE Robotics & Automation Magazine, Vol. 19, No. 3, pp. 46–56, 2012.

[WangEtAl10] Wang, Y.; Boyd, S.: Fast Model Predictive Control Using Online Optimization. IEEE Transactions on Control Systems Technology, Vol. 18, No. 2, pp. 267–278, 2010.

[WillsEtAl04] Wills, A.G.; Heath, W.P.: Barrier function based model predictive control. Automatica, Vol. 40, No. 8, pp. 1415–1422, 2004.

[WillsEtAl05] Wills, A.G.; Heath, W.P.: Application of barrier function based model predictive control to an edible oil refining process. Journal of Process Control, Vol. 15, No. 2, pp. 183–200, 2005.

[WuEtAl16a] Wu, G.; Sreenath, K.: Safety-Critical Control of a 3D Quadrotor With Range-Limited Sensing. In Volume 1: Advances in Control Design Methods, Nonlinear and Optimal Control, Robotics, and Wind Energy Systems; Aerospace Applications; Assistive and Rehabilitation

Robotics; Assistive Robotics; Battery and Oil and Gas Systems; Bioengineering Applications; Biomedical and Neural Systems Modeling, Diagnostics and Healthcare; Control and Monitoring of Vibratory Systems; Diagnostics and Detection; Energy Harvesting; Estimation and Identification; Fuel Cells/Energy Storage; Intelligent Transportation, p. V001T05A006, ASME, 2016.

[WuEtAl16b] Wu, G.; Sreenath, K.: Safety-critical control of a planar quadrotor. In 2016 American Control Conference (ACC), pp. 2252–2258, IEEE, 2016.

[WuEtAl18] Wu, Z.; Albalawi, F.; Zhang, Z.; Zhang, J.; Durand, H.; Christofides, P.D.: Control Lyapunov-Barrier Function-Based Model Predictive Control of Nonlinear Systems. In 2018 Annual American Control Conference (ACC), pp. 5920–5926, 2018.

[YamaneEtAl09] Yamane, K.; Hodgins, J.: Simultaneous tracking and balancing of humanoid robots for imitating human motion capture data. In 2009 IEEE/RSJ International Conference on Intelligent Robots and Systems, pp. 2510–2517, IEEE, 10.10.2009 - 15.10.2009.

[ZhangEtAl18] Zhang, Z.; Liu, S.; Tsai, G.; Hu, H.; Chu, C.C.; Zheng, F.: PIRVS: An Advanced Visual-Inertial SLAM System with Flexible Sensor Fusion and Hardware Co-Design. In K. Lynch (Ed.) 2018 IEEE International Conference on Robotics and Automation (ICRA), pp. 1–7, Piscataway, NJ: IEEE, 2018.